\documentclass[10pt,twocolumn,letterpaper]{article}

\usepackage[pagenumbers]{cvpr}

\usepackage{xspace}
\usepackage[dvipsnames]{xcolor}
\usepackage{comment}
\usepackage{adjustbox}
\usepackage{array}
\usepackage{makecell}
\usepackage{pifont}
\usepackage{gensymb}
\usepackage{multirow}
\usepackage{flushend}
\usepackage{colortbl}

\newcommand{\Rbb}{\ensuremath{\mathbb{R}}}

\newcommand{\inv}[1]{\ensuremath{\frac{1}{#1}}}

\newcommand{\norm}[1]{\ensuremath{\left\| #1\right\|}}
\newcommand{\abs}[1]{\ensuremath{\left| #1 \right|}}

\newcommand{\set}[1]{\ensuremath{\mathcal{#1}}}

\newcommand{\backthreed}{\ensuremath{\phi_{\rm 3D}}}
\newcommand{\backtwod}{\ensuremath{\phi_{\rm 2D}}}
\newcommand{\headthreed}{\ensuremath{\psi_{\rm 3D}}}

\newcommand{\featthreed}{\ensuremath{f}}
\newcommand{\feattwod}{\ensuremath{g}}
\newcommand{\dimfeatthreed}{\ensuremath{F_{\rm 3D}}}
\newcommand{\dimfeattwod}{\ensuremath{F_{\rm 2D}}}
\newcommand{\proj}[1]{\tilde{#1}}

\def\ours{ScaLR\xspace}
\definecolor{myblue}{rgb}{0.19, 0.55, 0.91}
\definecolor{myred}{rgb}{0.82, 0.1, 0.26}
\newcommand{\cmark}{\textcolor{myblue}{\ding{51}}}
\newcommand{\xmark}{\textcolor{myred}{\ding{55}}}
\newcommand{\smallparagraph}[1]{\medskip\noindent\textbf{#1}~}
\makeatletter
\renewcommand\smallparagraph{\@startsection
    {paragraph}{4}{\z@}%
    {0.3ex \@plus0.5ex \@minus.2ex}%
    {-1em}%
    {\normalfont\normalsize\bfseries}}
\makeatother

\definecolor{cvprblue}{rgb}{0.21,0.49,0.74}
\usepackage[pagebackref,breaklinks,colorlinks,citecolor=cvprblue]{hyperref}

\usepackage[capitalize]{cleveref}
\crefname{section}{Sec.}{Secs.}
\Crefname{section}{Section}{Sections}
\Crefname{table}{Table}{Tables}
\crefname{table}{Tab.}{Tabs.}

\title{Three Pillars improving Vision Foundation Model Distillation for Lidar}

\author{%
Gilles Puy$^1$
\and
Spyros Gidaris\textsuperscript{1}
\and
Alexandre Boulch\textsuperscript{1}
\and
Oriane Siméoni\textsuperscript{1}
\and
Corentin Sautier\textsuperscript{1,3}
\and
Patrick Pérez\textsuperscript{2}\thanks{Work done at valeo.ai.}
\and
Andrei Bursuc\textsuperscript{1}
\and
Renaud Marlet\textsuperscript{1,3}
\vspace{1mm}
\and
\textsuperscript{1}valeo.ai, Paris, France \hspace{5mm} \textsuperscript{2}Kyutai, Paris, France
\vspace{1mm}
\\
\textsuperscript{3}LIGM, Ecole des Ponts, Univ Gustave Eiffel, CNRS, Marne-la-Vall\'ee, France
}

\begin{document}
\maketitle

\begin{abstract}
Self-supervised image backbones can be used to address complex 2D tasks (e.g., semantic segmentation, object discovery) very efficiently and with little or no downstream supervision. Ideally, 3D backbones for lidar should be able to inherit these properties after distillation of these powerful 2D features. The most recent methods for image-to-lidar distillation on autonomous driving data show promising results, obtained thanks to distillation methods that keep improving. Yet, we still notice a large performance gap when measuring the quality of distilled and fully supervised features by linear probing. In this work, instead of focusing only on the distillation method, we study the effect of three pillars for distillation: the 3D backbone, the pretrained 2D backbones, and the pretraining dataset. In particular, thanks to our scalable distillation method named ScaLR, we show that scaling the 2D and 3D backbones and pretraining on diverse datasets leads to a substantial improvement of the feature quality. This allows us to significantly reduce the gap between the quality of distilled and fully-supervised 3D features, and to improve the robustness of the pretrained backbones to domain gaps and perturbations. The code is available at \url{https://github.com/valeoai/ScaLR}.
\end{abstract}

\section{Introduction}

\begin{figure*}
\centering
    \begin{minipage}{0.22\linewidth}
    \centering
    \small nuScenes\\
    \includegraphics[width=\linewidth]{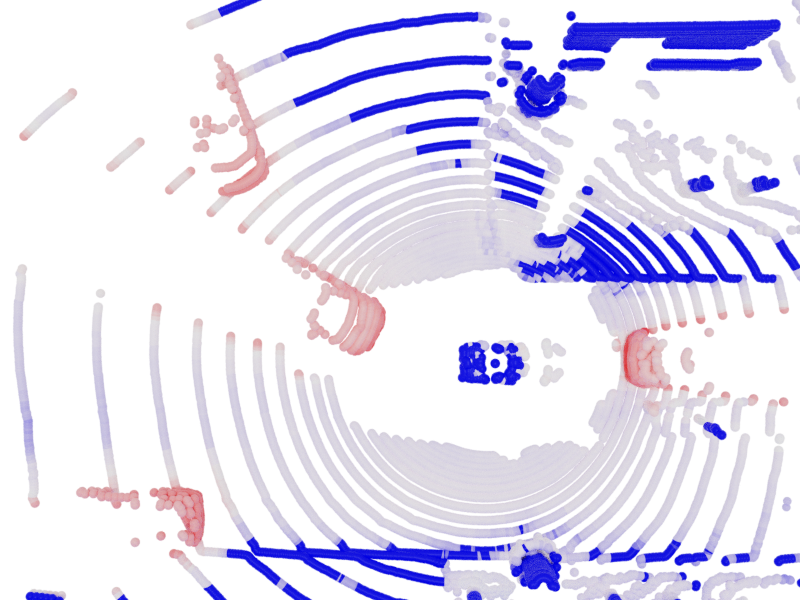}
    \end{minipage}
    \hfill
    \begin{minipage}{0.22\linewidth}
    \centering
    \small SemanticKITTI\\
    \includegraphics[width=\linewidth]{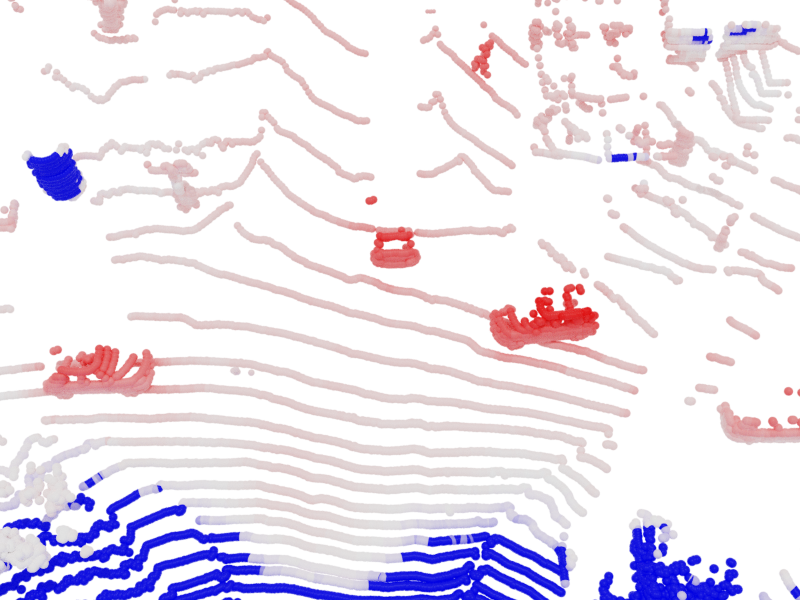}
    \end{minipage}
    \hfill
    \begin{minipage}{0.22\linewidth}
    \centering
    \small Pandar 64\\
    \includegraphics[width=\linewidth]{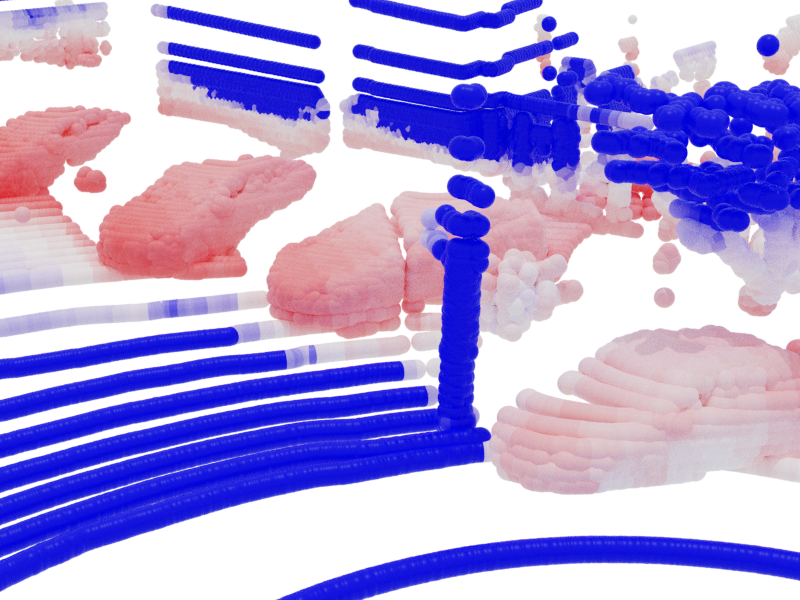}
    \end{minipage}
    \hfill
    \centering
    \begin{minipage}{0.22\linewidth}
    \centering
    \small Pandar GT\\
    \includegraphics[width=\linewidth]{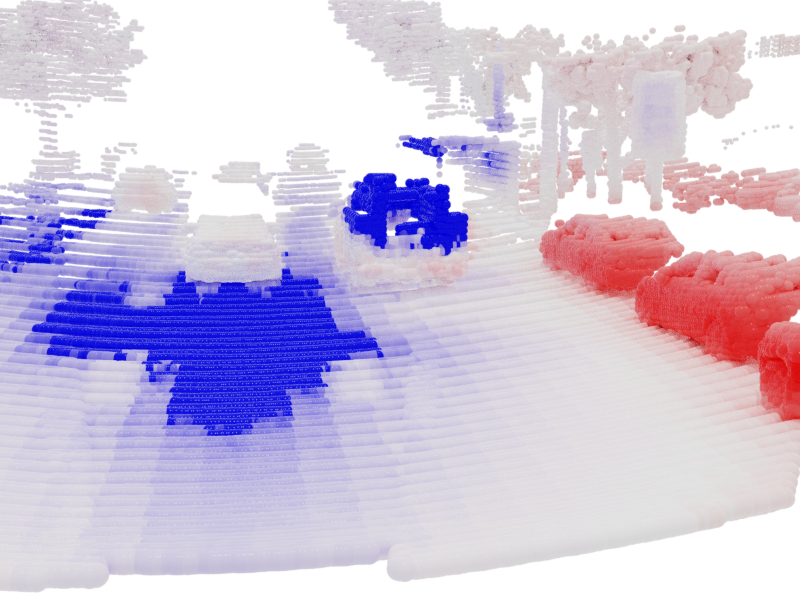}
    \end{minipage}
    \caption{\textbf{Correlation properties of distilled 3D features}. Correlation maps with a point located on a car on four different scenes extracted from nuScenes \cite{nuscenes}, SemanticKITTI \cite{behley2019semantickitti}, Pandar64 and PandarGT \cite{pandaset}, respectively. The features used to compute these maps are extracted from a \emph{single} pretrained backbone on all four datasets with \ours. Color goes from blue to red for low and high values.}
    \label{fig:scan_degree_cor}
\end{figure*}

Lidars capture the 3D geometry of a scene with high accuracy while being little sensitive to adverse light conditions, which is useful for advanced driver assistance systems. However, annotating lidar point clouds to train deep neural networks is notoriously long and expensive \cite{behley2019semantickitti}.

More frugal learning can be achieved by pretraining backbone networks with self-supervision on a pretext task~\cite{caron2021dino, caron2020unsupervised, grill2020byol, he2020momentum, he2022masked} and a non-annotated dataset. The pretrained network can then be finetuned on various downstream tasks and other datasets, with much less supervision for the same level of performance, or with a higher performance than when trained from scratch. Such pretraining has been particularly successful for image backbones~\cite{grill2020byol, chen2021mocov3, caron2021dino}. In particular, the gap between supervised and self-supervised representations, as evaluated by linear probing, has been closed on ImageNet~\cite{caron2021dino, caron2020unsupervised, chen2021mocov3, grill2020byol, he2020momentum, gidaris2021obow}.

On autonomous driving (AD) data, we distinguish mainly two categories of self-supervised methods for 3D backbones: \emph{(1)} methods leveraging only lidar data and defining a pretext task at the level of a single \cite{depthcontrast,segcontrast,also} or multiple scans \cite{pointcontrast,bevcontrast,tarl,stssl}, and \emph{(2)} methods exploiting images acquired in synchronization with the point clouds and distilling self-supervised image representations to a 3D backbone \cite{ppkt,sautier22SLidR,mahmoud2023st_SLidR,liu2023segment}. Our focus here is on the second category of methods, thanks to which 3D backbones should be able to inherit the powerful properties of self-supervised image backbones. 

Image-to-lidar distillation has improved a lot recently, in particular by designing better distillation losses. Nevertheless, we still observe a large gap between distilled and supervised 3D representations when directly measuring their quality by linear probing: respectively $45.0\%$ and $74.7\%$ mIoU in linear probing on the validation of nuScenes in \cite{liu2023segment} for a MinkUNet \cite{minkowskicnn} with cylindrical voxels \cite{cylinder3d}.

In this work, in order to improve the quality of distilled features, we explore the effect of three other pillars, instead of focusing only on the distillation method. These pillars are: (i) the 3D backbone, (ii) the pretrained 2D backbones and (iii) the pretraining dataset. In particular, we show that scaling the 2D and 3D backbones and pretraining on diverse datasets
leads to a considerable improvements of the feature quality. The role of these pillars is actually more important than the distillation method itself, which we propose to simplify for easier scaling. 

After proposing and studying a scalable distillation method, which we call \ours for \underline{Sca}labale \underline{L}idar \underline{R}epresentation, we make the following contributions.

First, we are able to significantly reduce the gap between distilled and supervised lidar representations. We reach an mIoU of $67.8\%$ in linear probing on the validation set of nuScenes with a WaffleIron-48-768 backbone \cite{puy23waffleiron}, i.e., an increase of $22.8$ points compared to the score obtained with the best distillation method \cite{liu2023segment}. We recall that the best\footnote{without test time augmentation.} reported mIoU on the validation of nuScenes obtained under supervision is $78.4\%$ \cite{spheretransformer}, to the best of our knowledge. We are thus only about $10$ points away from this upper bound. 

Second, we show that it is possible to pretrain a \emph{single} backbone on a mixture of datasets, performing similarly or better than different backbones specialized on each dataset individually. The capacity of this backbone in providing good features across multiple datasets in illustrated in \cref{fig:scan_degree_cor}. For each scene in this figure, we pick a point located on a car and present the feature correlation map with respect to this point. We notice that the most correlated points also belong to cars on all datasets, illustrating the capacity of our single pretrained backbone to correctly distinguish objects on multiple datasets.

Third, we thoroughly study the properties of our distilled features. We show that they are robust to both domain gaps and perturbations, actually leading to new state-of-the-art results on the benchmark of Robo3D \cite{robo3d}. We also show that pretraining on diverse datasets improves the robustness. 

Finally, we show that a possible way to get even better features is to distill the knowledge from multiple vision foundation models at the same time, which can be easily done with our scalable distillation strategy.

\section{Related work}

\smallparagraph{Pretraining 2D backbones.}
Supervised pretraining of 2D networks~\cite{krizhevsky2012imagenet, he2016deep, dosovitskiy2020image} with large-size data~\cite{imagenet} produces powerful image representations that can transfer well to various downstream tasks in the image domain. Alternatively, self-supervised pretraining has been proven able to rival or surpass supervised pretraining while using only raw unlabeled image data. Two prominent approaches conduct self-supervised pretraining via either discriminative tasks that train a 2D network to extract features invariant to augmentations~\cite{chen2020simple, he2020momentum, grill2020byol, caron2021dino}, or via masked image modeling tasks~\cite{bao2022beit, he2022masked} that hide part of an image and then train a network to reconstruct the missing fragment. An emerging pretraining paradigm is via language-image contrastive learning (CLIP) \cite{radford2021clip}, in which the 2D network is trained with a text network to match image data with their paired captions. 

\smallparagraph{Pretraining 3D backbones.}
Several self-supervised techniques for training 3D backbones have appeared recently. While the first techniques were often limited to dense scans of single objects~\cite{chen2021shape,poursaeed2020self,sauder2019self,depthcontrast}, contrastive self-supervision has enabled significant improvements in self-supervision on large indoor and outdoor datasets~\cite{pointcontrast,depthcontrast,proposalcontrast,segcontrast,tarl,liu2023fac}. Instead of working on a single scan, some works also take advantage of the temporal dimension to construct their contrastive pretext tasks \cite{strl,tarl,stssl,bevcontrast}. Finally, other strategies consist in reconstructing missing information hidden either artificially \cite{voxelmae1,voxelmae2,pointmae,pointm2ae,pointbert}, or intrinsically because of the sparse point sampling in lidar acquisitions \cite{also}.

\smallparagraph{Using 2D pretrained features to train 3D backbones.}
Exploiting pretrained image backbones to train 3D networks has drawn a lot of attention recently. For example, in \cite{zhang2022learning,dong2022autoencoders}, 3D autoencoders are pretrained on object-level point clouds where features extracted from a 2D backbone are used to build the supervision signal. Several methods \cite{sautier22SLidR,mahmoud2023st_SLidR,ppkt,liu2023segment} distill the knowledge of 2D self-supervised backbones on large indoor or outdoor point clouds thanks to a contrastive loss that aligns 2D and 3D representations for pairs of corresponding point and pixel, while making the representation as dissimilar as possible for non matching point-pixel pairs. The same principle can also be applied for object-level point cloud \cite{afham2022crosspoint}. Another type of strategy consists in pseudo-labeling 3D point clouds with class labels provided by 2D backbones trained under supervision \cite{yu2022data}. Recently, several methods, e.g., \cite{zhang2022pointclip,chen2023clip2scene, Peng2023OpenScene}, proposed to distill the knowledge of supervised vision-language models for open-vocabulary recognition or semantic segmentation. In addition to distillation, these methods also need to preserve alignment with the text representations. While our focus is not on open-vocabulary semantic segmentation, it is probable that some recipes discovered in this work can improve the distillation of vision-language models as well.

\smallparagraph{Robustness.} Making sure that trained models are robust to various type of perturbations is important for safety-critical applications, such as advanced driver assistance systems. Recently, several works \cite{Yu_2023_CVPR,shuangzhi23,albreiki22,Dong_2023_CVPR,yan2023benchmarking} have proposed ways to evaluate or improve robustness of 3D perception models. In this work, we evaluate the robustness of our pretrained models on the Robo3D benchmark~\cite{robo3d}, which considers eight corruption types of different intensities meant to mimic different cases and degrees of distribution shifts from the original training distribution. As mentioned earlier, we show that distilling 2D features on data collected from multiple different lidars improves the robustness of our semantic segmentation models.

\smallparagraph{Domain generalization.} 3D backbones trained on lidar data are sensitive to domain shifts such as the one induced by the different lidar sensors. Several techniques have thus been developed to improve the generalization capabilities of 3D models, e.g., \cite{Kim_2023_CVPR, Sanchez_2023_ICCV,Xiao_2023_CVPR,Saltori_2023_ICCV}. Alternatively, some techniques adapt these models to a new target domain using non-annotated target lidar data only, e.g., \cite{completeandlabel,cosmix,saluda}, or with the help of image data, e.g., \cite{peng2021sparse,jaritz2020xmuda,LIU2021211} . While our method is not a domain generalization or adaptation technique per se, we consider the protocol used in this related literature to evaluate the capacity of our pretrained backbones to generalize to different lidar sensors.

\section{Scalable Distillation Strategy}
\label{sec:method}

\subsection{Motivation and Principle}
\label{sec:motivation}

Our goal is to study the benefit of using high-capacity 2D and 3D backbones, and of mixing data acquired by multiple lidars. To facilitate this study, it is important to use a distillation strategy that is scalable, with no or few hyperparameters to tune on each dataset, while remaining competitive.

While the state-of-the-art methods \cite{ppkt,sautier22SLidR,mahmoud2023st_SLidR,liu2023segment} use a contrastive loss defined at the level of semantically coherent segments, in this work we relate pixel and point features directly using a simple cosine similarity-based loss, whose usage actually appears naturally when distilling CLIP features \cite{Peng2023OpenScene}. This loss has the advantage to have no hyperparameters and, as we will see later, is very competitive without the cost of pre-computing segments.

Finally, AD datasets often contain images captured by multiple cameras synchronized with the lidar. The best practice for distillation is, for each lidar scan, to load all available synchronized images during batch loading. Nevertheless, the number of cameras varies from one dataset to the other, which creates some implementation difficulties when pretraining on a mix of different datasets, as in this work. Instead, we propose to have batches where each sample is created as follows: one camera is selected at random and we only load the corresponding image and points viewed in this camera. Therefore, data loading becomes strictly identical whatever the number of cameras available in the pretraining dataset, which makes multi-datasets pretraining easier to implement. Furthermore, this strategy saves memory, making it easier to train large backbones.

\subsection{Formal description} 

\smallparagraph{Backbones.} The 3D backbone $\backthreed$ takes as input a point cloud $(p_1,\ldots, p_N) \in \Rbb^{N \times (3 + 1)}$, where each point $p_i$ holds three Cartesian coordinates and the laser return intensity. It outputs deep features of dimension~$\dimfeatthreed$ for all input points: $(\featthreed_1,\ldots, \featthreed_N) \in \Rbb^{N \times \dimfeatthreed}$. Similarly, the 2D backbone $\backtwod$ takes as input an RGB image of $M$ pixels $(u_1, \ldots, u_M) \in \Rbb^{M \times 3}$ and outputs deep features of dimension~$\dimfeattwod$ for all pixels: $(\feattwod_1,\ldots, \feattwod_M) \in \Rbb^{M \times \dimfeattwod}$. In practice, we use bilinear interpolation to increase the resolution of the 2D feature map to the resolution of the input image.

\smallparagraph{3D Projection head.} A linear projection head is added at the end of the 3D networks. We denote this projection head by $\headthreed$. At distillation time, it projects 3D features into the output space of the 2D backbone. The projection head $\headthreed$ is removed after distillation and replaced by another one dedicated for the task of interest. We denote $(\proj\featthreed_1,\ldots, \proj\featthreed_N) \in \Rbb^{N \times \dimfeattwod}$ the output of $\headthreed \circ \backthreed$.

\smallparagraph{Point-pixel mapping.} We assume the lidar and the camera (selected at random among all available cameras during batch creation) are calibrated so that we can put in correspondence 3D points and pixels. We can thus compute a point-to-pixel mapping $\rho: \{1, \ldots, N\} \rightarrow \{-1\} \cup \{1, \ldots, M\}$. We are then able to retrieve the pixel feature $\feattwod_{\rho(i)}$ associated to each point $p_i$. By convention, $\rho(i) = -1$ means that the point $p_i$ is not viewed in the camera.

\smallparagraph{Similarity loss.} In our experiments, we use the following loss, which relies on pointwise cosine similarity
\begin{equation}
\label{eq:cosine_loss}
    \set{L}_{\rm sim} = \inv{\abs{\set{V}}} \sum_{i \in \set{V}} \norm{\proj\featthreed_i - \feattwod_{\rho(i)}}_2,
\end{equation}
where $\proj\featthreed_i$ and $\feattwod_j$ have been $\ell_2$-normalized beforehand, and $\set{V}$ is the subset of all visible points in the camera, i.e., such that $\rho(i) \neq -1$.

\begin{table}[t]
\centering
\small
\setlength{\tabcolsep}{4.5pt}
\begin{tabular}{lcccccc}
\toprule
    & \makecell{2D\\head}
    & Con.
    & Cos.
    & \makecell{No.\\cam.}
    & {Mem.~/~Time $\downarrow$}
    & mIoU\% $\uparrow$
\\
\midrule
\multirow{4}{*}{\rotatebox[origin=c]{90}{\scriptsize MinkUNet}}
    & \cmark 
    & \cmark
    & -
    & 6
    & 1.0 ~/~ 1.0 
    & 39.3
\\
    &\xmark
    & \cmark
    & -
    & 6
    & 1.6 ~/~ 1.0 
    & 43.4
\\
    &\xmark
    & -
    & \cmark
    & 6
    & 1.6 ~/~ 1.1 
    & 43.6 
\\
&\xmark
    & -
    & \cmark
    & 1
    & 0.8 ~/~ 0.3
    & 42.1
\\
\midrule
\multirow{3}{*}{\rotatebox[origin=l]{90}{\scriptsize WI-48-256}}
    & \cmark 
    & \cmark
    & -
    & 1
    & -
    & 50.0 {\scriptsize ($\pm 0.8$)}
\\
    &\xmark 
    & \cmark
    & -
    & 1
    & -
    & 54.1 {\scriptsize \color{white} ($\pm 0.0$)}
\\
    &\xmark 
    & -
    & \cmark
    & 1
    & -
    & 56.7 {\scriptsize ($\pm 0.9$)}
\\
\bottomrule
\end{tabular}
\caption{\textbf{Analysis of distillation losses}. Effect on the linear probing performance (mIoU), memory usage, training speed of the distillation loss and number of images loaded per scan. We show relative gain in memory and training time with respect to the first row (lower value = less consumption). We report the standard deviation between parentheses evaluated over 3 different pretrainings for some selected experiments.}
\label{tab:distillation-loss}
\end{table}

\subsection{Analysis of our Distillation Strategy}
\label{sec:analysis_distillation}

In this section, we conduct an analysis of different pretraining options to validate our choice of distillation strategy.

We conduct our experiments on 2 backbones: the MinkUNet \cite{minkowskicnn} with cylindrical voxels \cite{zhu2021cylindrical} used in \cite{sautier22SLidR,mahmoud2023st_SLidR,liu2023segment} and WaffleIron-48-256 \cite{puy23waffleiron} (48 layers with 256-dim.\ features). We perform our study on nuScenes \cite{nuscenes}. We split the original 700 training scenes in a mini-train set and a mini-val set of 600 and 100 scenes, respectively (see \cite{sautier22SLidR} for details). All backbones are pretrained and linearly probed on the mini-train set and the results are reported on the mini-val set. For all settings, we distill features obtained from the DINO-pretrained ViT-S/8 \cite{caron2021dino}.

The results are presented in \cref{tab:distillation-loss}. We start from the baseline \cite{ppkt} which uses a contrastive loss and a 2D projection head on top of the 2D backbone. We then remove this 2D projection head and finally replace the loss by ours (Eq.\,\ref{eq:cosine_loss}). Note that the use of a 2D projection head is not compatible with \eqref{eq:cosine_loss} as an optimal but degenerated solution would be obtained by letting the parameters of the 2D and 3D projection heads go to zero. Finally, we measure the effect of loading only one image instead of six per batch. Based on the results of \cref{tab:distillation-loss}, we notice that our distillation strategy performs well thanks to the following advantages. 

\smallparagraph{Preservation of the 2D feature space.} The results in \cref{tab:distillation-loss} show a clear detriment of the 2D projection for this unidirectional image-to-lidar distillation. The absence of 2D projection head preserves the structure of the 2D feature space: there is no loss of information on the 2D side anymore. Leveraging as much as possible the original 2D features, obtained after pretraining on very large image datasets, is thus key to produce good 3D features.

\smallparagraph{Absence of false negatives.} Even in absence of 2D projection head, the loss \eqref{eq:cosine_loss} performs better than the contrastive loss. We believe that this can be explained by the presence of false negatives in the contrastive loss. Indeed, with the cosine similarity loss, a feature $f_i$ of a 3D point $p_i$ sampled, e.g., on a car will be made as similar as possible to the corresponding 2D feature $g_{\rho(i)}$ falling at the same location on the car. However, with the contrastive loss, an additional mechanism comes into play: the feature $f_i$ will also be made as dissimilar as possible to all other 2D features $g_{\rho(j)}, j \neq i$, even when these 2D features are originally similar, e.g., when they fall on the same object. This known drawback of the contrastive loss seems to harm its performance significantly. 

\smallparagraph{Scalability to large point-pixel pairs.} A drawback of the contrastive loss is that it cannot be computed using all possible point-pixel pairs because of memory constraints. One needs to subsample these pairs, e.g., randomly but then increasing the risk of ignoring small objects in the loss. The effects of this sub-optimality have been reduced by constructing the loss at the level of semantically coherent segments in \cite{sautier22SLidR,mahmoud2023st_SLidR,liu2023segment}, but at the cost of extra pre-processing time and additional parameters to tune. This is a cost we wish to avoid to be able to pretrain more easily on multiple datasets. The cosine similarity loss does not require selecting point-pixel pairs or pre-extracting image/lidar segments.

\smallparagraph{Reduced memory usage.} nuScenes provides images acquired from six different cameras, offering a 360\degree{} view of the scene. To reduce the memory requirement, we propose to use only one image (chosen at random among all available cameras) and only the points viewed in this camera when loading one element of a batch. Our result shows a slight drop of performance when batching with single cameras, but a significant gain in speed and memory. As explained in \cref{sec:motivation}, this change has several benefits. First, the savings in memory and compute time make it easier to train large backbones, which, as we will see, is essential to reach good performance. Second, the data loading process becomes strictly identical whatever the number of cameras available in the pretraining dataset, which makes multi-datasets pretraining easier to implement.

To summarize, our distillation strategy does not suffer from the presence of false negatives, does not require a selection of point-pixel pairs, does not need extra pre-processing time (e.g., extracting segments), does not introduce extra hyperparameters to tune (such as the temperature in the contrastive loss or number of segments), and saves compute time and memory.

\section{Experiments}
\label{sec:experiments}

In this section, we show that scaling the 2D and 3D backbones and pretraining on diverse datasets leads to a substantial improvement of the feature quality. First, we describe the backbones and datasets that we use. Second, we show that our scalable distillation strategy is competitive compared to other state-of-the-art methods. Third, we scale the 2D and 3D backbones and show that it results in distilled features of much better quality. Fourth, we pretrain a single backbone on multiple datasets captured by four different lidars and demonstrate that it performs as well as, or better than, backbones specialized to each lidar. Fifth, we evaluate the downstream finetuning performance, robustness and generalization properties of our pretrained backbones. Finally, we show that our method allows an easy combination of multiple 2D teachers to further boost the performance.

\subsection{Three Pillars}

\smallparagraph{3D backbones.} We experiment with two very different architectures designed for 3D point clouds: MinkUNet \cite{minkowskicnn} with cylindrical voxels \cite{zhu2021cylindrical} and WaffleIron (WI) \cite{puy23waffleiron}. For WI, we fix the depth to $48$, as it led to the best results in \cite{puy23waffleiron}, and vary the feature size: $\dimfeatthreed \in \{96, 256, 384, 768\}$. We denote the corresponding backbone by WI-$\dimfeatthreed$. Note that we noticed numerical instabilities during training at $\dimfeatthreed = 768$, which we solved by replacing all batchnorms by layernorms (except in the embedding and classification layers) at this feature size.

\smallparagraph{2D backbones.} We concentrate on the distillation of self-supervised ViT features obtained with DINO~\cite{caron2021dino}, DINOv2~\cite{oquab2023dinov2} or MAE~\cite{he2022masked}. Nevertheless, we also conduct one experiment with the MaskCLIP ViT-B/16 \cite{zhou2022extractMaskclip} to show the benefit of using multiple 2D teachers.

\smallparagraph{Pretraining datasets and lidars.} We conduct experiments by pretraining on data collected from four different lidars: Velodyne-32 of nuScenes~\cite{nuscenes}, Velodyne-64 of SemanticKITTI~\cite{behley2019semantickitti}, Pandar 64 and Pandar GT of PandaSet~\cite{pandaset}. On nuScenes and SemanticKITTI, we use the official train/val split and list of classes. We detail the train/val split and list of classes we used on PandaSet in the supplementary material as we noticed different practices on this dataset in the literature.

\subsection{Implementation Details}

For all experiments conducted with MinkUNet, we use the implementation provided by \cite{sautier22SLidR} and the same hyperparameters, unless otherwise stated. We describe below the main implementation details used to pretrain, linear probe and finetune WI backbones. More information, such as learning rate schedules, weight decay, number of epochs are provided in the supplementary material.

\smallparagraph{Data augmentation.} We do not apply any image augmentation, beyond systematic resizing to $224 \times 448$. During pretraining, finetuning and linear probing, we apply the following standard point cloud augmentations: random rotation around the $z$-axis, random flip of the $x$ and $y$ axes. During finetuning and linear/MLP probing, we also globally scale the coordinates by a random factor chosen uniformly in $[0.9, 1.1]$.

\smallparagraph{Linear probing.} We remove the projection layer $\headthreed$ and replace it by a batch normalization layer directly followed by a linear classification layer. We denote the combination of these last layers by $\kappa$. Note that the combination of batch normalization followed by linear classification acts as a linear layer at inference. The batch normalization layer makes the results less sensitive to the choice of downstream learning rate \cite{lee2023rethinking}. While maintaining $\backthreed$ fixed, we train $\kappa \circ \backthreed$ using ground-truth labels.

\smallparagraph{Finetuning.} For finetuning, we remove the distillation layer $\headthreed$ and replace it by a batch normalization layer directly followed by a linear classification layer. As before, we denote the combination of these last layers by $\kappa$ and we train $\kappa \circ \backthreed$ using ground-truth labels. The backbone is finetuned using a layer-wise learning rate decay~\cite{clark2020electra, balestriero2023cookbook}, a technique commonly used when finetuning pretrained ViT backbones~\cite{bao2022beit,he2022masked,zhou2022ibot}.

\begin{table}
\small
\centering
\setlength{\tabcolsep}{5pt}
\begin{tabular}{lllcc}
\toprule
3D Back.
    & Method
    & 2D Back.
    & Lin. Prob.
    & 1\%
\\
\midrule
\multirow{4}{*}{MinkUNet}
    & PPKT \cite{ppkt}
    & ResNet-50
    & \it 36.4 
    & \it 37.5
\\
    & SLidR \cite{sautier22SLidR}
    & ResNet-50
    & \it 38.8
    & \it 39.0 
\\
    & ST-SLidR \cite{mahmoud2023st_SLidR}
    & ResNet-50
    & \it 40.5
    & \it 40.8 
\\
& Seal \cite{liu2023segment}
    & ResNet-50
    & \it 45.0
    & \it 45.8 
\\
\midrule
\multirow{3}{*}{MinkUNet}
    & PPKT \cite{ppkt}
    & ViT-S/8
    & 38.6 
    & 40.6
\\
    & SLidR \cite{sautier22SLidR}
    & ViT-S/8
    & 39.3 
    & 39.0
\\
    & \textbf{\ours} (ours)
    & ViT-S/8
    & 42.4
    & 40.5
\\
\midrule
WI-96
    & \textbf{\ours} (ours)
    & ViT-S/8
    & 46.8
    & 38.8
\\
WI-256
    & \textbf{\ours} (ours)
    & ViT-S/8
    & 54.2
    & 41.4
\\
\bottomrule
\end{tabular}
\caption{\textbf{Comparison of different pretraining methods and effect of 2D/3D backbones}. The mIoU\% reported in italic are found in the literature while the others are our production. All methods are pretrained on the same 600 scenes of nuScenes. Linear probing is done using the 700 training scenes of nuScenes. Finetuning is done using only 1\% of nuScenes training dataset. All scores are reported on the official validation split.}
\label{tab:nuscenes_sota}
\end{table}

\subsection{Comparison of Distillation Methods}
\label{sec:comparison_distillation}

In this section, we verify that our distillation strategy remains competitive with respect to the state-of-the-art in image-to-lidar distillation. All experiments are conducted using the experimental protocol used in \cite{sautier22SLidR, mahmoud2023st_SLidR,liu2023segment}, i.e., pretraining on the reduced training set of nuScenes (600 scenes) defined in \cite{sautier22SLidR} and using the official train/val split for downstream linear probing and finetuning.

We start by distilling the representations of DINO ViT-S/8 in a MinkUNet. These representations are distilled using three different methods: PPKT~\cite{ppkt}, SLidR~\cite{sautier22SLidR} and using our cosine-based similarity loss (but still loading all six images as in PPKT and SLidR). The MinkUNet backbones are pretrained with a batch size of 12 (we scale the learning rate proposed in \cite{sautier22SLidR} accordingly) during 25 epochs. The quality of the distilled features is assessed by linear probing. We also finetune the backbones using 1\% the complete training set of nuScenes. We report the corresponding scores in \cref{tab:nuscenes_sota}, where we also present the results obtained in the literature by distilling ResNet-50 representations.

First, we notice that changing the 2D backbones from ResNet-50 to ViT-S/8 helps both PPKT and SLidR, which both reach a higher mIoU in linear probing and finetuning after this change. The improvement seems however more significant for PPKT than for SLidR.

Second, we remark that \textbf{\ours leads to better 3D features than those obtained with PPKT and SLidR}: we reach a higher score in linear probing; we also obtain a better result in finetuning with 1\% of the training data than SLidR and a level of performance similar to PPKT. This confirms that using the simple cosine-based similarity loss of \cref{eq:cosine_loss} is a competitive choice.

\begin{table}[t]
\small
\centering
\setlength{\tabcolsep}{8.5pt}
\begin{tabular}{lllrlr}
\toprule
\multirow{3}{*}{3D Back.}
    & \multirow{3}{*}{2D Back.}
    & \multicolumn{2}{c}{nuScenes}
    & \multicolumn{2}{c}{SemKITTI}
\\
\cmidrule(lr){3-4}
\cmidrule(lr){5-6}
    & 
    & LP
    & \textcolor{myblue}{$\Delta\downarrow$}
    & LP
    & \textcolor{myblue}{$\Delta\downarrow$}
\\
\midrule
\rowcolor{gray!20}\multicolumn{6}{l}{No pretraining. Best fully supervised WI.}
\\
WI-$\dimfeatthreed$
    & --
    & \multicolumn{2}{l}{78.7 \tiny ($\dimfeatthreed$ = 768)}
    & \multicolumn{2}{l}{65.1 \tiny ($\dimfeatthreed$ = 256)}
\\
\midrule
\rowcolor{gray!20}\multicolumn{6}{l}{MAE-pretrained}
\\
WI-256
    & ViT-B/16
    & 47.8
    & \textcolor{myblue}{30.9}
    & --
    & --
\\
\rowcolor{gray!20}\multicolumn{6}{l}{DINO-pretrained}
\\
WI-256
    & ViT-S/8
    & 54.4
    & \textcolor{myblue}{24.3}
    & --
    & --
\\
WI-384
    & ViT-S/8
    & 57.8
    & \textcolor{myblue}{20.9}
    & --
    & --
\\
WI-768
    & ViT-S/8
    & 61.1
    & \textcolor{myblue}{17.6}
    & --
    & --
\\
\rowcolor{gray!20}\multicolumn{6}{l}{DINOv2-pretrained}
\\
WI-256
    & ViT-S/14
    & 57.7
    & \textcolor{myblue}{21.0}
    & --
    & --
\\
WI-256
    & ViT-B/14
    & 60.2
    & \textcolor{myblue}{18.5}
    & --
    & --
\\
WI-256
    & ViT-L/14
    & 61.6
    & \textcolor{myblue}{17.1}
    & 48.3
    & \textcolor{myblue}{16.8}
\\
WI-384
    & ViT-L/14
    & 63.9
    & \textcolor{myblue}{14.8}
    & 50.6
    & \textcolor{myblue}{14.5}
\\
WI-768
    & ViT-L/14
    & \textbf{66.8}
    & \textcolor{myblue}{\bf11.9}
    & \textbf{51.1}
    & \textcolor{myblue}{\bf14.0}
\\
\bottomrule
\end{tabular}
\caption{\textbf{Effect of the 3D network width and of the 2D features' quality}. We report the mIoU\% obtained by linearly probing distilled features. The networks are trained on nuScenes or SemanticKITTI using \ours. The scores are reported on the corresponding official validation split. The column \textcolor{myblue}{$\Delta$} shows the gap with respect to the best fully supervised mIoU\% reached with WI on the corresponding dataset.}
\label{tab:capacity}
\end{table}

Third, we also pretrained WI-96 and WI-256 backbones on this dataset using our full scalable distillation protocol. We notice that \textbf{changing MinkUNet to WI further boosts linear probing performance}. We remark that WI-96 has the same output feature dimension than the MinkUNet backbone but fewer trainable parameters (1M vs 35M). Hence, the improvement of performance in linear probing cannot be solely explained by the 3D feature dimension or the capacity of the backbone; other elements in the architecture must come into play but identifying the exact origin of this improvement is beyond the scope of this work. 

Fourth, when finetuning on 1\% of nuScenes training set, \textbf{our \ours WI-256 model surpasses prior methods}. The only exception is Seal \cite{liu2023segment} that leverages a supervised backbone (segment-everything-everywhere model \cite{seem}) to create segments. It is probable that we could reach even higher performance by using such extra supervision. Yet, we leave this research direction for future work as our scalable distillation strategy is already competitive enough for the purpose of our study.

Finally, we already notice that distilling features in WI backbones of increasing width helps both the linear probing and finetuning performance. We continue to study this property in the next section with WI, as this is the backbone which led to the best performance with our technique.

\subsection{2D Backbone Choice \& 3D Backbone Scale}
\label{sec:scaling_backbones}

\begin{table}[t]
\small
\centering
\setlength{\tabcolsep}{2.0pt}
\begin{tabular}{llrrrr}
\toprule
    & \multirow{2}{*}{\parbox{10.7mm}{Pretrain. Dataset}}
    & \multicolumn{4}{c}{Downstream \& Test Dataset}
\\
\cmidrule{3-6}
    &
    & nuScenes
    & SemKITTI
    & Pandar 64
    & Pandar GT
\\
\midrule
\multirow{3}{*}{\rotatebox[origin=t]{90}{WI-$\dimfeatthreed$}}
    & \multicolumn{5}{l}{\cellcolor{gray!20}\emph{No Pretraining - Best fully supervised WI}}
\\
    & \multirow{2}{*}{--}
    & 78.7
    & 65.1
    & 47.8
    & 40.6
\\
    &
    & \scriptsize $\dimfeatthreed$ = 768
    & \scriptsize $\dimfeatthreed$ = 256
    & \scriptsize $\dimfeatthreed$ = 768
    & \scriptsize $\dimfeatthreed$ = 256
\\
\midrule
\multirow{7}{*}{\rotatebox[origin=t]{90}{WI-256}}
    & \multicolumn{5}{l}{\cellcolor{gray!20}\emph{Pretraining with DINO-ViT-S/8 and linear probing}}
\\
    & nuScenes
    & \underline{54.4}
    & 28.8
    & 26.9
    & 25.2
\\
    & KITTI
    & 39.5
    & \underline{46.6}
    & 25.3
    & 25.7
\\
    & Pandar 64
    & 39.6
    & 25.6
    & \underline{30.0}
    & 24.7
\\
    & Pandar GT
    & 29.9
    & 26.9
    & 23.5
    & \underline{28.5}
\\
    & nuSc. \& KITTI
    & 54.4
    & 50.1
    & 29.3
    & 28.9
\\
    & All datasets
    & \bf54.6
    & \bf50.6
    & \bf33.1
    & \bf32.3
\\
\midrule
\multirow{3}{*}{\rotatebox[origin=t]{90}{WI-768}}
    & \multicolumn{5}{l}{\cellcolor{gray!20}\emph{Pretraining with DINOv2-ViT-L/14 and linear probing}}
\\
    & nuScenes
    & \underline{\bf67.8}
    & 43.1
    & 33.9
    & 29.9
\\
    & All datasets
    & \textbf{67.8}
    & \textbf{55.8}
    & \textbf{37.9}
    & \textbf{34.5}
\\
\bottomrule
\end{tabular}
\caption{\textbf{Benefit of pretraining on diverse datasets -- Linear probing}. Performance after linearly probing distilled features. The backbones are pretrained on nuScenes, SemanticKITTI, Pandar 64, Pandar GT, nuScenes \& SemanticKITTI, or all these datasets together, and then linearly probed on each individual dataset. The reported mIoU\% is computed on the val split of each dataset. The underlined \underline{mIoU}\% highlights the score obtained by pretraining and linear probing on the same dataset.}
\label{tab:multi-dataset}
\end{table}

In this section, we show that increasing the width of WI leads to substantial improvements of the feature quality. This effect combined with a good choice of 2D backbone significantly reduces the gap between the quality of distilled 3D features and fully-supervised 3D features.

We conduct experiments with three different WaffleIron backbones: WI-256, WI-384, WI-768. We pretrain these backbones on the complete training set of nuScenes or SemanticKITTI, and evaluate the quality of the distilled features on the validation set of the respective datasets. The evaluation consists of linear probings using 100\% of the annotated training scans. The results obtained by distilling MAE, DINO, and DINOv2 features are presented in \cref{tab:capacity}.

First, we observe on nuScenes that pretraining WI-256 backbones with ViTs pretrained with DINOv2 instead of DINO consistently improves the scores. We also notice that the linear probing performance improves when distilling DINOv2 ViTs of increasing capacities. This indicates that 3D backbones pretrained with \textbf{our method can directly benefit from an improved quality of the image features}. Interestingly, the MAE-pretrained ViT leads to the worst performance. We attribute this to the fact that MAE features are less linearly separable, as reported in~\cite{he2022masked}.

Second, we notice that \textbf{the quality of our distilled features increases significantly with the WI width} for ViT-S/8 and ViT-L/14 on nuScenes and SemanticKITTI. Noticeably, we reach a mIoU of $66.8$ by linear probing on nuScenes. This is only about $11.9$ points away from the mIoU reached by WI-768 when trained under full supervision. Note that WI-768 is actually well-performing on this dataset as it reaches a mIoU of $78.7\%$ while the best published score, without test time augmentations, we know of on this dataset is $78.4\%$ (reported in \cite{spheretransformer}). It shows that \textbf{distilling 2D features without manual annotations can bridge the gap with the quality of fully supervised 3D features}.

\subsection{Pretraining on Multiple Datasets}
\label{sec:multi_datasets}

\begin{table}[t]
\small
\centering
\setlength{\tabcolsep}{3.2pt}
\begin{tabular}{lccccccc}
\toprule
\multirow{2}{*}{\parbox{10.7mm}{Pretrain. Dataset}}
    & \multicolumn{3}{c}{nuScenes}
    & \multicolumn{2}{c}{SemKITTI}
    & \multicolumn{1}{c}{Pan. 64}
    & \multicolumn{1}{c}{Pan. GT}
\\
\cmidrule(lr){2-4}
\cmidrule(lr){5-6}
\cmidrule(lr){7-7}
\cmidrule(lr){8-8}
    & 1\%
    & 10\%
    & 100\%
    & 1\%
    & 100\%
    & 100\%
    & 100\%
\\
\midrule
\multicolumn{8}{l}{\cellcolor{gray!20}\emph{No Pretraining - Fully supervised WI-768}}
\\
- 
    & 35.2
    & 62.2
    & \textbf{78.7}
    & 49.6
    & 63.4
    & 47.8
    & 40.0
\\
\midrule
\multicolumn{8}{l}{\cellcolor{gray!20}\emph{Pretraining WI-768 with DINOv2-ViT-L/14 and finetuning}}
\\
nuScenes
    & \textbf{51.0}
    & \textbf{70.5}
    & 77.9
    & 49.7
    & 63.9
    & \textbf{49.4}
    & \textbf{42.1}
\\
All
    & 50.7
    & 69.2
    & 78.4
    & \textbf{56.8}
    & \textbf{65.8}
    & 48.3
    & 41.1
\\
\bottomrule
\end{tabular}
\caption{\textbf{Finetuning performance}. The backbones are pretrained on nuScenes alone or combined with SemanticKITTI, Pandar 64, Pandar GT (`All'), and finetuned on the train split of each dataset for different amounts of data. The reported mIoU\% is computed on the val split of each dataset.}
\label{tab:multi-dataset-finetuning}
\end{table}

In this section, we show that we can pretrain a single backbone on multiple datasets that performs as well or better than individual backbones specialized to each dataset.

We conduct our main experiment with WI-256 and the DINO-pretrained ViT-S/8. The datasets we consider (nuScenes, SemanticKITTI, Pandar 64 and Pandar GT) have different sizes. To avoid drawing conclusions which could be explained by different amount of pretraining, we adjust the number of pretraining epochs on each dataset so that the backbones are pretrained for the same number of iterations (the batch size is fixed to 16). 

\begin{table}[t]
\small
\centering
\setlength{\tabcolsep}{2pt}
\begin{tabular}{lllccc}
\toprule
Method
    & 3D Back.
    & \parbox{10mm}{Pretrain dataset}
    & mCE\% \scriptsize$\downarrow$
    & mRR\% \scriptsize$\uparrow$
    & \parbox{10mm}{\centering Avg. mIoU\%}
\\
\midrule
--
    & Cylind3D~\cite{cylinder3d}
    & --
    & \it105.6
    & \it78.1  
    & \it57.4
\\
--
    & 2DPASS~~\,\cite{2dpass}
    & --
    & \it\phantom{1}98.6
    & \it75.2   
    & \it58.6
\\
--
    & SPVCNN~\cite{spvnas}
    & --
    & \it\phantom{1}97.5 
    & \it75.1  
    & \it57.5
\\
--
    & GFNet~~~~~~\cite{gfnet}
    & -- 
    & \it\phantom{1}92.6
    & \it83.3
    & \it64.0
\\
--
    & WI-768
    & --
    & \phantom{1}90.9
    & 80.6
    & 63.5
\\
\midrule
PPKT
    & MinkUNet
    & nuScenes
    & \it105.6
    & \it76.1
    & \it56.6
\\
SLidR
    & MinkUNet
    & nuScenes
    & \it106.1
    & \it76.0
    & \it56.8
\\
Seal
    & MinkUNet
    & nuScenes
    & \it\phantom{1}92.6
    & \it83.1
    & \it62.8
\\
\midrule
\bf\ours
    & WI-768
    & nuScenes
    & \phantom{1}89.1
    & 83.7
    & 65.2
\\
\bf\ours
    & WI-768
    & Multiple
    & \bf\phantom{1}87.4
    & \bf83.8
    & \bf65.7
\\
\bottomrule
\end{tabular}
\caption{\textbf{Robustness to corruptions}. The evaluation is done on nuScenes-C from the Robo3D benchmark \cite{robo3d}. We report the mCE\%, mRR\%, and the mIoU\% averaged over all eight corruptions. The scores in italic are obtained from \cite{liu2023segment,robo3d}. PPKT, SLidR and Seal use a MoCov2 ResNet-50. We use DINOv2 ViT-L/14.}
\label{tab:robo3d}
\end{table}

We start by pretraining WI-256 on each dataset individually and evaluate the quality of the distilled features by linear probing on each of the datasets. The results are presented in \cref{tab:multi-dataset}. We notice that the pretrained backbones are quite sensitive to the domain gap induced by different lidars. For example, the backbone pretrained on nuScenes performs significantly worse on SemanticKITTI than the backbone pretrained on SemanticKITTI: 28.8\% vs 46.6\% in mIoU. The observation is valid whatever pairs of pretraining/downstream datasets we take. However, by pretraining a backbone on both nuScenes and SemanticKITTI, we do not notice any drop in linear probing performance on nuScenes and achieve even better results on SemanticKITTI. The best overall performance is obtained when pretraining on all datasets together. This experiment shows that \textbf{a single backbone easily benefits from diverse lidar data without performance loss in linear probing}.

We conduct one more experiment using the best combination of 3D and 2D backbone discovered in the last section: WI-768 and DINOv2 ViT-L/14. We pretrain this backbone either on nuScenes alone or the combination of all considered datasets. Due to memory constraints at this scale when training on all datasets, we reduce the batch size to 8 for both pretrainings. We thus also used a longer training schedule. The results are presented in the last section of \cref{tab:multi-dataset}. We notice again that \textbf{pretraining on multiple datasets leads to better linear probing performance across all downstream datasets}. We also remark that reducing the batch size and training for longer allowed to gain 1 mIoU point when training only on nuScenes (see WI-768 in \cref{tab:capacity} and \cref{tab:multi-dataset}). The remaining gap with respect to the best fully-supervised WI baseline is less than 10.9 mIoU points on all datasets. Notably, on nuScenes, we reach 67.8 in mIoU while the best reported score so far in linear probing was 45.0 (see \cref{tab:all_results}), reached by \cite{liu2023segment} with the help of fully supervised 2D backbones to create semantically coherent segments.

\subsection{Properties of Distilled Features}
\label{sec:properties}

In this section, we use the WI-768 backbones pretrained with \ours using DINOv2 ViT-L/14. We consider two pretraining datasets: (1) nuScenes alone and (2) the combination of nuScenes, SemanticKITTI, Pandar 64/GT (denoted by `Multiple' in pretraining dataset column of the tables).

\smallparagraph{Performance in finetuning.} We present in \cref{tab:multi-dataset-finetuning} the results obtained by finetuning our pretrained WI-768 on different datasets for different amounts of data. First, we notice that \textbf{pretraining always improves the score compared to the non-pretrained fully supervised baseline}, except on nuScenes when using all the available annotations. We also notice that WI-768 pretrained on multiple datasets: (1) significantly outperforms the nuScenes-pretrained WI-768 when finetuning on SemanticKITTI; (2) performs nearly as well on nuScenes as the model nuScenes-pretrained WI-768, i.e., the model specialized to this dataset. Interestingly, WI-768 pretrained on nuScenes seems to have a small advantage for finetuning on Pandar. We recall nevertheless that the model pretrained on multiple datasets still improves the performance compared to no pretraining. More importantly, as will be seen in the next experiment, \textbf{multi-dataset pretraining produces more robust backbones against perturbations and domain gaps}.

\begin{table}[t]
\begin{minipage}[t]{0.24\linewidth}
\small
\setlength{\tabcolsep}{2.5pt}
\vspace*{-3.2mm}
\begin{tabular}{lcc}
\toprule
\parbox{10mm}{Pretrain Dataset}
    & KITTI
    & POSS
\\
\midrule
--
    & 28.6
    & 54.0
\\
nuScenes
    & 29.6
    & 57.1
\\
Multiple
    & 36.6
    & 59.1
\\
\bottomrule
\end{tabular}
\end{minipage}
\hspace{13mm}
\begin{minipage}[t]{0.24\linewidth}
\small
\centering
\setlength{\tabcolsep}{2.5pt}
\begin{tabular}{lccc}
\toprule
\multirow{2}{*}{\parbox{10mm}{3D Back}}
    & \multicolumn{2}{c}{2D Back}
    & \multirow{2}{*}{LP}
\\
\cmidrule{2-3}
    & MaskCLIP
    & DINOv2
\\
\midrule
WI-256
    & \cmark
    & \xmark
    & 59.4
\\
WI-256
    & \xmark
    & \cmark
    & 61.6
\\
WI-256
    & \cmark
    & \cmark
    & 62.5
\\
\midrule
WI-384
    & \cmark
    & \xmark
    & 60.9
\\
WI-384
    & \xmark
    & \cmark
    & 63.9
\\
WI-384
    & \cmark
    & \cmark
    & 65.1
\\
\bottomrule
\end{tabular}
\end{minipage}
\caption{\textbf{Domain generalization properties (left) and benefit of pretraining with multiple teachers (right)}. \emph{Left}: mIoU\% on SemanticKITTI (abbreviated KITTI) and SemanticPOSS (abbreviated POSS) of WI-768 pretrained either on nuScenes alone or on all considered pretraining datasets, and then finetuned on nuScenes. The performance of WI-768 trained under full supervision on nuScenes without pretraining is also reported. \emph{Right}: Linear probing mIoU\% obtained on nuScenes after pretraining WI with \ours on nuScenes. The considered 2D backbones are DINOv2 ViT-L/14 and MaskCLIP ViT-B/16.}
\label{tab:domain_generalization}
\end{table}

\smallparagraph{Robustness to corruptions.} We evaluate the robustness of the pretrained WI-768 to corruptions on the Robo3D benchmark \cite{robo3d}. The results are presented in \cref{tab:robo3d}. We notice that WI-768 trained under full supervision on nuScenes, without any pretraining, already performs well compared to the other 3D backbones trained in the same condition. WI-768 achieves the best results in terms of mean corruption error (mCE) and is second in terms of mean resilience rate (mRR). Then, we remark that pretraining always improves the robustness of WI-768. Finally, the best performance is achieved thanks to our multiple datasets pretraining with an mCE of 87.4\% and a mRR of 83.8\%, i.e., new state-of-the-art results on this benchmark.

\smallparagraph{Domain generalization properties.} We evaluate the capacity of our pretrained WI-768 to generalize to different lidars. The backbones are finetuned on nuScenes after pretraining. We measure the performance of the resulting models directly on the val set of SemanticKITTI (train set seen during pretraining) or SemanticPOSS (not seen during pretraining) using respectively 10 and 6 aggregated classes, as done, e.g., \cite{completeandlabel,saluda}. We notice in \cref{tab:domain_generalization} that pretraining always helps generalization and that the best performance are obtained when pretraining on multiple datasets.

\subsection{Pretraining with multiple teachers}
\label{sec:multiple_teachers}

We show in \cref{tab:domain_generalization} that using multiple 2D teachers 
is a way to further improve the performance. 
Distillation was done using the two modifications in \ours: (1) On the 2D side, we concatenate the pixel features obtained from both backbones; (2) On the 3D side, we use a 2-layer MLP for $\headthreed$ with an output size matching the dimension of the concatenated pixel features. More details about this MLP are available in the supp.~material.

\section{Conclusion}

\begin{table}[t]
\small
\centering
\setlength{\tabcolsep}{2.2pt}
\begin{tabular}{lllccccccccc}
\toprule
\multirow{2}{*}{Method}
    & \multirow{2}{*}{3D}
    & \multirow{2}{*}{Pretrain}
    & \multicolumn{4}{c}{nuScenes}
    & \multicolumn{1}{c}{KITTI}
\\
\cmidrule(lr){4-7}
\cmidrule(lr){8-8}
    &
    &
    & LP
    & 1\%
    & 10\%
    & 100\%
    & 1\%
\\
\midrule
--
    & MinkUNet
    & --
    & \it~8.1
    & \it30.3
    & \it56.2
    & \it74.7
    & \it39.5
\\
--
    & WI-768
    & --
    & --
    & 35.2
    & 62.2
    & \textbf{78.7}
    & 49.6
\\
\midrule
\midrule
PPKT
    & MinkUNet
    & nuScenes
    & \it35.9
    & \it37.8
    & \it60.3
    & \it74.5
    & \it44.0
\\
SLidR
    & MinkUNet
    & nuScenes
    & \it38.8
    & \it38.3
    & \it59.8
    & \it74.8
    & \it44.6
\\
ST-SLidR
    & MinkUNet
    & nuScenes
    & \it40.5
    & \it40.8
    & \it60.8
    & \it75.1
    & \it44.7
\\
Seal
    & MinkUNet
    & nuScenes
    & \it45.0
    & \it45.8
    & \it63.0
    & \it75.6
    & \it46.6
\\
\midrule
\bf\ours
    & WI-768
    & Multiple
    & \bf67.8
    & \bf50.7
    & \bf69.2
    & \underline{78.4}
    & \bf56.8
\\
\bottomrule
\end{tabular}
\caption{\textbf{Progress made with \ours}. We report the mIoU\% obtained by linear probing (LP) or finetuning on different amounts of data. The scores in italic are reported from \cite{liu2023segment}. WI-768 is pretrained using DINOv2 ViT-L/14 on nuScenes, SemanticKITTI (abbreviated KITTI), Pandar 64, Pandar GT. The MinkUNet backbones are pretrained using MoCov2 ResNet-50 on nuScenes.}
\label{tab:all_results}
\end{table}

We have seen that scaling 2D and 3D backbones and pretraining on multiple datasets lead to better and more robust distilled features. We summarize in \cref{tab:all_results} the scores obtained by \cite{ppkt,sautier22SLidR,mahmoud2023st_SLidR,liu2023segment} and with \ours to highlight the significant progress we have made. The most notable improvement appears when linear probing and finetuning on small amounts of data. Note that we \emph{do not} claim that the concurrent distillation methods cannot surpass our proposed scalable distillation method; the take-away message is that scaling the 2D and 3D backbones and pretraining on diverse datasets actually lead to significantly better pretrained backbones even with a simple distillation method. These are three pillars that should not be overlooked to get good pretrained backbones for lidars. We hope that these findings, combined with more powerful distillation methods, will lead to even better 3D backbones in the future.

\smallparagraph{Acknowledgement} This project has benefited from an access to the HPC resources of CINES under the allocation GDA2213 for the Grand Challenges AdAstra GPU made by GENCI. We thank Mathieu Cloirec and Jean-Christophe Penalva at CINES for helping us accessing the cluster and creating the good setup to run our experiments smoothly. This research also received the support of EXA4MIND project, funded by a European Union´s Horizon Europe Research and Innovation Programme, under Grant Agreement N° 101092944. Views and opinions expressed are however those of the author(s) only and do not necessarily reflect those of the European Union or the European Commission. Neither the European Union nor the granting authority can be held responsible for them. We acknowledge EuroHPC Joint Undertaking for awarding us access to Karolina at IT4Innovations, Czech Republic. We also acknowledge the support of the French Agence Nationale de la Recherche (ANR), under grant ANR-21-CE23-0032 (project MultiTrans).

{
\small
\bibliographystyle{ieeenat_fullname}
\bibliography{main}
}

\maketitlesupplementary
\appendix
\section{Robustness}

We present in \cref{tab:robo3d_detailed} a detailed version of Tab.~\textcolor{red}{6} with the mIoU\% attained by for the eight different types of corruptions considered in \cite{robo3d}. 

We observe that WI-768 pretrained on multiple datasets is second or third for each type of corruption, except for motion blur, which permits it to achieve the best mCE\% and mRR\%.

\begin{table*}[t]
\small
\centering
\setlength{\tabcolsep}{3.5pt}
\scalebox{0.98}{
\begin{tabular}{lrllllcccccccccc}
\toprule
    \multirow{2}{*}{Method}
    &
    & \multirow{2}{*}{2D Back.}
    & \multirow{2}{*}{3D Back.} 
    & \multirow{2}{*}{\parbox{10.7mm}{Pretrain. dataset}}
    & \multirow{2}{*}{mCE\% $\downarrow$}
    & \multirow{2}{*}{mRR\% $\uparrow$}
    & \multicolumn{8}{c}{Corruptions (mIoU\% $\uparrow$)}
\\
\cmidrule(lr){8-15}
    &
    &
    &
    &
    &
    &
    & Fog 
    & Wet
    & Snow
    & \!\!Motion\!\!
    & Beam
    & Cross
    & Echo
    & Sensor
\\
\midrule
    --
    &
    & --
    & Cylinder3D \cite{cylinder3d}
    & --
    & \it105.6
    & \it78.1
    & \it61.4
    & \it71.0
    & \it58.4
    & \it56.0
    & \it64.2
    & \it45.4
    & \it60.0
    & \it43.0
\\
    --
    &
    & --
    & 2DPASS~~~~~\,\cite{2dpass}
    & --
    & \it\phantom{1}98.6
    & \it75.2 
    & \it64.5 
    & \it76.8 
    & \it54.5
    & \it\cellcolor{myblue!15}62.0 
    & \it67.8 
    & \it34.4
    & \it63.2
    & \it45.8
\\
    --
    &
    & --
    & SPVCNN~~~~\cite{spvnas}
    & --
    & \phantom{1}97.5 
    & \it75.1 
    & \it55.9
    & \it74.0
    & \it42.0
    & \it\cellcolor{myblue!85}74.6 
    & \it69.0
    & \it28.1 
    & \it\cellcolor{myblue!85}65.0
    & \it\cellcolor{myblue!15}51.6
\\
    --
    &
    & --
    & GFNet~~~~~~~~~\cite{gfnet}
    & -- 
    & \it\phantom{1}92.6
    & \it\cellcolor{myblue!15}83.3
    & \it69.6
    & \it75.5
    & \it\cellcolor{myblue!85}71.8
    & \it59.4
    & \it64.5
    & \it\cellcolor{myblue!85}66.8
    & \it61.9 
    & \it42.3
\\
    --
    &
    & --
    & WI-768
    & --
    & \cellcolor{myblue!15}\phantom{1}90.9
    & 80.6
    & \cellcolor{myblue!50}72.2
    & \cellcolor{myblue!85}78.0
    & 66.6
    & 55.2
    & \cellcolor{myblue!85}70.4
    & 48.7
    & \cellcolor{myblue!50}64.7
    & \cellcolor{myblue!85}52.4
\\
\cmidrule{1-15}
    PPKT
    & \cite{ppkt}
    & ResNet-50
    & MinkUNet
    & nuScenes
    & \it105.6
    & \it76.1
    & \it64.0
    & \it72.2
    & \it59.1
    & \it57.2
    & \it63.9
    & \it36.3
    & \it60.6
    & \it39.6
\\
    SLidR
    & \cite{sautier22SLidR}
    & ResNet-50
    & MinkUNet
    & nuScenes
    & \it106.1
    & \it76.0
    & \it65.4
    & \it72.3
    & \it56.0
    & \it56.1
    & \it62.9
    & \it41.9
    & \it61.2
    & \it38.9
\\
    Seal
    & \cite{liu2023segment}
    & ResNet-50
    & MinkUNet
    & nuScenes
    & \it\phantom{1}92.6
    & \it83.1
    & \it\cellcolor{myblue!85} 72.7
    & \it74.3
    & \it66.2
    & \it\cellcolor{myblue!50}66.1
    & \it66.0
    & \it57.4
    & \it59.9
    & \it39.9
\\
\cmidrule{1-15}
    \ours
    & \llap{(ours)}
    & ViT-L/14
    & WI-768
    & nuScenes
    & \cellcolor{myblue!50}\phantom{1}89.1 
    & \cellcolor{myblue!50}83.7
    & 70.8
    & \cellcolor{myblue!15}77.2
    & \cellcolor{myblue!15}67.1
    & 55.9
    & \cellcolor{myblue!15}70.0
    & \cellcolor{myblue!50}65.7
    & 63.9
    & 51.1
\\
    \ours
    & \llap{(ours)}
    & ViT-L/14
    & WI-768
    & Multiple
    & \cellcolor{myblue!85}\phantom{1}87.4
    & \cellcolor{myblue!85}83.8
    & \cellcolor{myblue!50}72.2
    & \cellcolor{myblue!50}77.9
    & \cellcolor{myblue!50}69.1
    & 57.4
    & \cellcolor{myblue!50}70.1
    & \cellcolor{myblue!15}62.7
    & \cellcolor{myblue!15}64.0
    & \cellcolor{myblue!50}52.2
\\
\bottomrule
\end{tabular}}
\caption{\textbf{Robustness to corruptions}. The evaluation is conducted on nuScenes-C from the Robo3D benchmark \cite{robo3d}. We report the mCE\%, mRR\%, and the mIoU\% attained for the eight corruptions, i.e., fog, wet ground, snow, motion blur, beam missing, crosstalk (among multiple sensors), incomplete echo, and cross-sensor (beam and point dropping). The scores in italic are obtained from \cite{liu2023segment,robo3d}. PPKT, SLidR and Seal use a MoCov2 ResNet-50. We use DINOv2 ViT-L/14.}
\label{tab:robo3d_detailed}
\end{table*}

\section{Training details}
\label{sec:training_details}

\subsection{Pandaset} 

Pandaset \cite{pandaset} is made of scans acquired in San Francisco and along El Camino Real from Palo Alto to San Mateo. We use the scans collected in San Francisco as train set and the rest as validation set. We also separate the scans collected with the Pandar64 from the scans collected with the PandarGT and treat them as different datasets. Finally, we merge the fine-grained original classes into 17 classes similar to those used in nuScenes and SemanticKITTI. These classes are: road, traffic sign, barrier, pedestrian, vegetation, road marking, sidewalk, manmade, traffic cone, car, motorcycle, truck, bus, bicycle, other vehicle, ground, and driveway.

\subsection{Pretraining} 

\smallparagraph{Shared setting.} The point tokens (see \cite{puy23waffleiron} for details) in the WaffleIron backbones are computed using 16 nearest neighbors and the following point features: lidar intensity, 3D Cartesian $xyz$ coordinates, radius/range. The field-of-view in the spatial mixing blocks is restricted to $[-64\;{\rm m}, + 64\;{\rm m}]$ along the $x$ and $y$ axes and $[-8\;{\rm m}, +8\;{\rm m}]$ along the $z$ axis; we use a grid of resolution 50 cm. 

We pretrain the WaffleIron backbones using AdamW \cite{adamw} with a weight decay of $3 \times 10^{-4}$ and a learning rate linearly increasing from 0 to 0.002, then decreasing to $10^{-5}$ following a cosine schedule. The number of iterations or epochs corresponding to that maximum learning rate, as well as the total number of iterations or epochs, are described below, depending on the setting.

\smallparagraph{Mono-dataset setting.} The mono-dataset setting is the setting used in Secs.~\textcolor{red}{3.3}, \textcolor{red}{4.3}, \textcolor{red}{4.4}, \textcolor{red}{4.7}. We pretrain the WaffleIron backbones by distilling 2D features during 19 epochs, with a batch size of 16. The learning rate reaches its maximum value after 2 epochs. The MinkUNet backbone is pretrained following SLidR \cite{sautier22SLidR} protocol with the following adjustments: we use a batch size of 12, an initial learning rate of 1.5 --the optimizer used in SLidR is SGD-- and 25 epochs.

\smallparagraph{Multi-dataset setting.} The multi-dataset setting is the setting used in Secs.~\textcolor{red}{4.5}, \textcolor{red}{4.6}. The pretraining dataset contains: 28,130 scans for nuScenes, 19,130 scans for SemanticKITTI,  3,920 scans for Panda64 and 3,920 scans for PandaGT. We pretrain WI-256 by distilling 2D features with a batch size of 16 during 19 epochs on nuScenes, 28 epochs on SemanticKITTI, 136 epochs on Panda 64 or Panda GT, 11 epochs on nuScenes \& SemanticKITTI, 10 epochs on the mix of all datasets. The number of epochs is adjusted so that the backbone is pretrained for (approximately) the same number of iterations. For WI-768, as the available GPU memory is not sufficient for some batches, we decrease the batch size to 8 and pretrain for 49 epochs on nuScenes and 25 epochs on the mix of all datasets. The learning rate reaches its maximum value after 3500 iterations in all cases.

\smallparagraph{2D teacher.} All the results presented in this paper with the DINO-pretrained ViT-S/8 are obtained by distilling the keys at the last attention layer. This choice was guided by the fact that the last keys have properties that enable the design of unsupervised object discovery algorithms \cite{simeoni21lost,wang2022tokencut,simeoni2023unsupervised}. The results obtained with the DINOv2-pretrained ViTs are obtained by distilling the features before the last normalization layer.

\smallparagraph{Multi-teacher distillation.} Let us denote the output feature dimension of both image teachers by $\dimfeattwod^{(1)}$ and $\dimfeattwod^{(2)}$, respectively. On the image side, we $\ell_2$-normalize the pixel features extracted by each teacher and concatenate them. On the point cloud side, the head $\headthreed$ is a 2-layer MLP where the hidden linear layer has size $2 \times \dimfeatthreed$ and is followed by a Layer Norm and a \texttt{ReLU}. The final linear layer of the MLP has size $\dimfeattwod^{(1)} + \dimfeattwod^{(2)}$ to match the size of the concatenated 2D features. These point features are then split into two parts of size $\dimfeattwod^{(1)}$ and $\dimfeattwod^{(2)}$, respectively. Each part is $\ell_2$-normalized independently. The normalized features are then re-concatenated. Finally, we distill the knowledge of the 2D features by applying Eq.~(\textcolor{red}{1}) directly on the features of size $\dimfeattwod^{(1)} + \dimfeattwod^{(2)}$.

\subsection{Linear probing} 

The linear head is trained with a batch size of 8, using AdamW with a weight decay of $3 \times 10^{-3}$. The learning rate linearly increases from 0 to 0.001 during the first 2 epochs and then decreases to $10^{-5}$ following a cosine schedule. We use 20 epochs on nuScenes and SemanticKITTI, and 50 epochs an Pandar64 and PandarGT.

\subsection{Finetuning} 

For finetuning the pretrained WaffleIron backbones, we use a batch size of $8$, using AdamW without weight decay. The learning rate linearly increases from $0$ to $0.002$ during the first tenth of epochs and then decreases to $0$ following a cosine schedule. During finetuning, we also use stochastic depth~\cite{stocdepth} with a layer drop probability of 0.2. We finetune the WaffleIron backbones for 45 epochs and a layer-wise learning rate decay parameter of $0.95$ when using 1\% and 10\% of available data, and for 25 epochs and a layer decay parameter of $0.99$ when all annotated data are available.

\section{Visual inspection}

\begin{figure*}
\small
\centering
    \begin{minipage}{3mm}
    \rotatebox[origin=c]{90}{nuScenes}
    \end{minipage}
    \begin{minipage}{0.24\linewidth}
    \includegraphics[width=\linewidth]{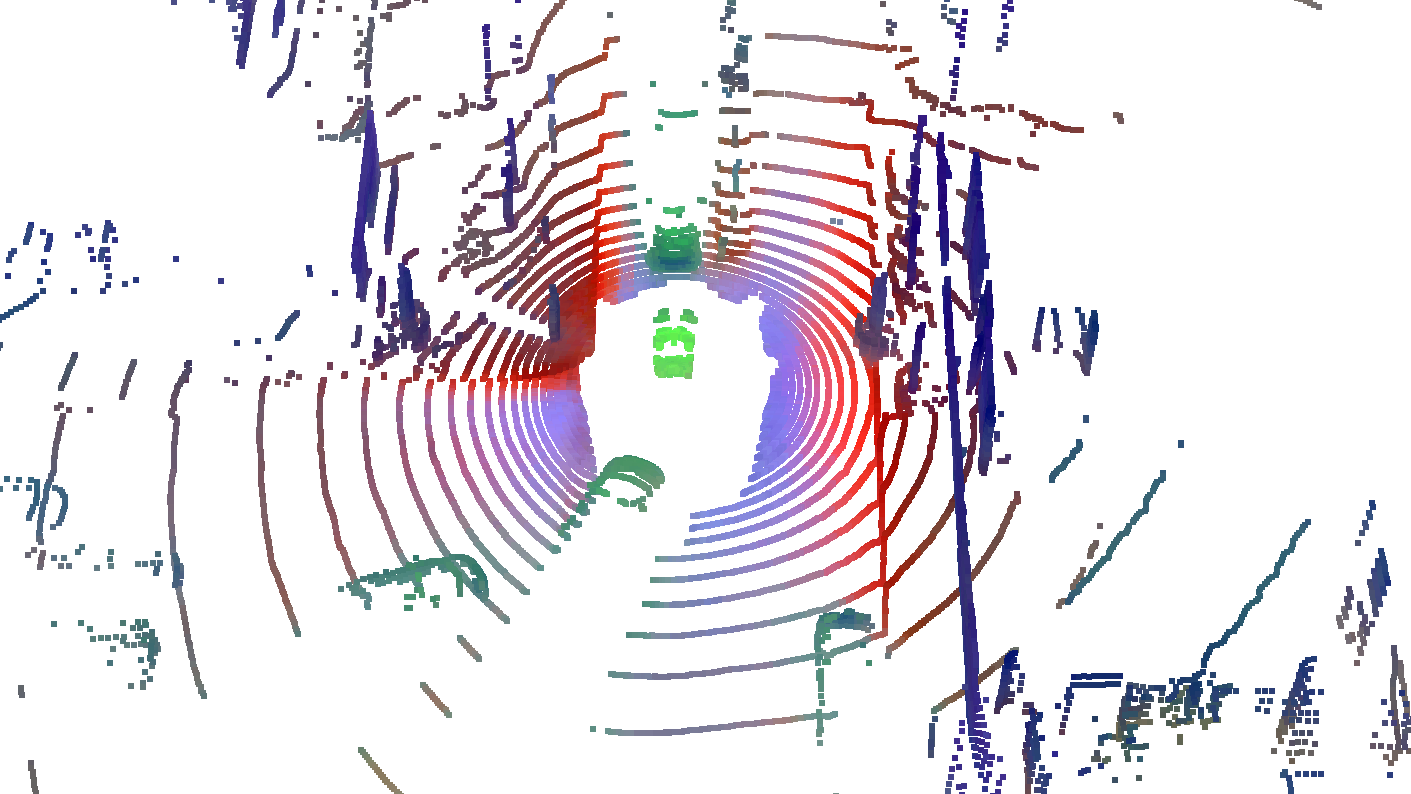}    
    \end{minipage}
    \begin{minipage}{0.24\linewidth}
    \includegraphics[width=\linewidth]{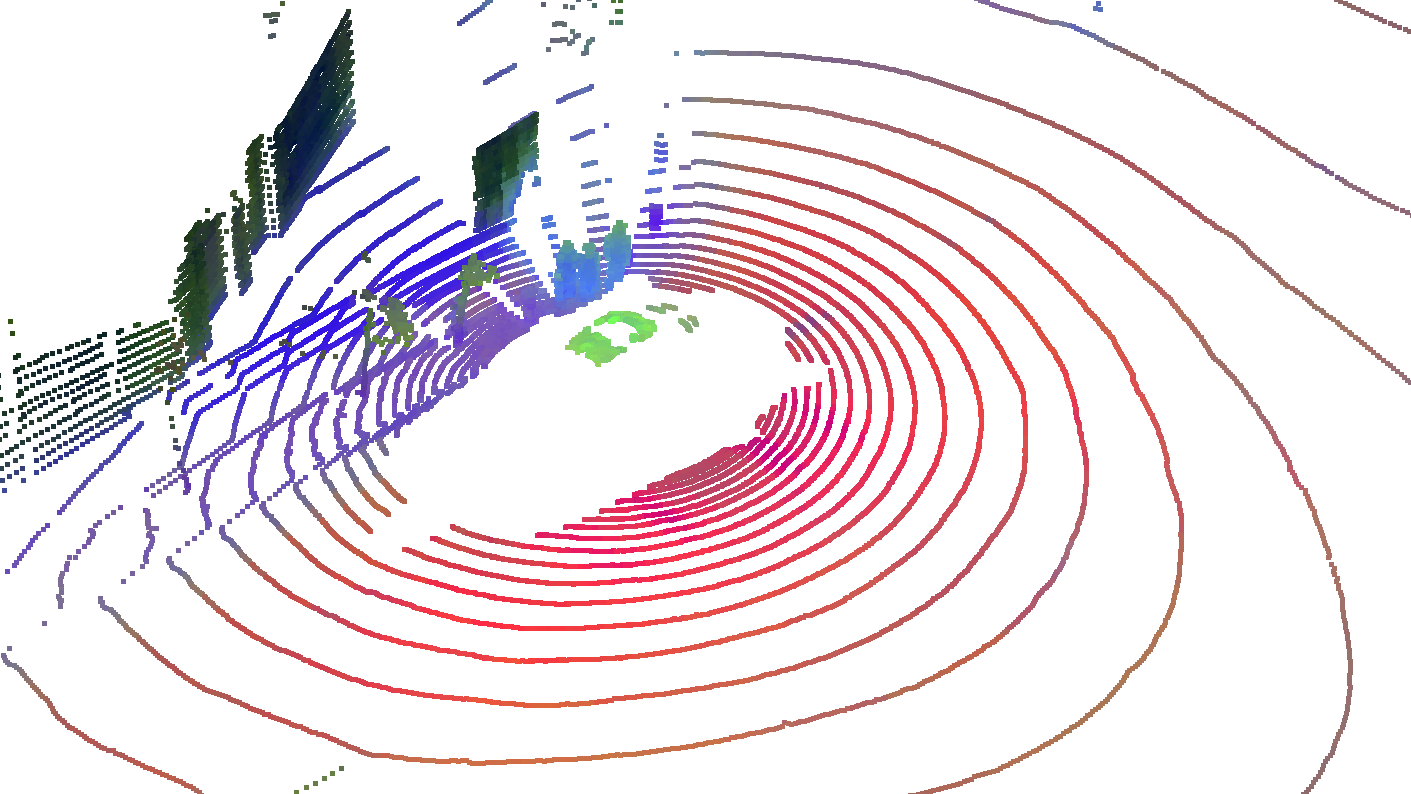}    
    \end{minipage}
    \begin{minipage}{0.24\linewidth}
    \includegraphics[width=\linewidth]{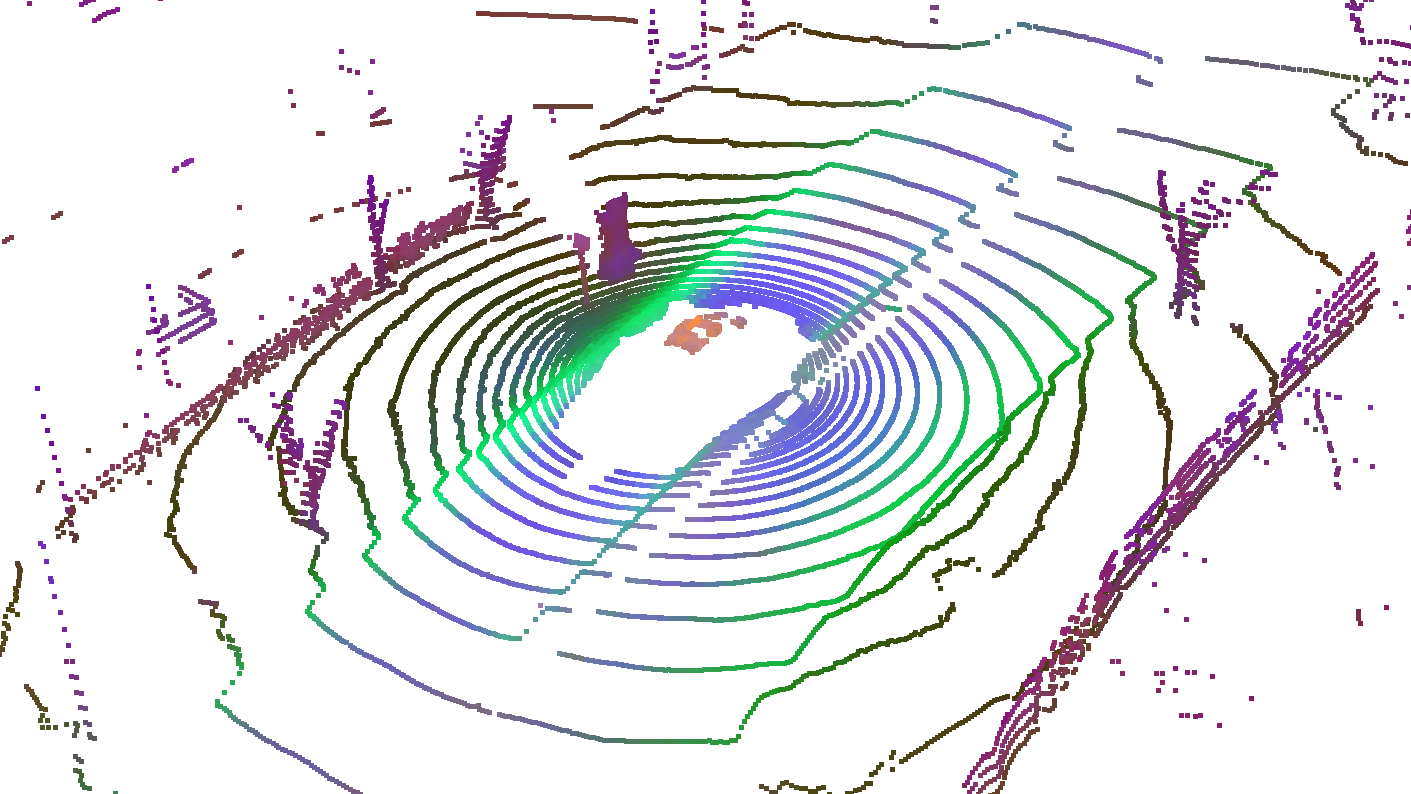}    
    \end{minipage}
    \begin{minipage}{0.24\linewidth}
    \includegraphics[width=\linewidth]{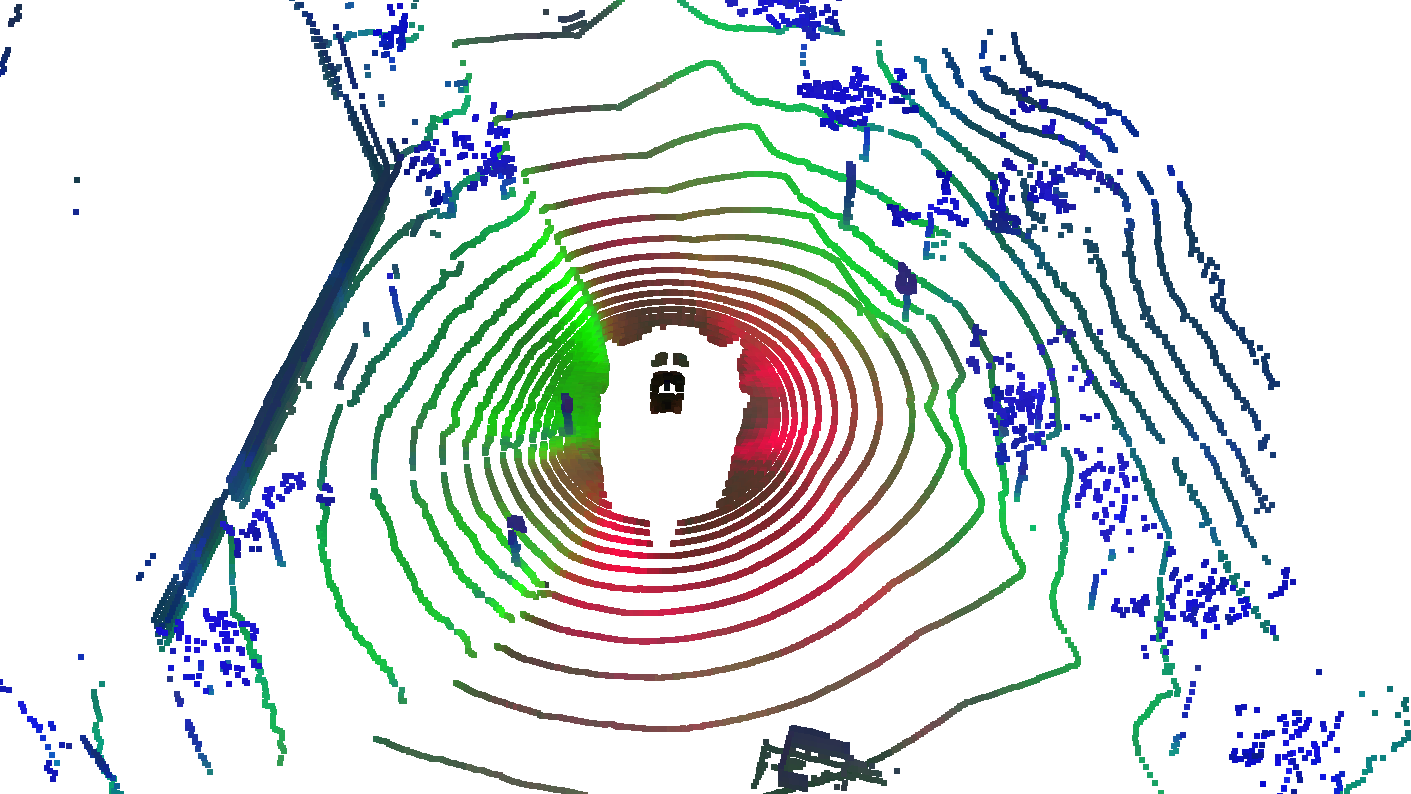}    
    \end{minipage}
    \\
    \begin{minipage}{3mm}
    \rotatebox[origin=c]{90}{SemanticKITTI}
    \end{minipage}
    \begin{minipage}{0.24\linewidth}
    \includegraphics[width=\linewidth]{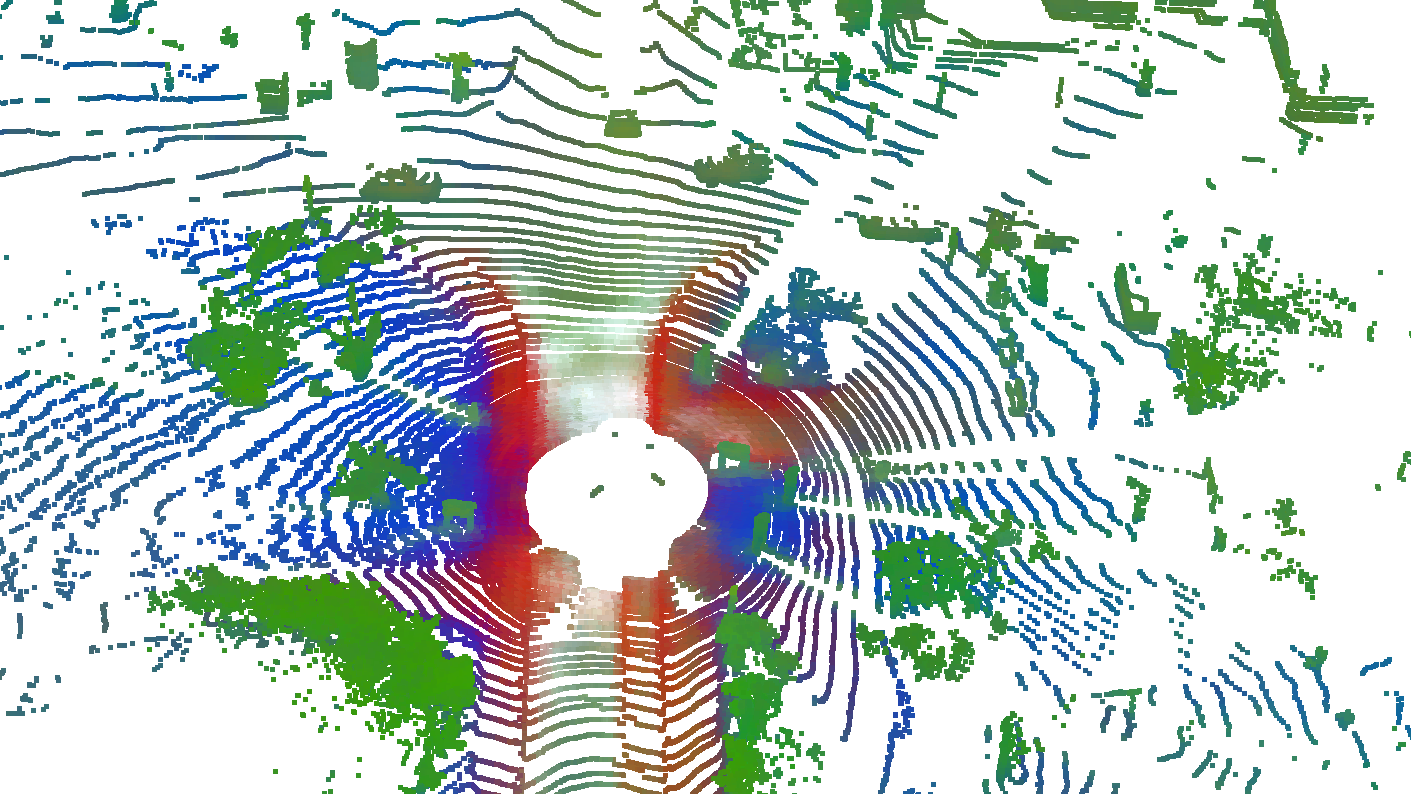}    
    \end{minipage}
    \begin{minipage}{0.24\linewidth}
    \includegraphics[width=\linewidth]{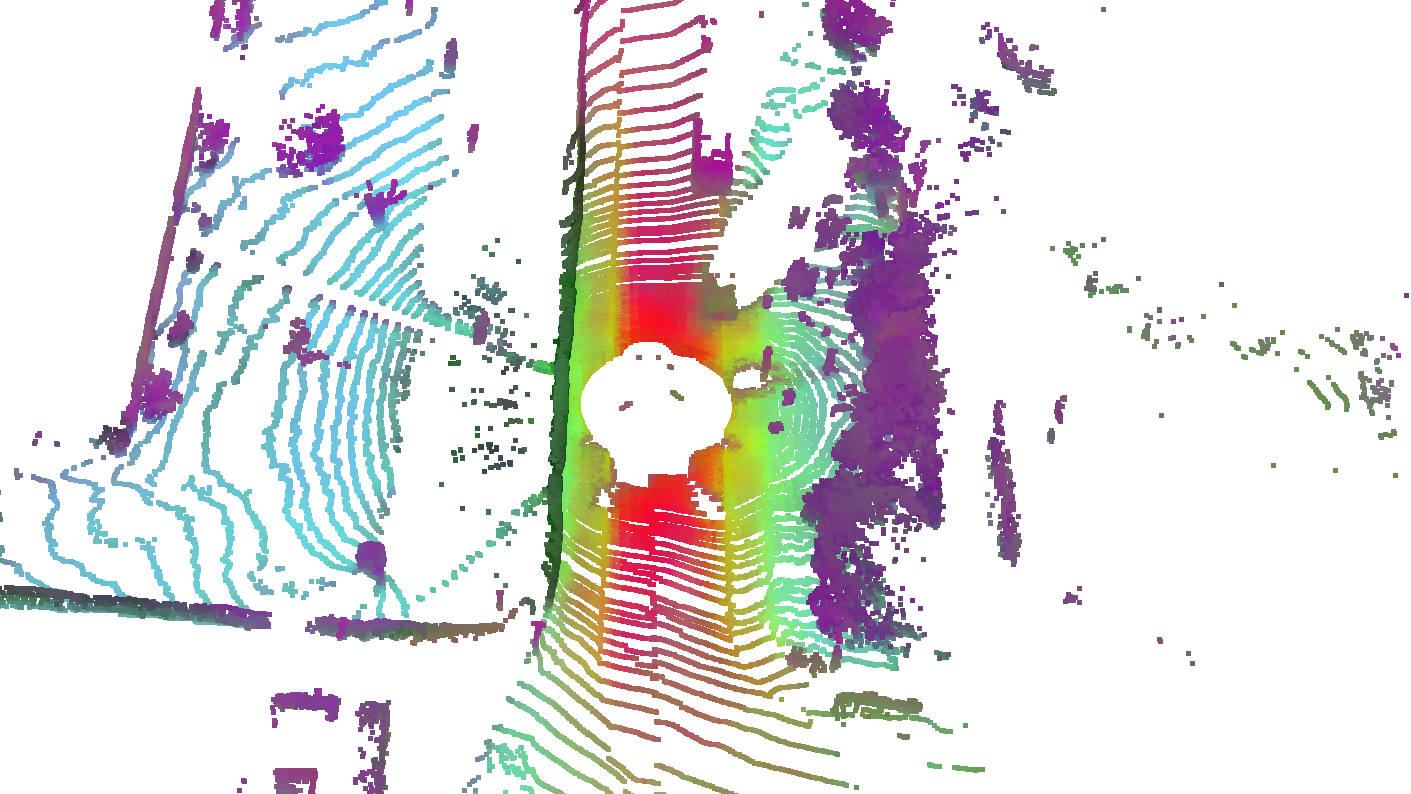}    
    \end{minipage}
    \begin{minipage}{0.24\linewidth}
    \includegraphics[width=\linewidth]{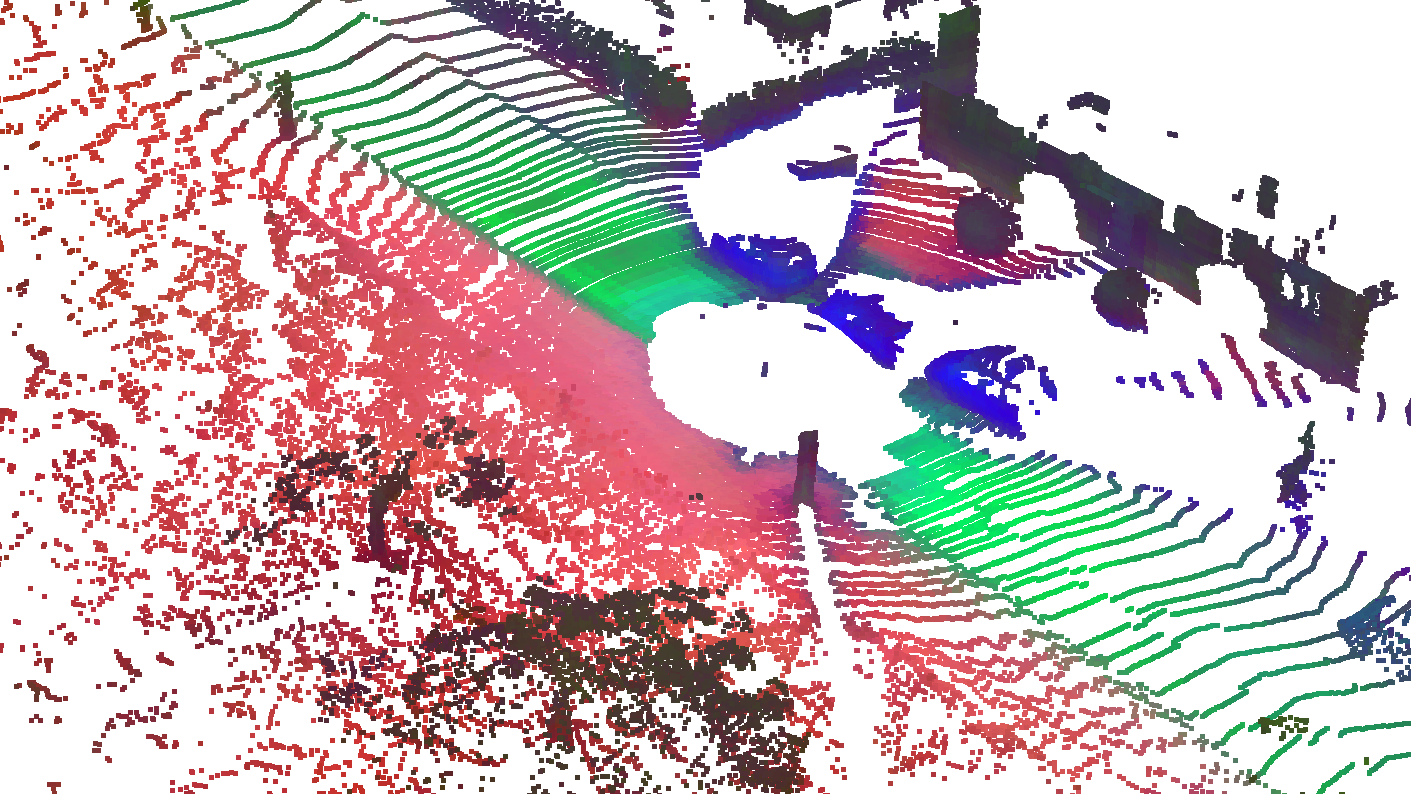}    
    \end{minipage}
    \begin{minipage}{0.24\linewidth}
    \includegraphics[width=\linewidth]{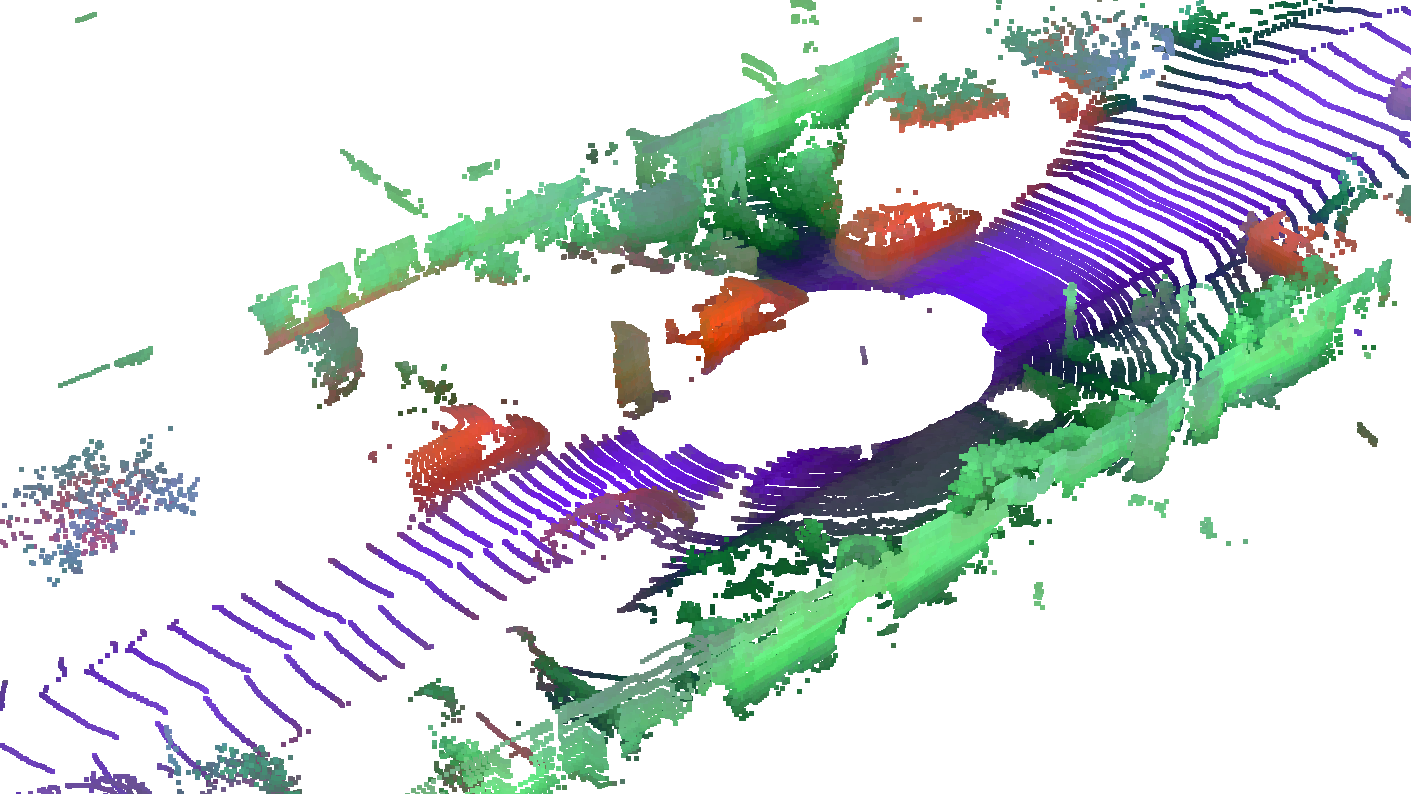}    
    \end{minipage}
    \\
    \begin{minipage}{3mm}
    \rotatebox[origin=c]{90}{Pandar64}
    \end{minipage}
    \begin{minipage}{0.24\linewidth}
    \includegraphics[width=\linewidth]{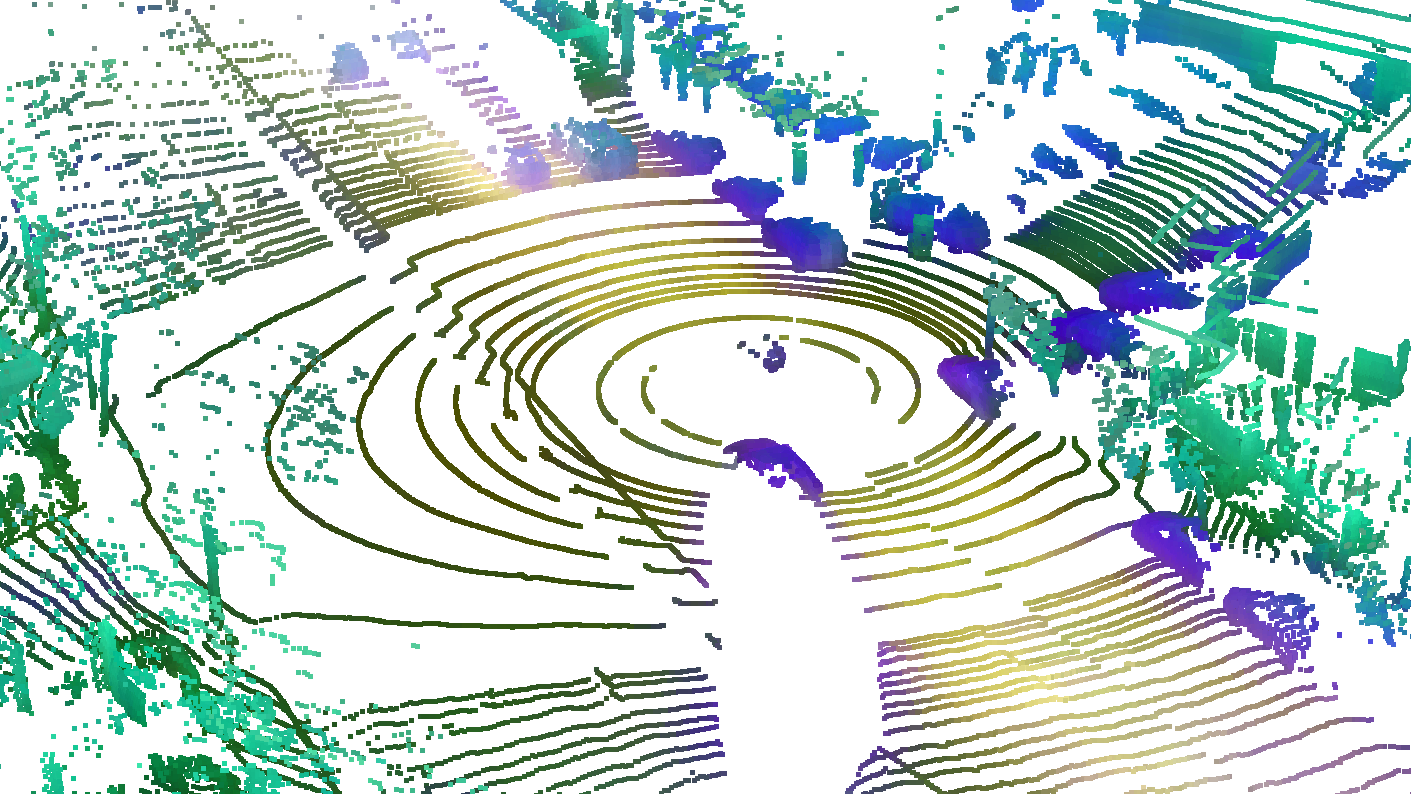}    
    \end{minipage}
    \begin{minipage}{0.24\linewidth}
    \includegraphics[width=\linewidth]{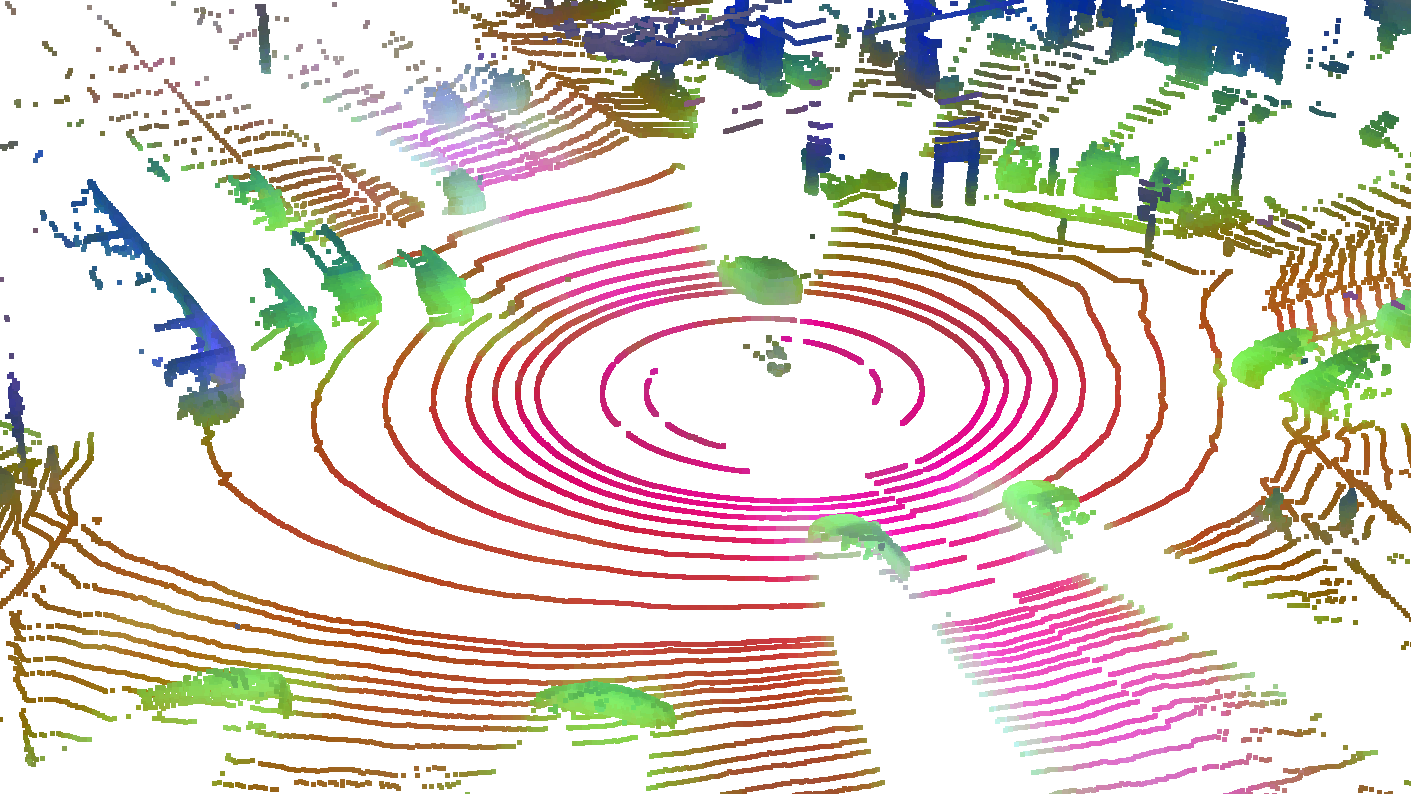}    
    \end{minipage}
    \begin{minipage}{0.24\linewidth}
    \includegraphics[width=\linewidth]{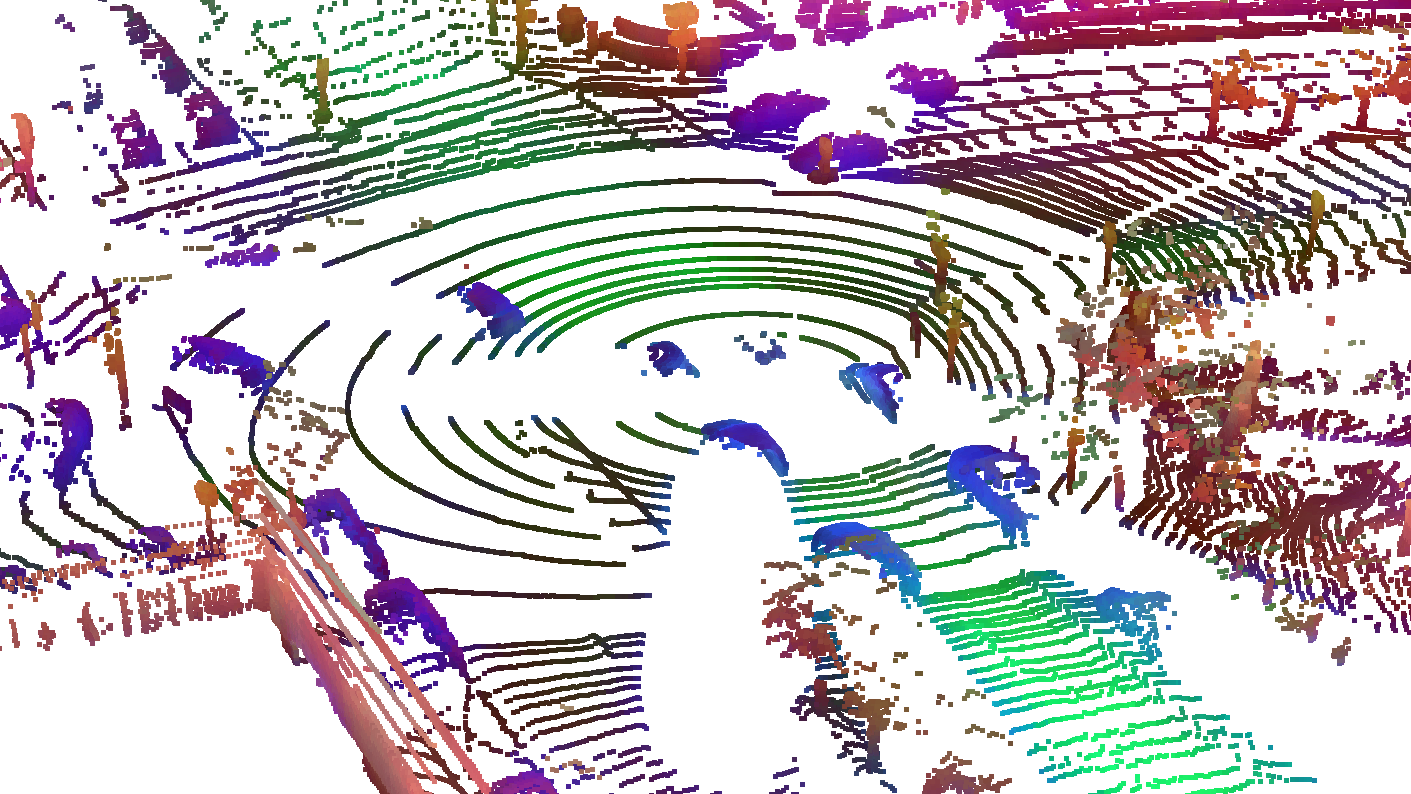}    
    \end{minipage}
    \begin{minipage}{0.24\linewidth}
    \includegraphics[width=\linewidth]{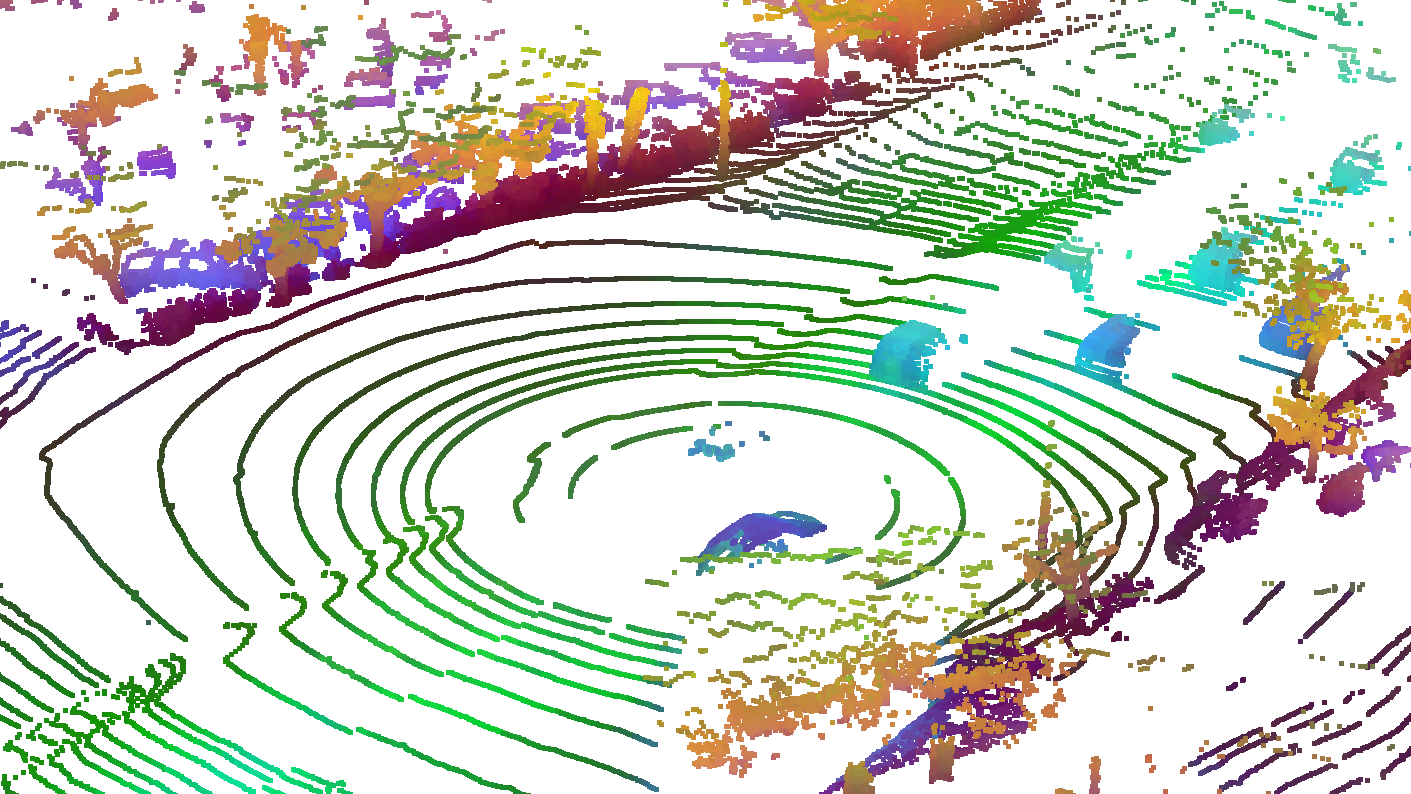}    
    \end{minipage}
    \\
    \begin{minipage}{3mm}
    \rotatebox[origin=c]{90}{PandarGT}
    \end{minipage}
    \begin{minipage}{0.24\linewidth}
    \includegraphics[width=\linewidth]{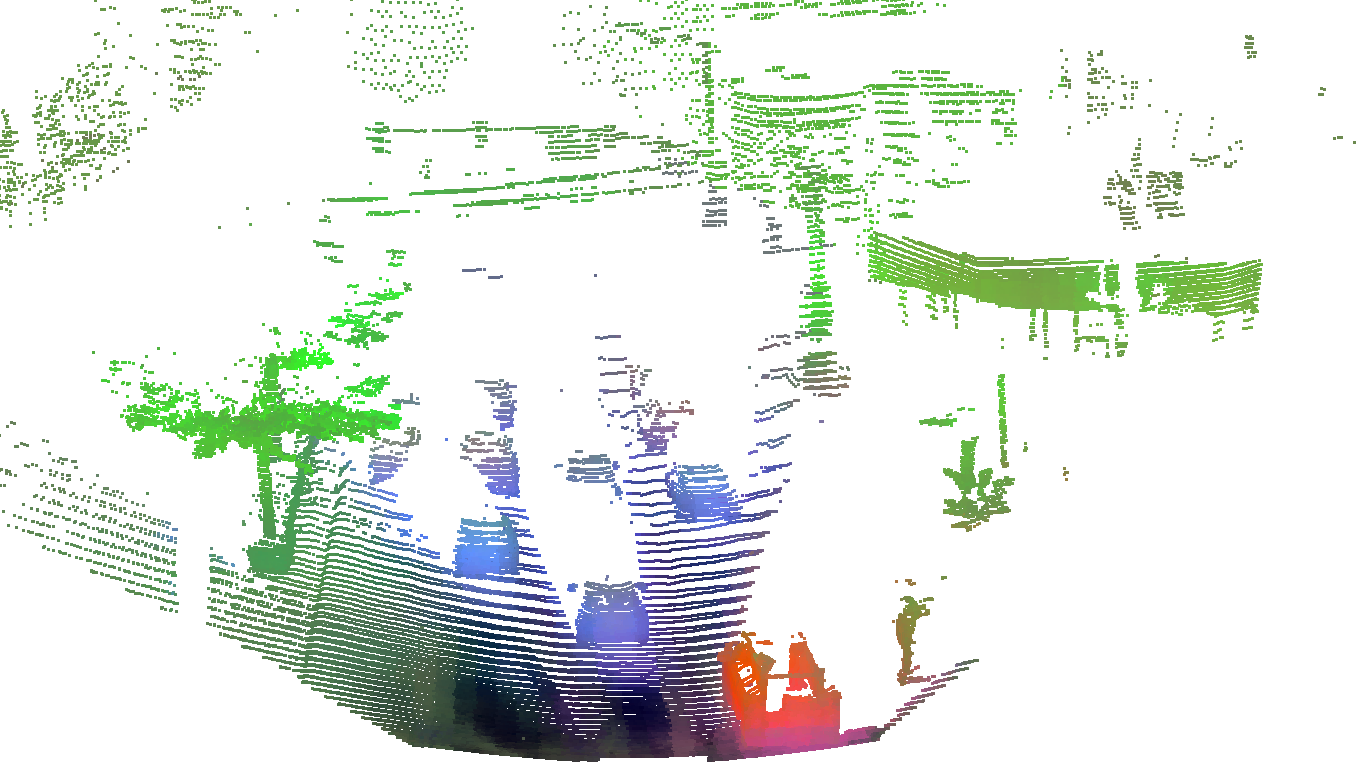}    
    \end{minipage}
    \begin{minipage}{0.24\linewidth}
    \includegraphics[width=\linewidth]{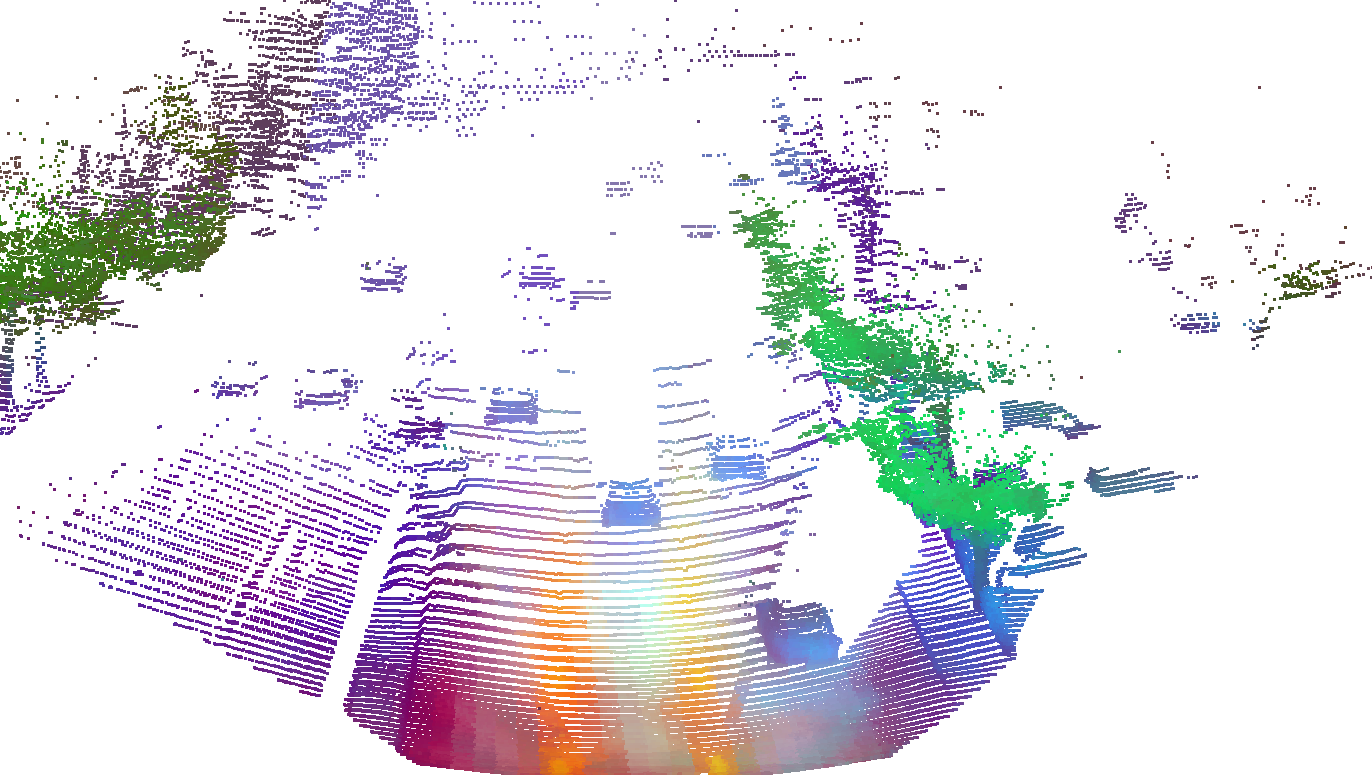}    
    \end{minipage}
    \begin{minipage}{0.24\linewidth}
    \includegraphics[width=\linewidth]{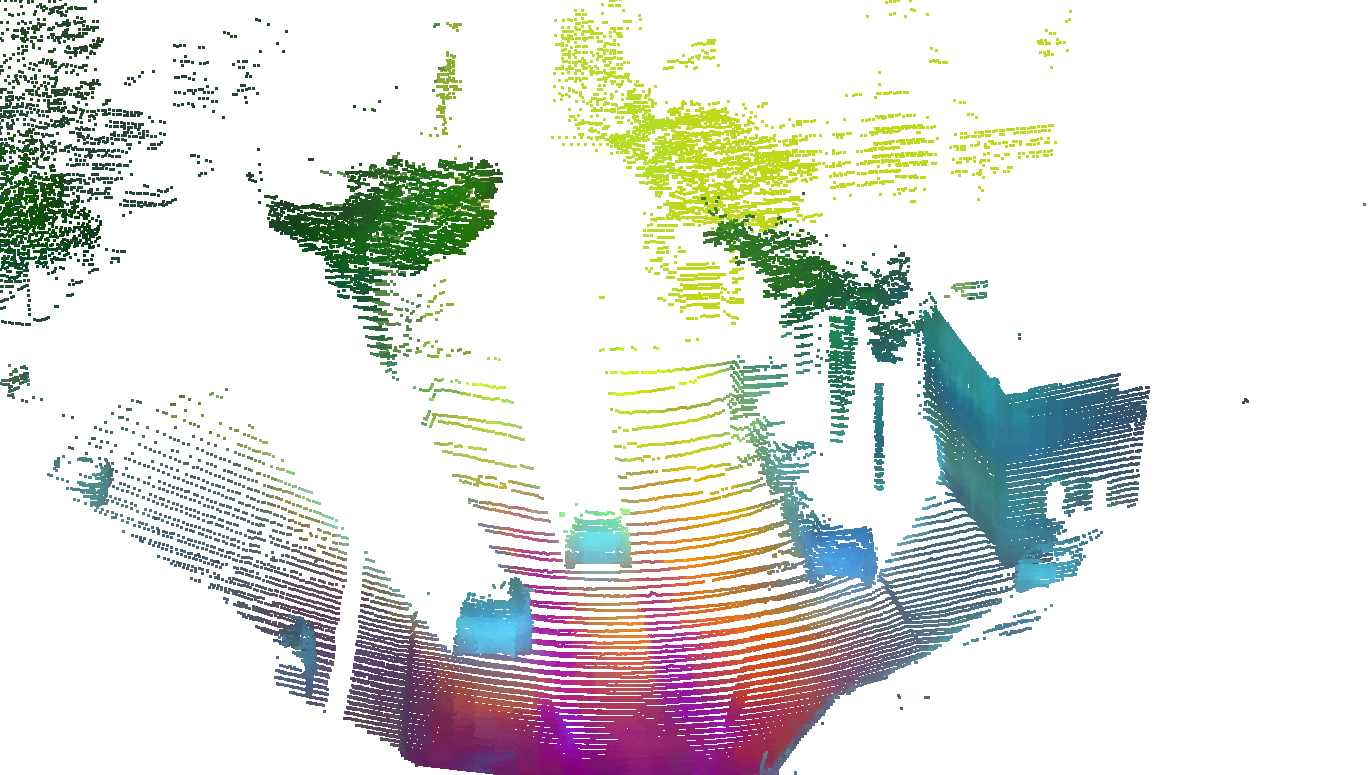}    
    \end{minipage}
    \begin{minipage}{0.24\linewidth}
    \includegraphics[width=\linewidth]{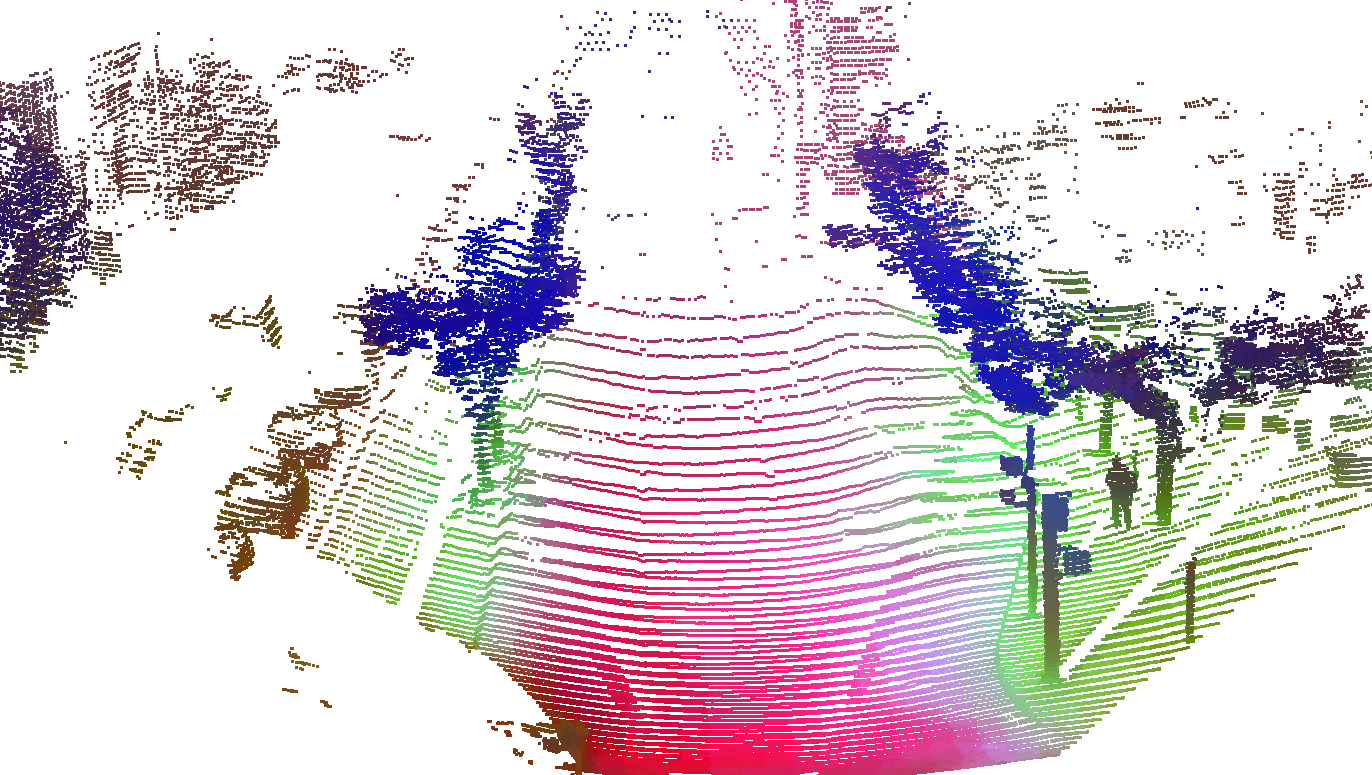}    
    \end{minipage}
\caption{\textbf{Distilled feature visualizations}. We project the features at the output of $\backthreed$ into a three-dimensional space by PCA. The projected value serves as RGB value to color the point clouds, i.e., the first, second and third components are used as the red, green and blue channels, respectively. Note that the PCA is done independently for each scan, which explains why the colors are not consistent from one scan to another. In this figure, we used the WI-768 pretrained on nuScenes, SemanticKITTI, Pandar 64 and Pandar GT with \ours.}
\label{fig:pca_features}
\end{figure*}

We provide a visualization of the features computed by a \ours-pretrained $\backthreed$ backbone in \cref{fig:pca_features}. We use our WI-768 pretrained on nuScenes, SemanticKITTI, Pandar64 and PandarGT. In this figure, the features are projected onto the space spanned by their 3 principal components and used as RGB values to color the point clouds. Note that the PCA is done independently on each scan, which explains why the colors are not consistent from one scan to another.

We notice that the feature space of our pretrained backbone is correctly structured as we can distinguish rather easily the main urban constructions and objects in these figures. For example, we notice that the points belonging to road and sidewalk have similar colors (per scan) on nuScenes and SemanticKITTI. On Pandar64 and PandarGT, we also notice that the cars have similar colors (per scan) as well. Let us mention that the road on PandarGT scans appears in a less uniform color than on the other datasets. It could be explained by a higher density of points on the road for this lidar, which might lead to more subtle differences between features after distillation and/or PCA.

\begin{figure*}
\small
\centering
    \begin{minipage}{0.24\linewidth}
    \centering \small Car
    \end{minipage}
    \begin{minipage}{0.24\linewidth}
    \centering \small Pedestrian
    \end{minipage}
    \begin{minipage}{0.24\linewidth}
    \centering \small Road
    \end{minipage}
    \begin{minipage}{0.24\linewidth}
    \centering \small Sidewalk
    \end{minipage}
    \\
    \begin{minipage}{3mm}
    \rotatebox[origin=c]{90}{nuScenes - Sim}
    \end{minipage}
    \begin{minipage}{0.24\linewidth}
    \includegraphics[width=\linewidth]{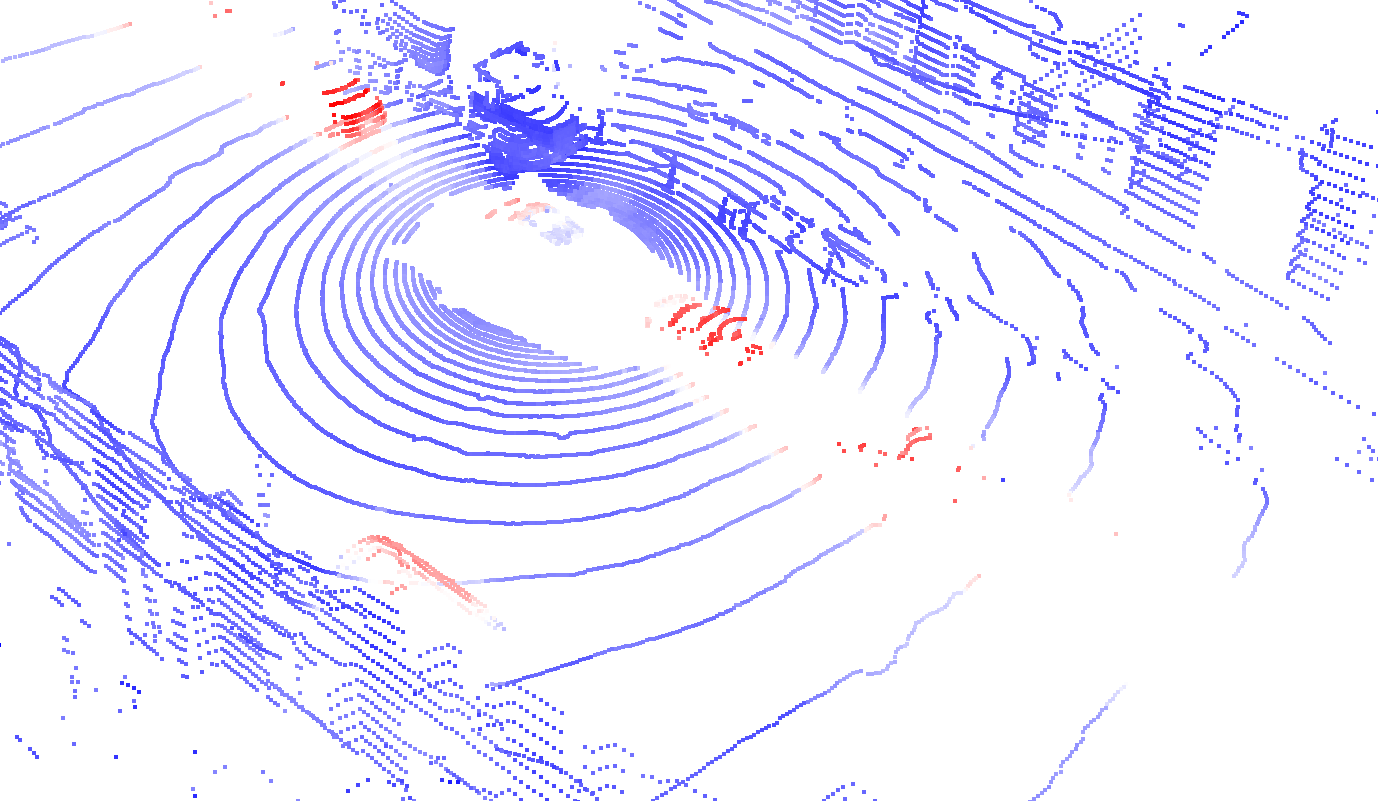}    
    \end{minipage}
    \begin{minipage}{0.24\linewidth}
    \includegraphics[width=\linewidth]{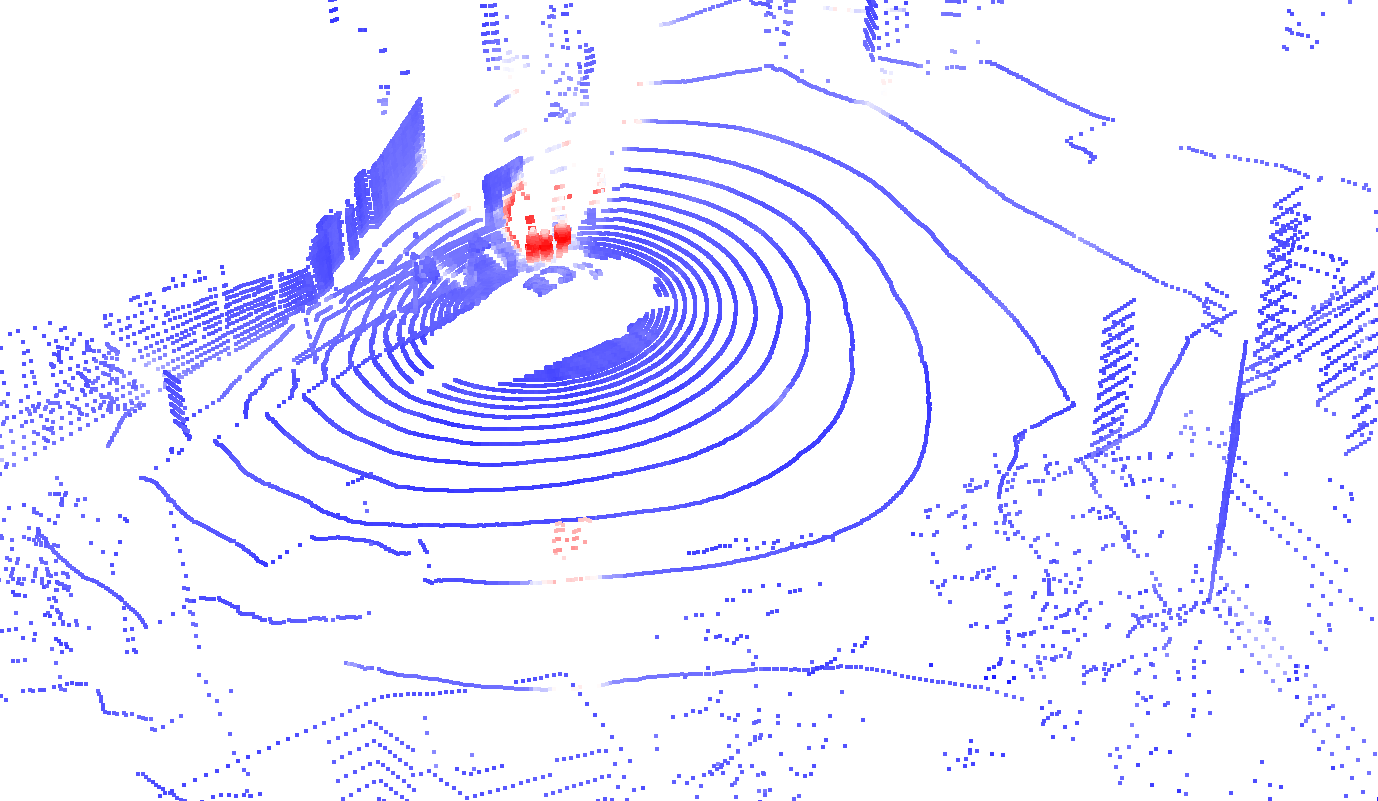}    
    \end{minipage}
    \begin{minipage}{0.24\linewidth}
    \includegraphics[width=\linewidth]{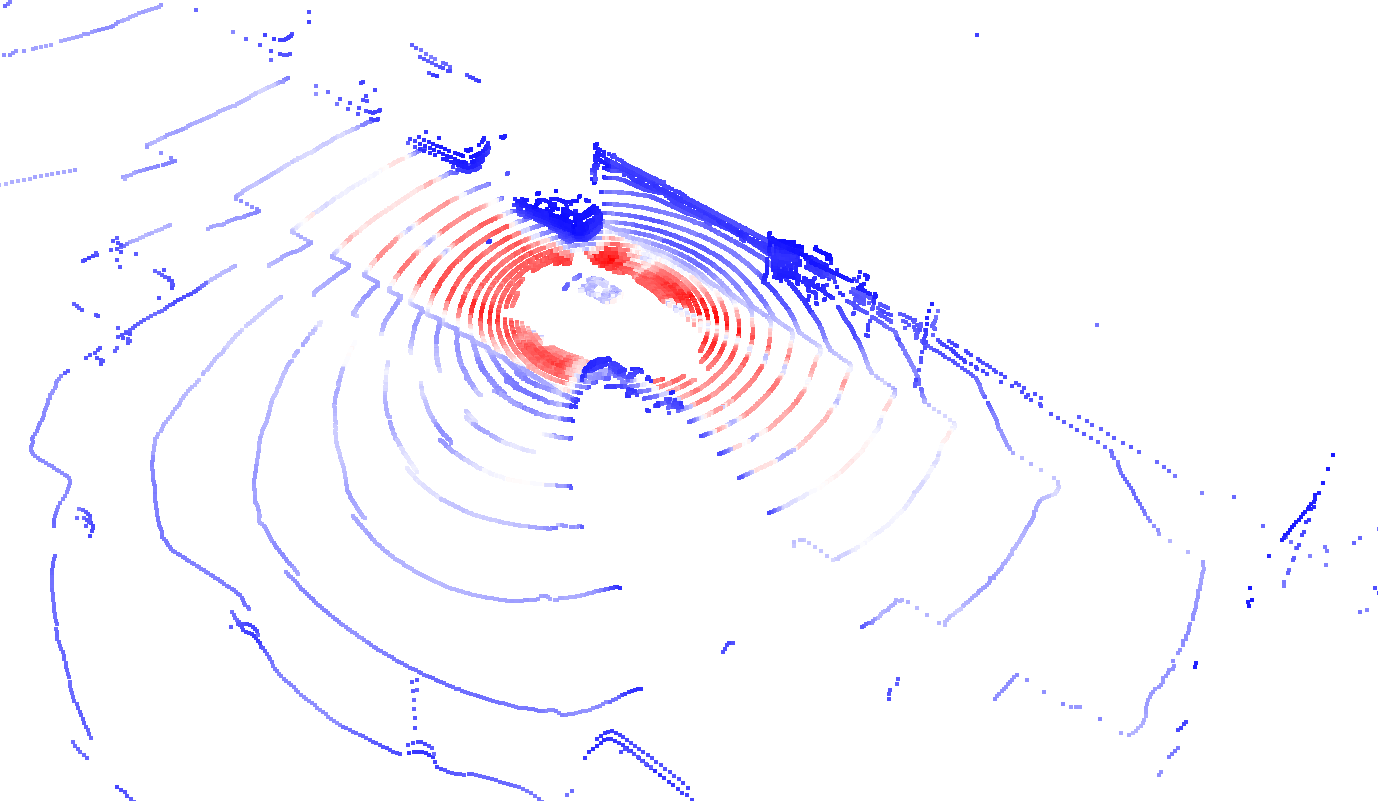}    
    \end{minipage}
    \begin{minipage}{0.24\linewidth}
    \includegraphics[width=\linewidth]{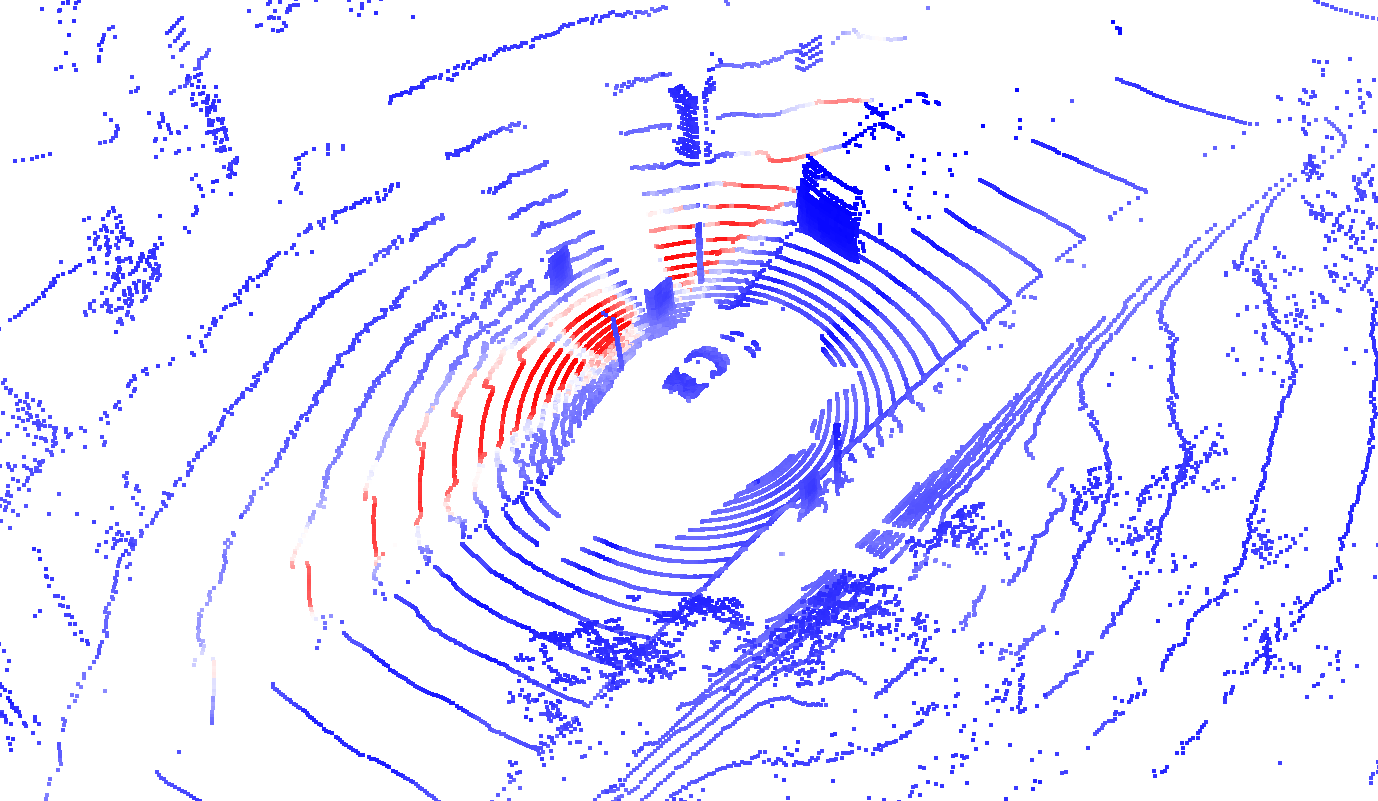}    
    \end{minipage}
    \\
    \rotatebox[origin=c]{90}{\small nuScenes - Label}
    \begin{minipage}{0.24\linewidth}
    \includegraphics[width=\linewidth]{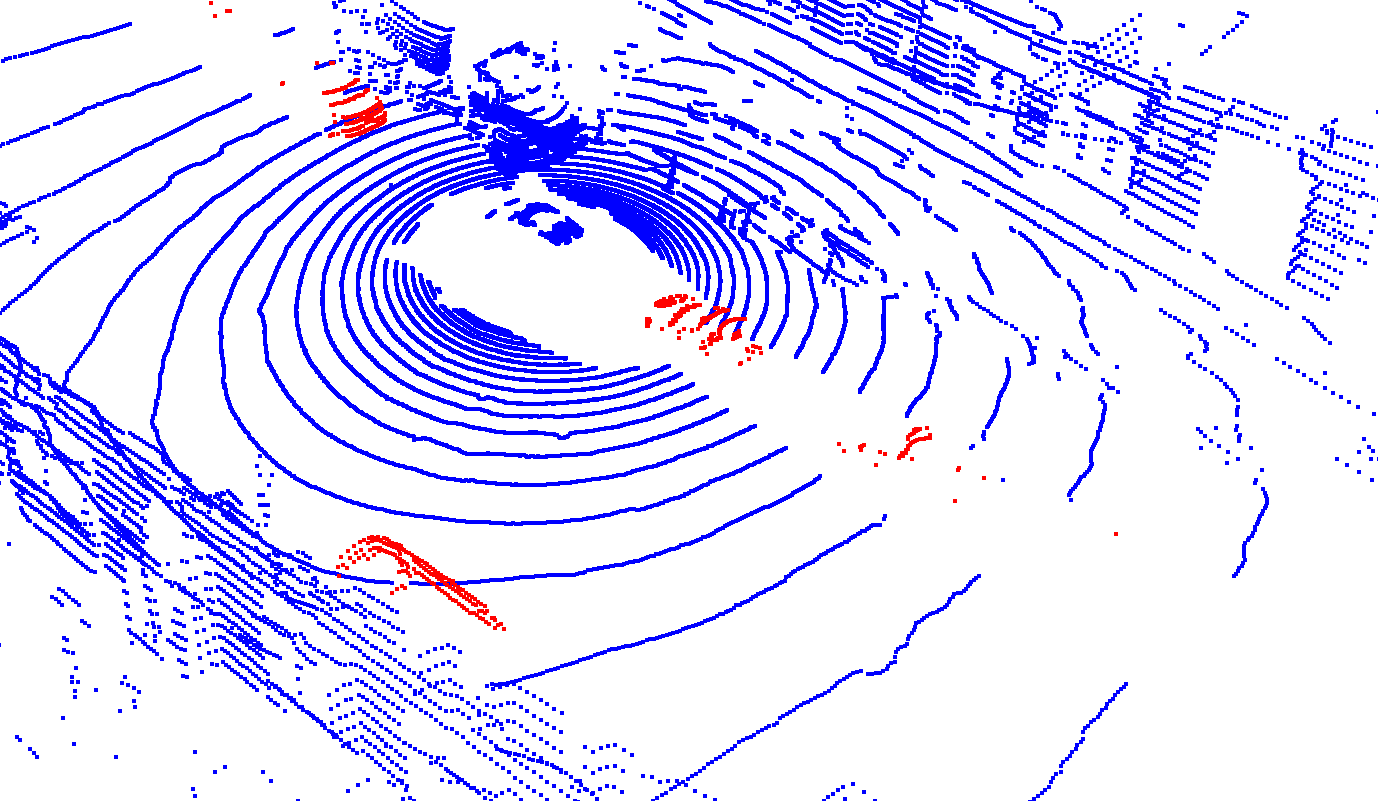}    
    \end{minipage}
    \begin{minipage}{0.24\linewidth}
    \includegraphics[width=\linewidth]{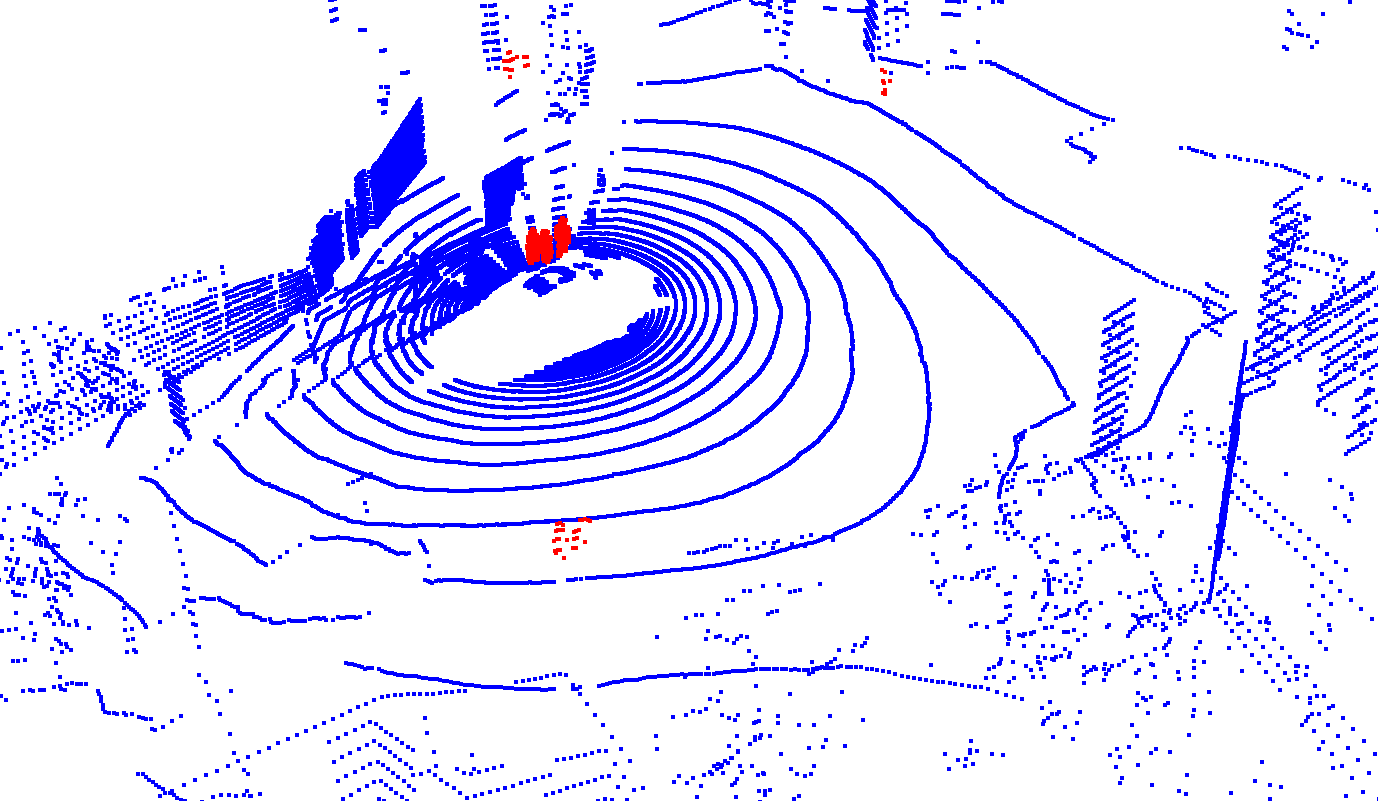}    
    \end{minipage}
    \begin{minipage}{0.24\linewidth}
    \includegraphics[width=\linewidth]{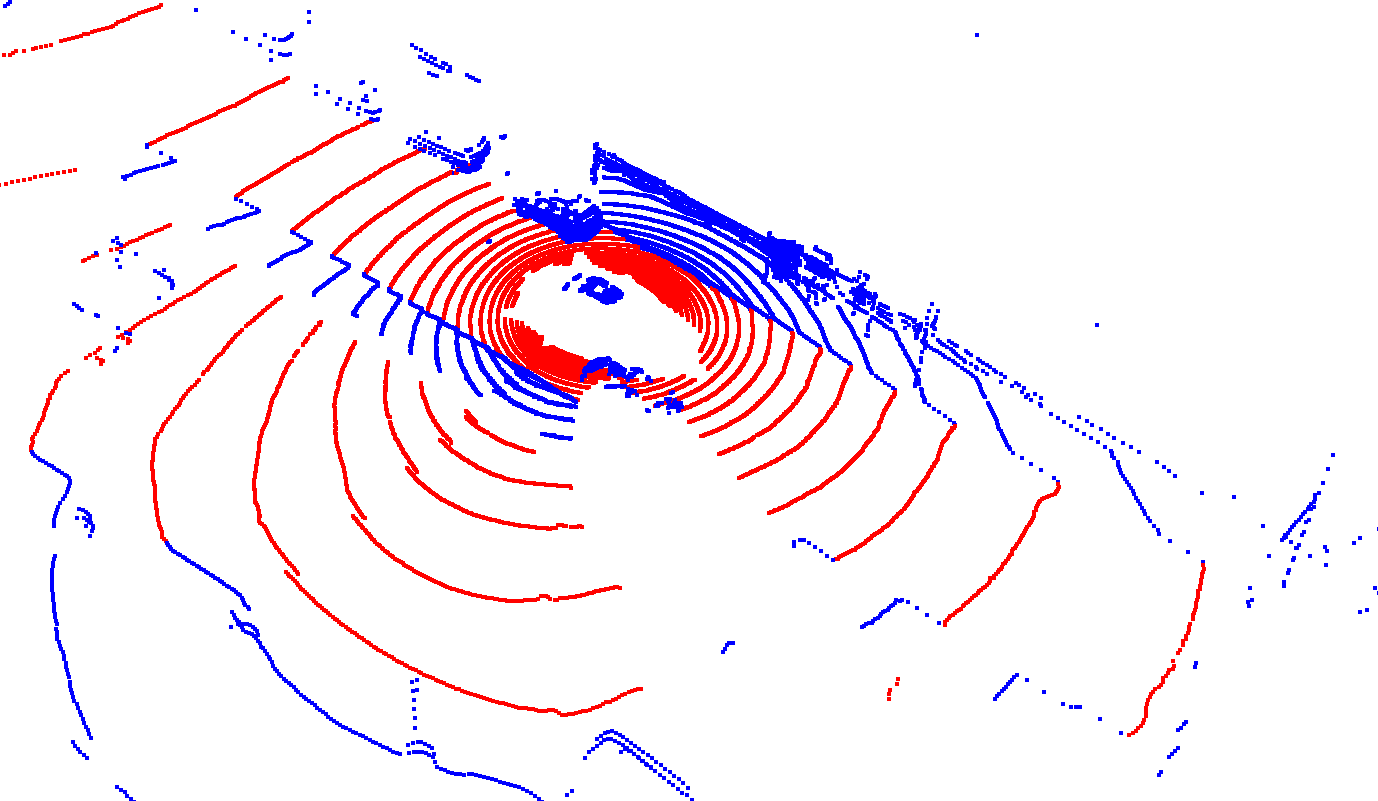}    
    \end{minipage}
    \begin{minipage}{0.24\linewidth}
    \includegraphics[width=\linewidth]{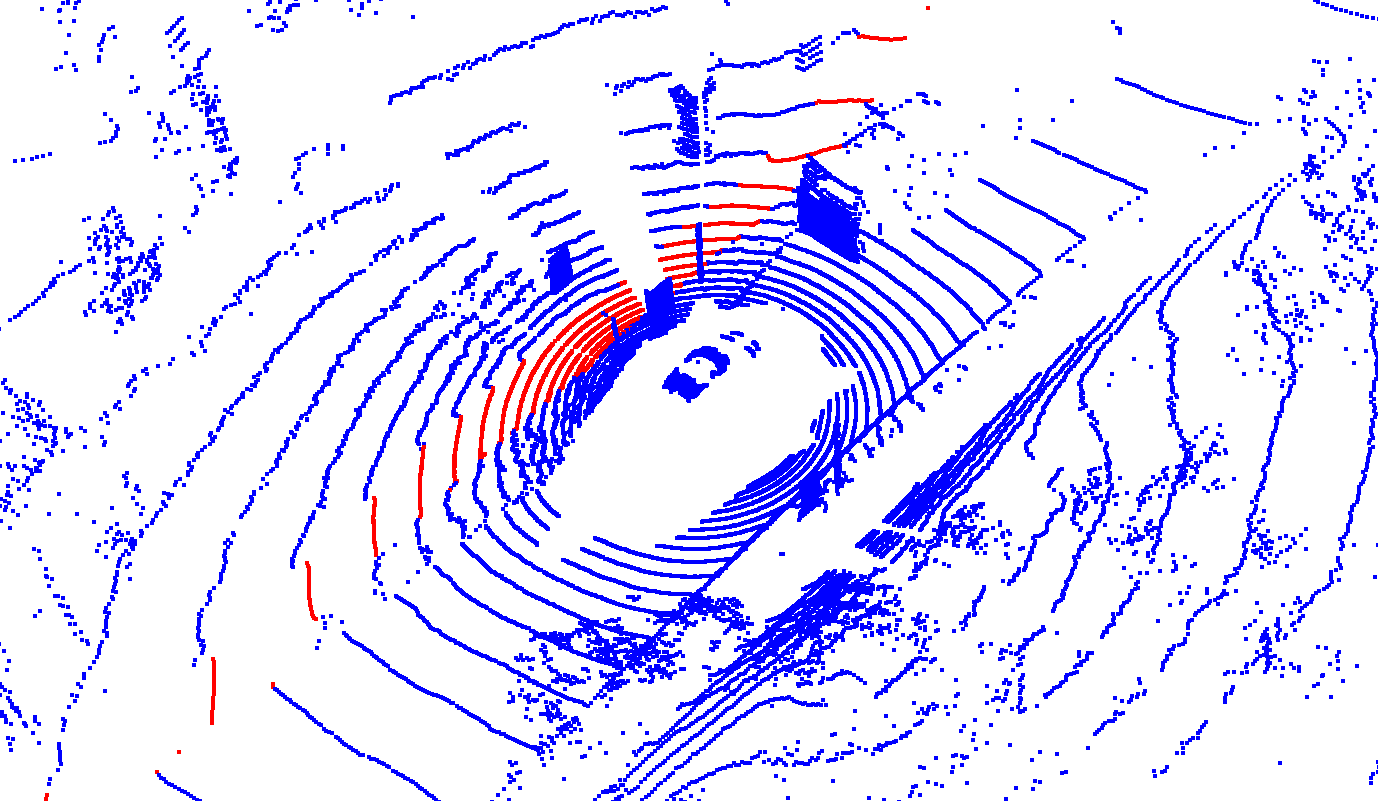}    
    \end{minipage}
    \vspace{1mm}
    \\    
    \hrule
    \vspace{1mm}
    \begin{minipage}{3mm}
    \rotatebox[origin=c]{90}{KITTI - Sim}
    \end{minipage}
    \begin{minipage}{0.24\linewidth}
    \includegraphics[width=\linewidth]{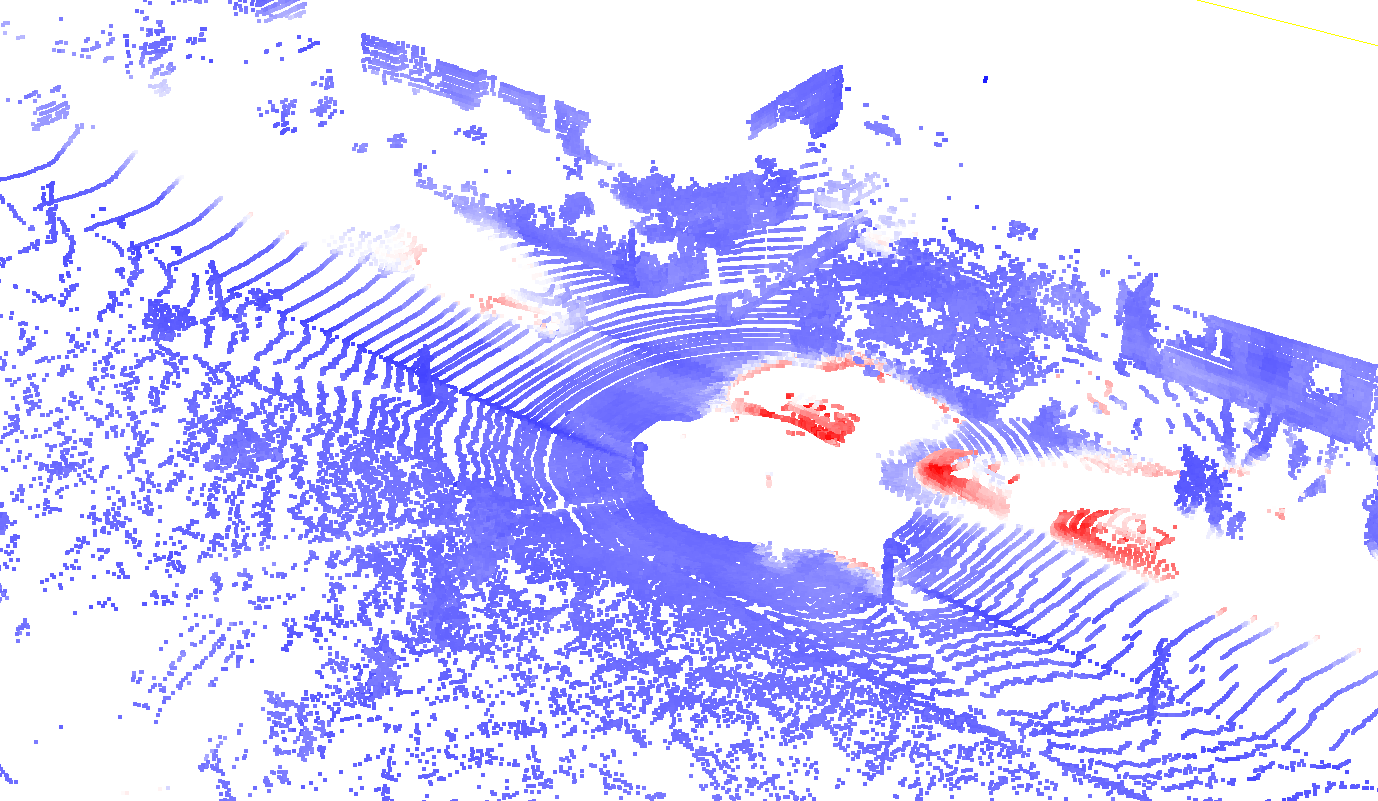}
    \end{minipage}
    \begin{minipage}{0.24\linewidth}
    \includegraphics[width=\linewidth]{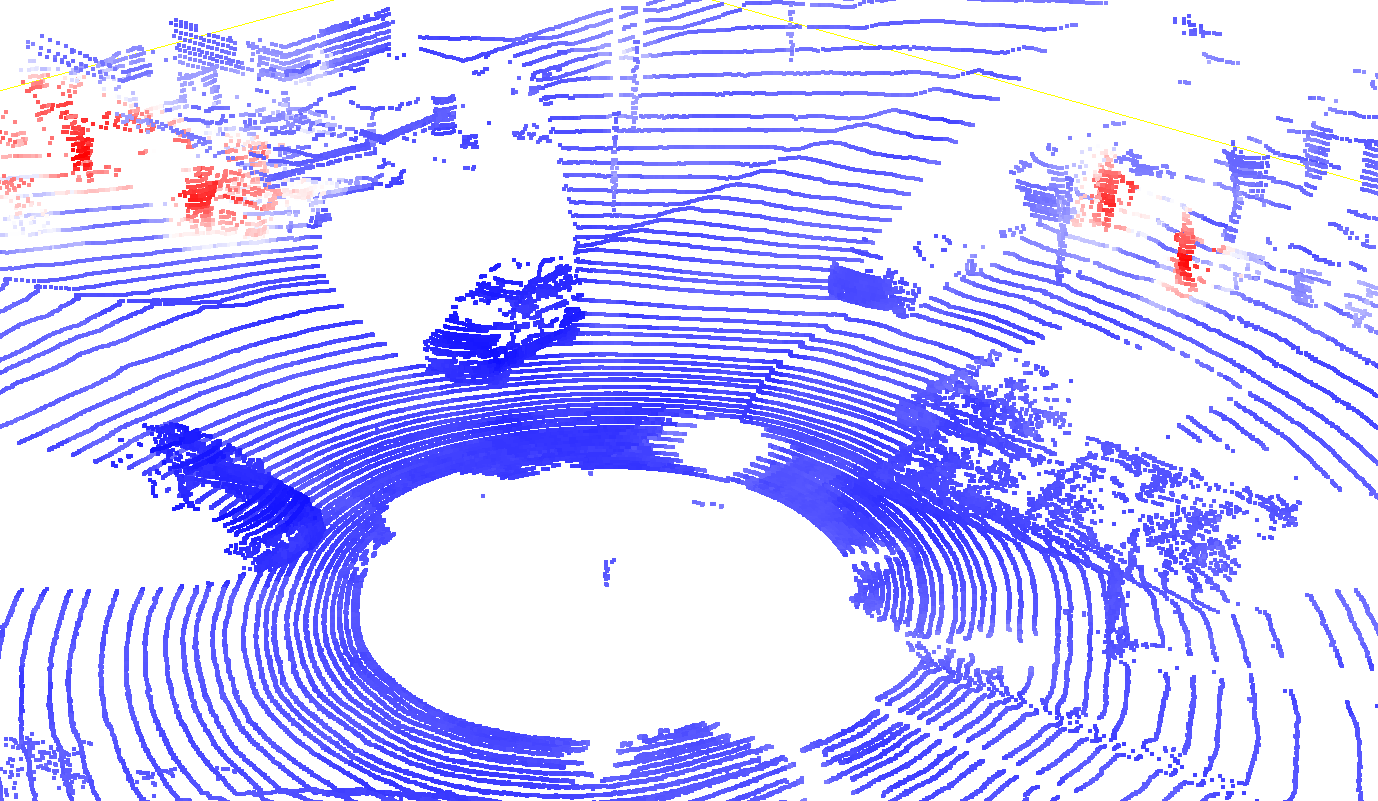}   
    \end{minipage}
    \begin{minipage}{0.24\linewidth}
    \includegraphics[width=\linewidth]{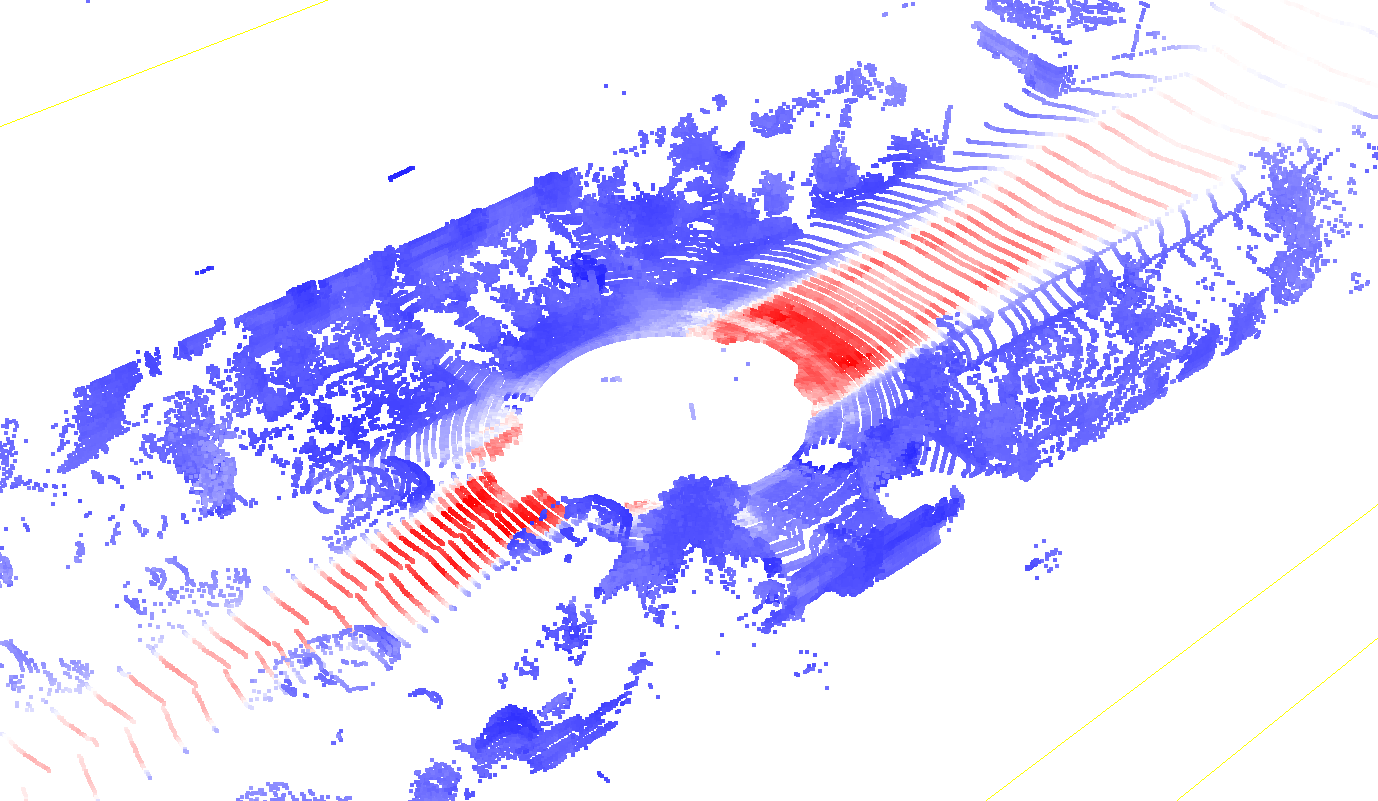} 
    \end{minipage}
    \begin{minipage}{0.24\linewidth}
    \includegraphics[width=\linewidth]{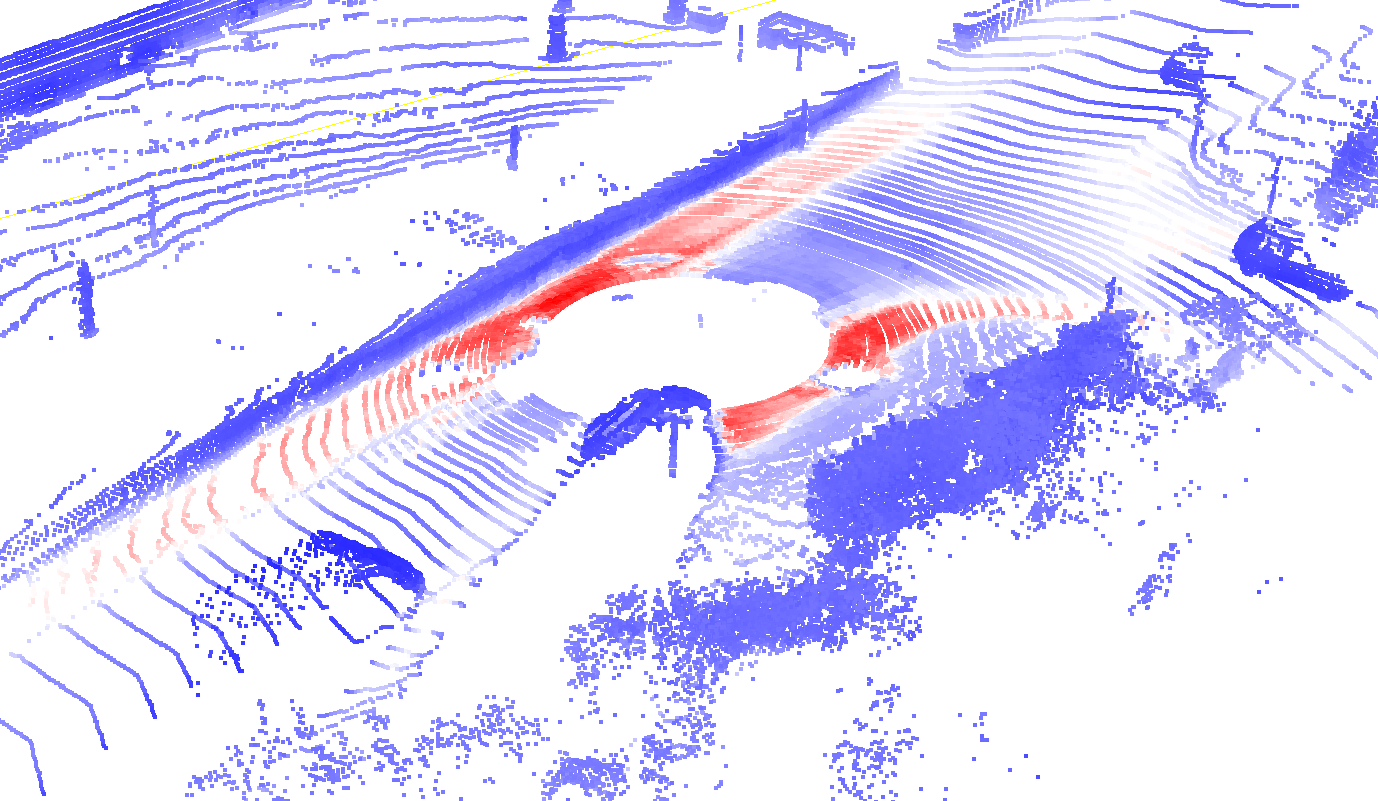}  
    \end{minipage}    
    \\
    \begin{minipage}{3mm}
    \rotatebox[origin=c]{90}{KITTI - Label}
    \end{minipage}
    \begin{minipage}{0.24\linewidth}
    \includegraphics[width=\linewidth]{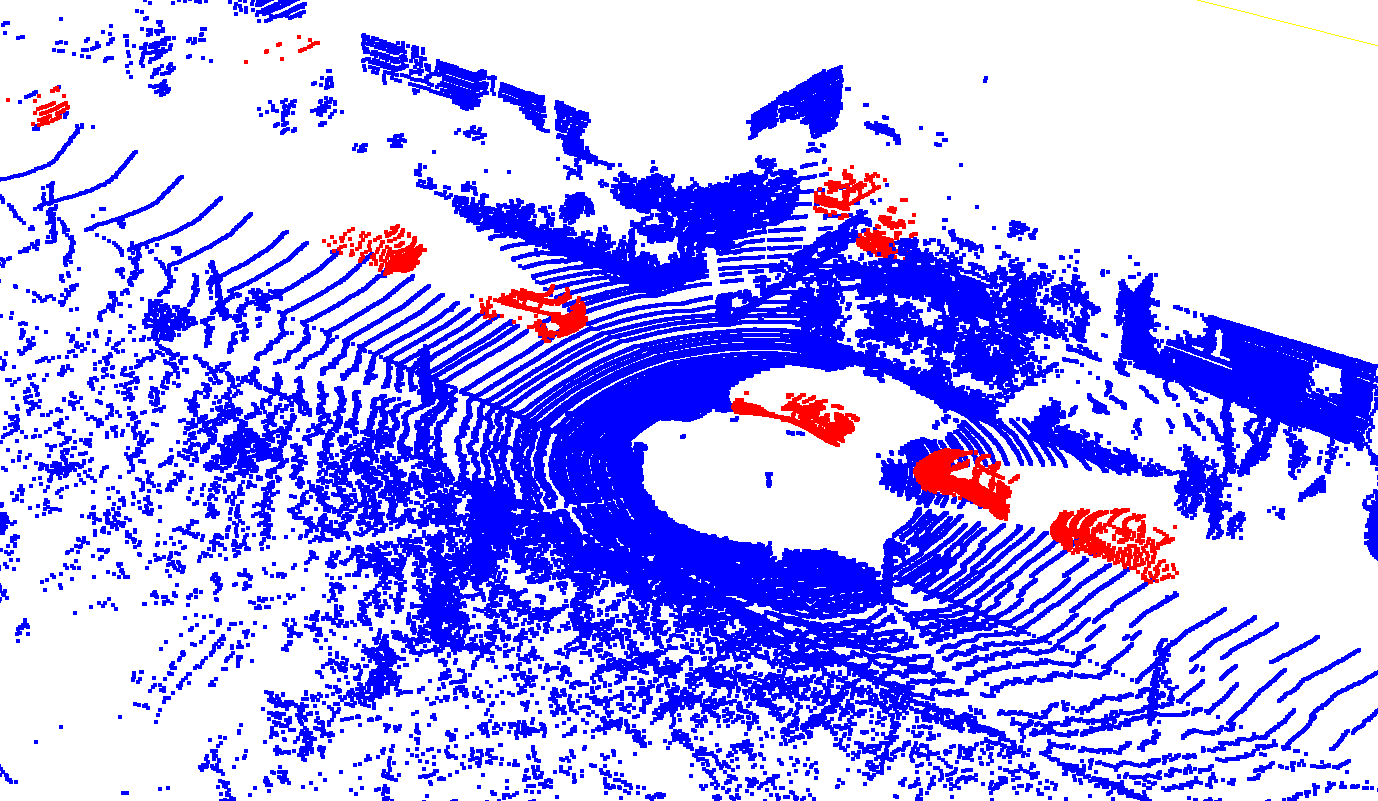} 
    \end{minipage}
    \begin{minipage}{0.24\linewidth}
    \includegraphics[width=\linewidth]{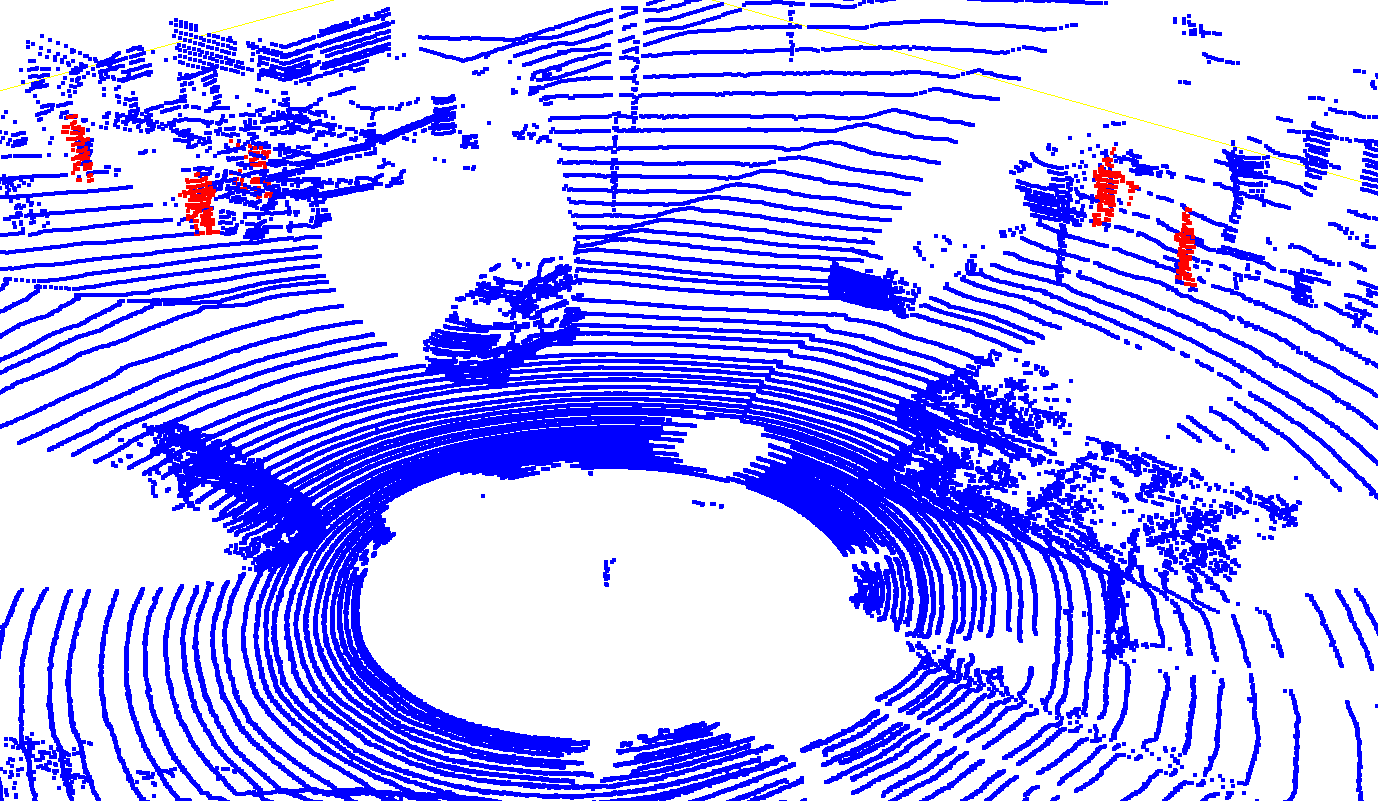}  
    \end{minipage}
    \begin{minipage}{0.24\linewidth}
    \includegraphics[width=\linewidth]{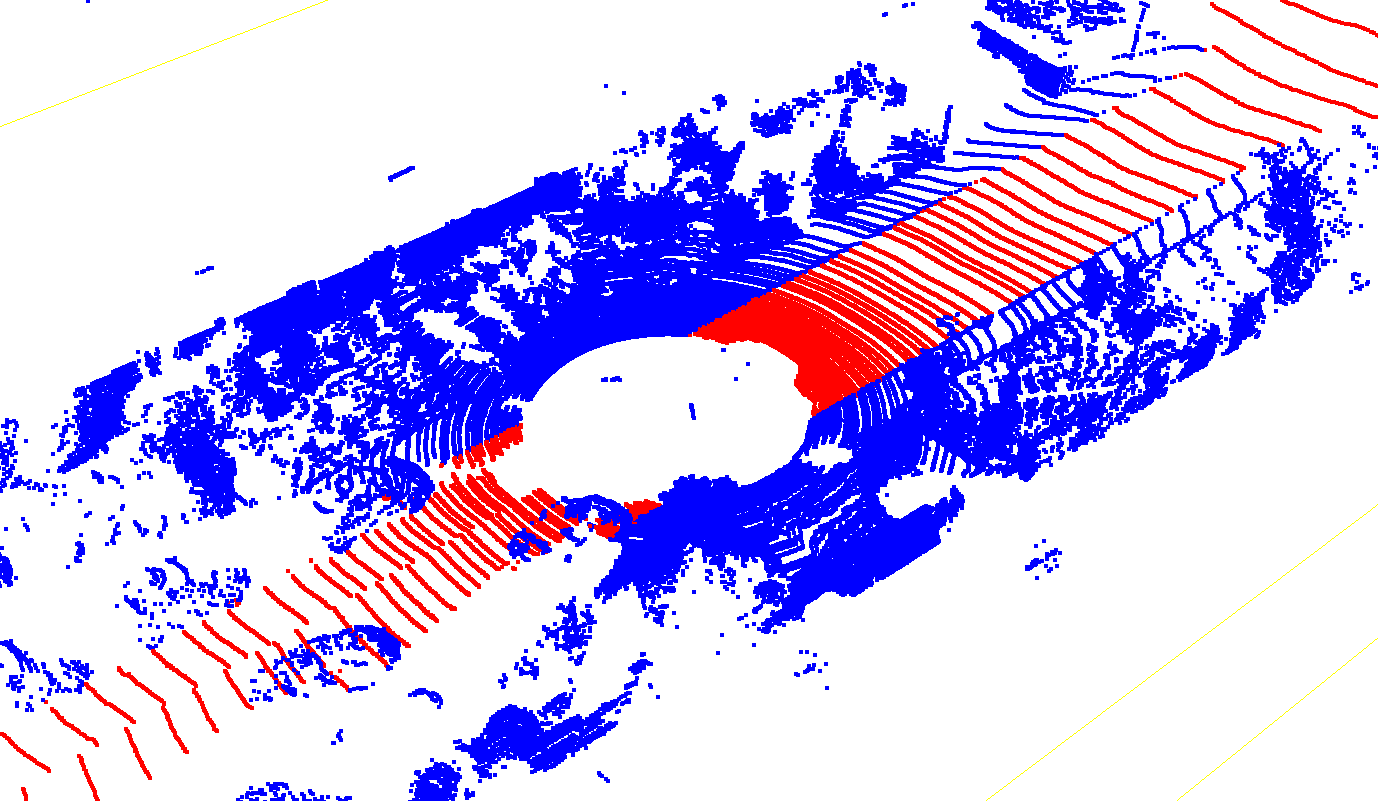}
    \end{minipage}
    \begin{minipage}{0.24\linewidth}
    \includegraphics[width=\linewidth]{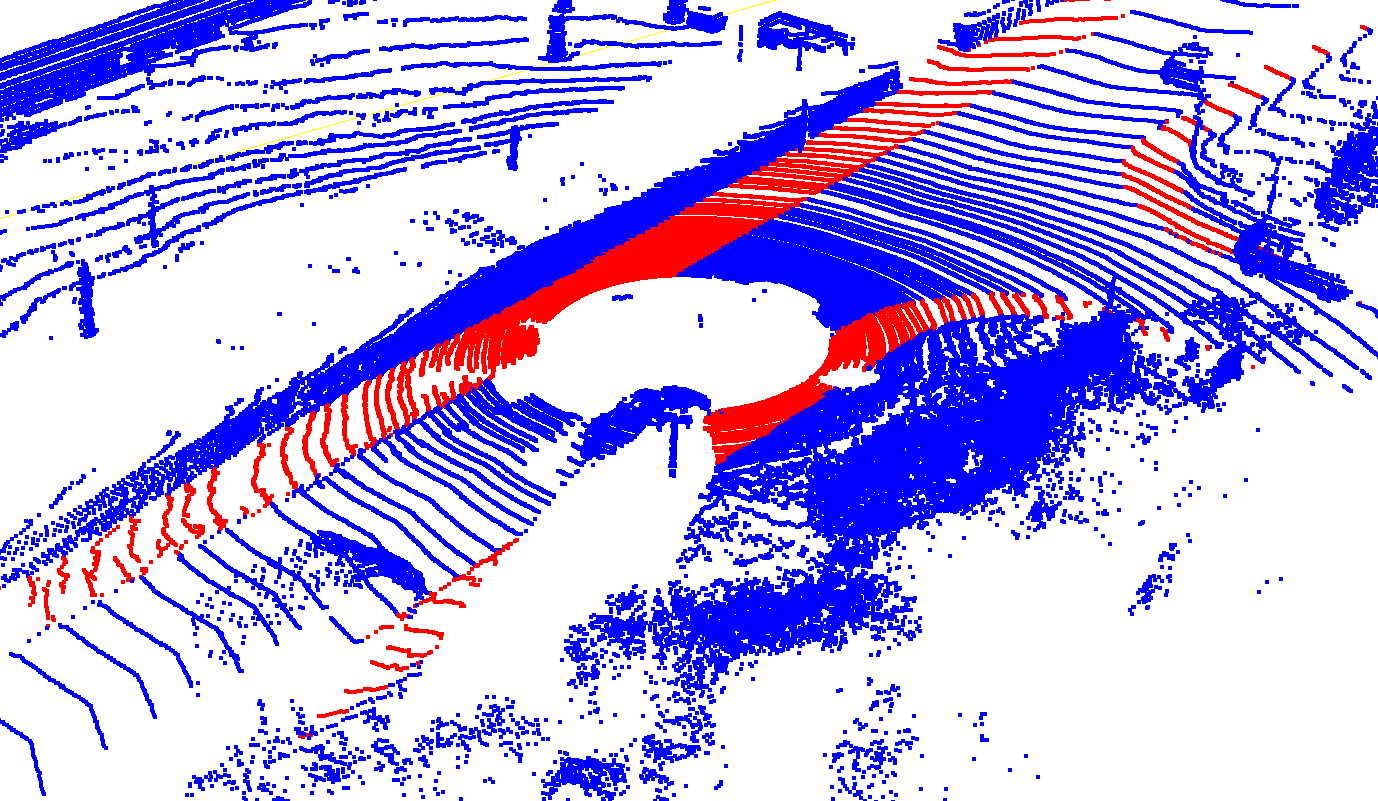}
    \end{minipage}
    \vspace{1mm}
    \\    
    \hrule
    \vspace{1mm}
    \begin{minipage}{3mm}
    \rotatebox[origin=c]{90}{Pan. 64 - Sim}
    \end{minipage}
    \begin{minipage}{0.24\linewidth}
    \includegraphics[width=\linewidth]{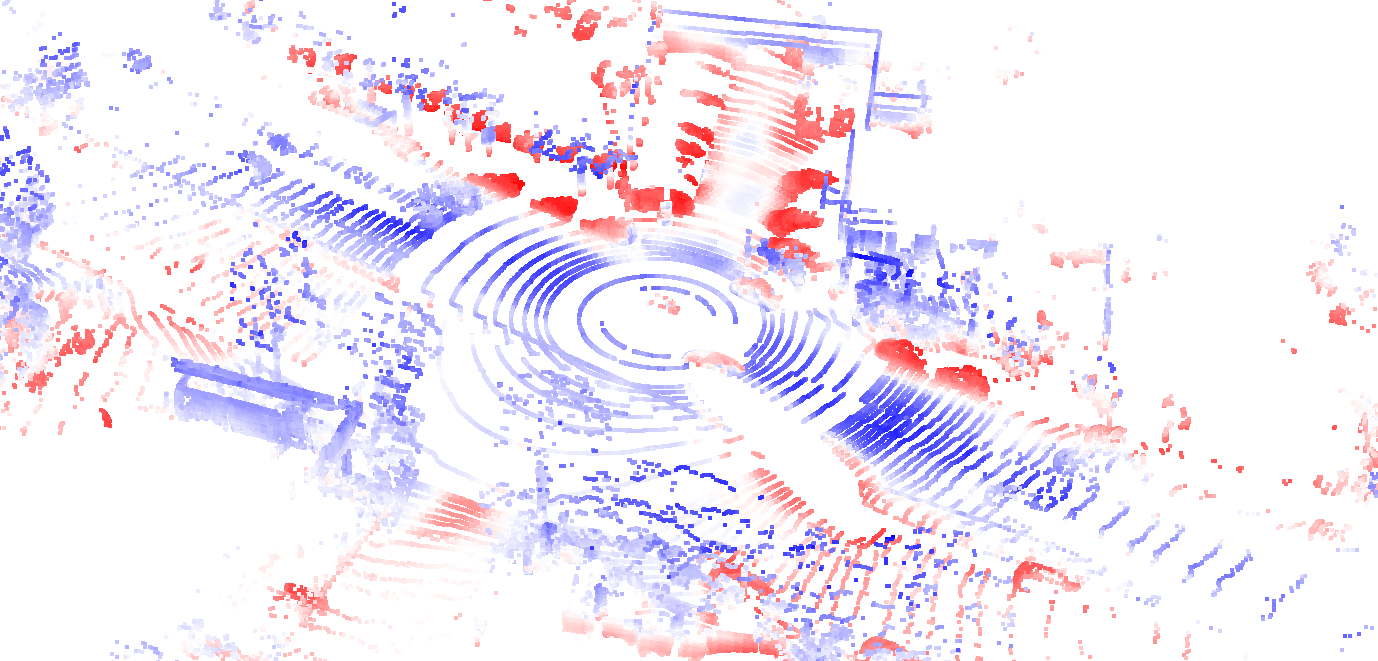}
    \end{minipage}
    \begin{minipage}{0.24\linewidth}
    \includegraphics[width=\linewidth]{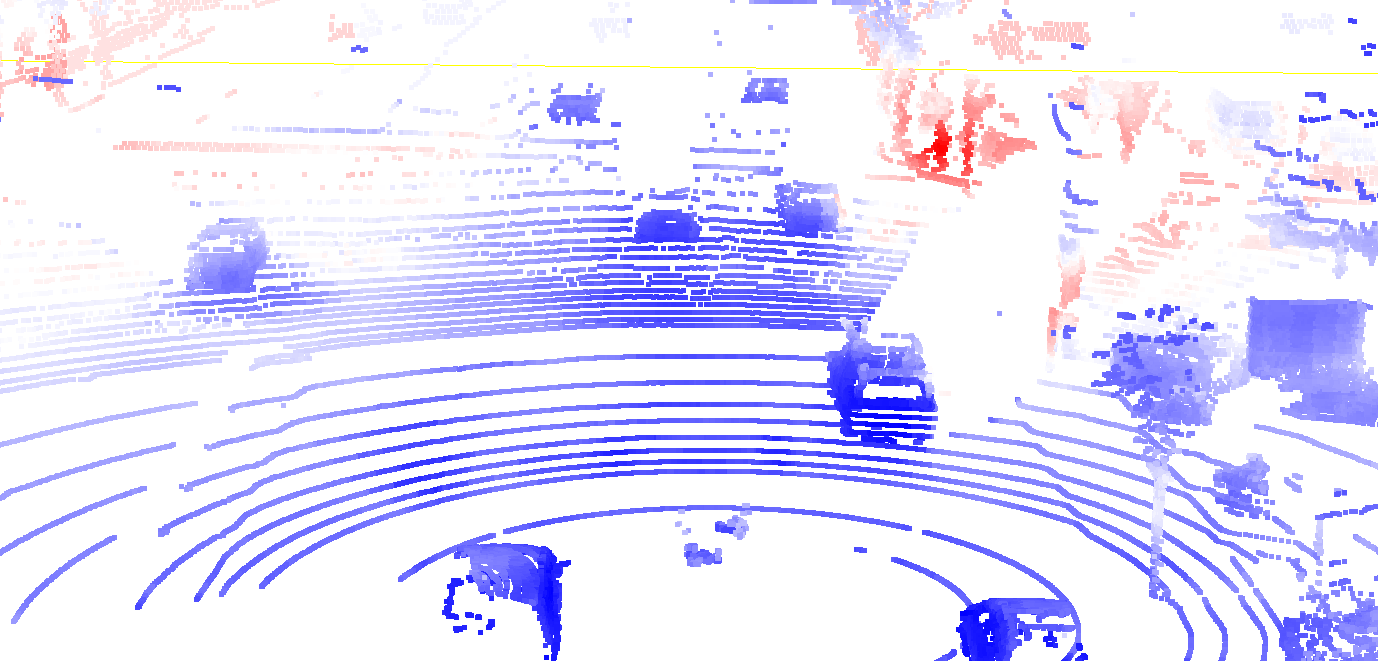}
    \end{minipage}
    \begin{minipage}{0.24\linewidth}
    \includegraphics[width=\linewidth]{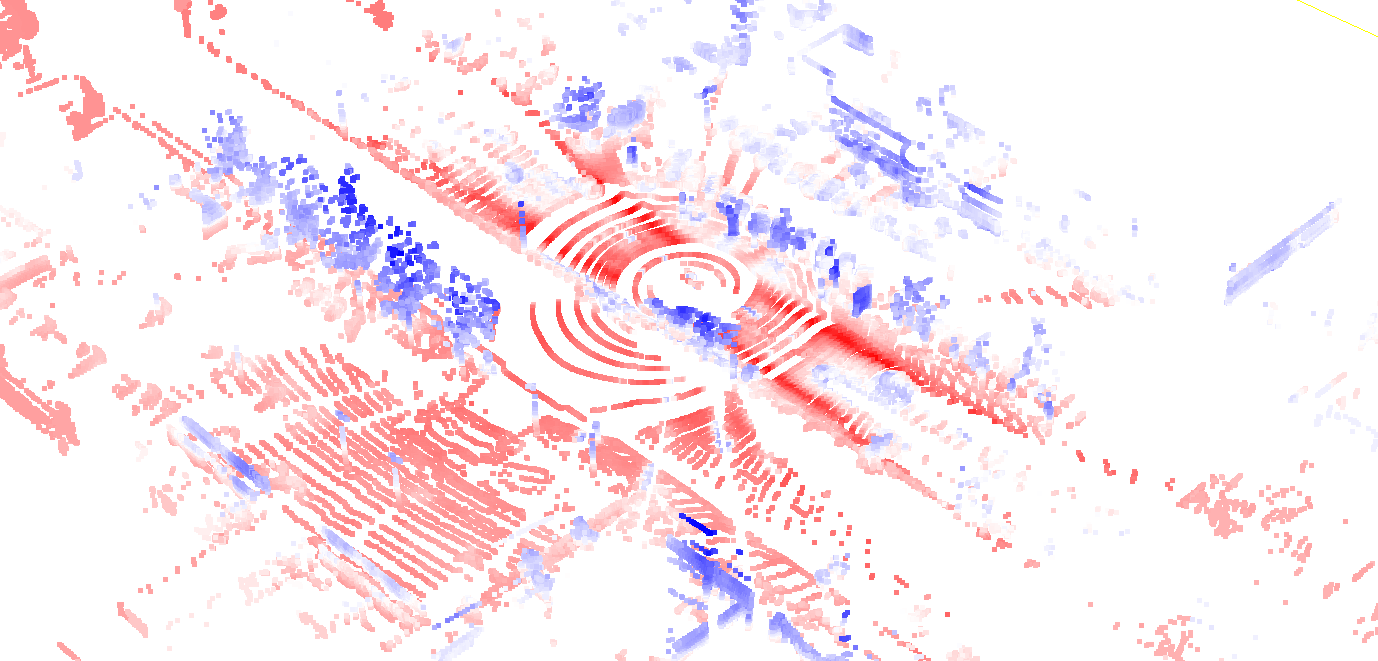}
    \end{minipage}
    \begin{minipage}{0.24\linewidth}
    \includegraphics[width=\linewidth]{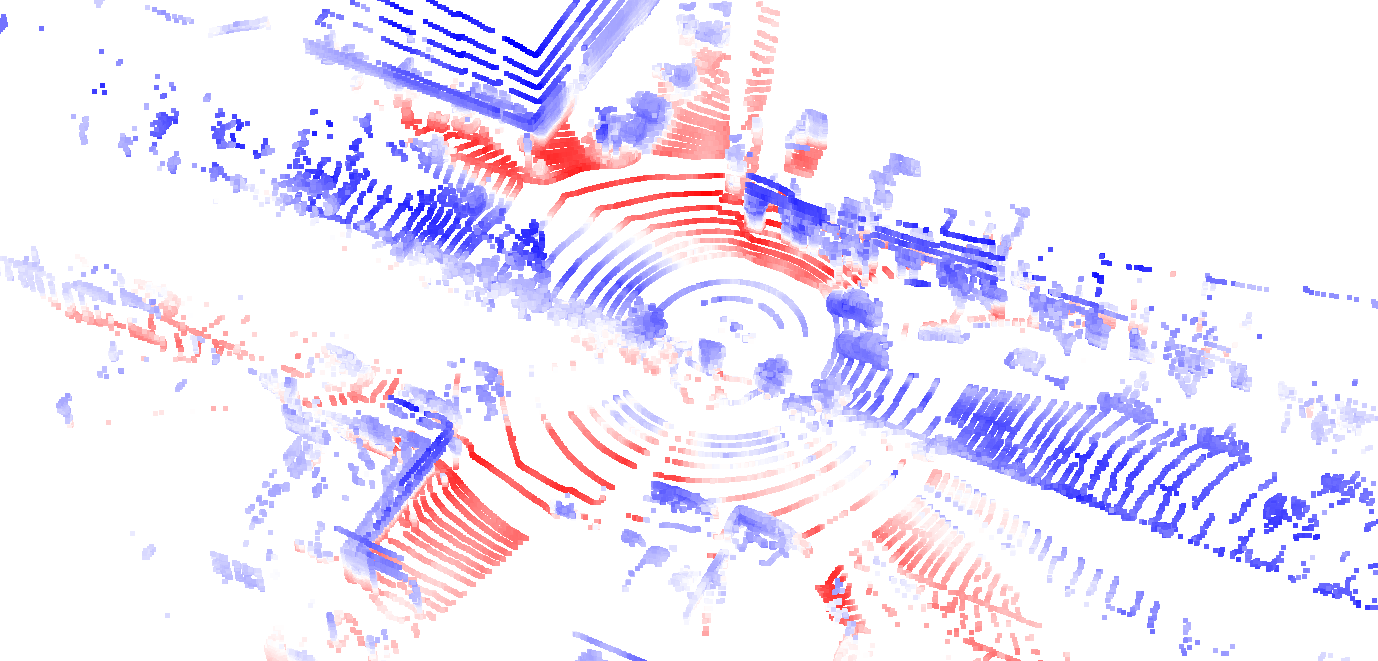}
    \end{minipage}    
    \\
    \begin{minipage}{3mm}
    \rotatebox[origin=c]{90}{Pan. 64 - Label}
    \end{minipage}
    \begin{minipage}{0.24\linewidth}
    \includegraphics[width=\linewidth]{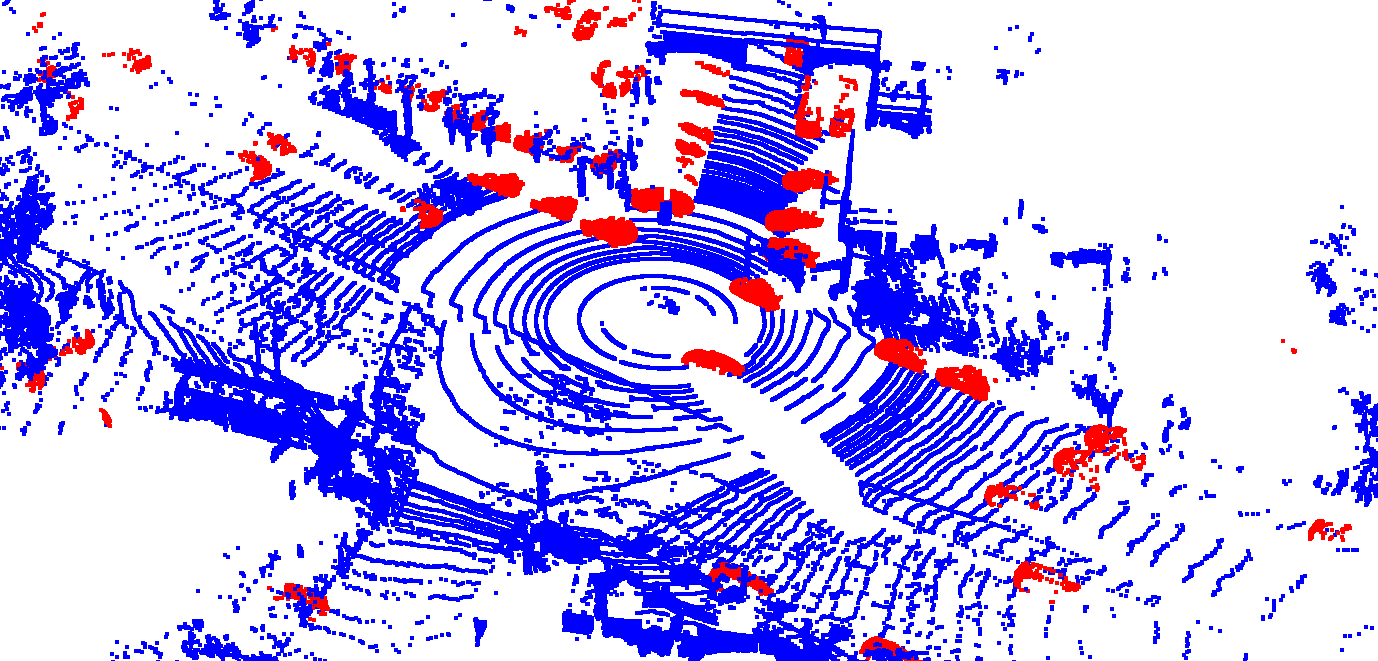}
    \end{minipage}
    \begin{minipage}{0.24\linewidth}
    \includegraphics[width=\linewidth]{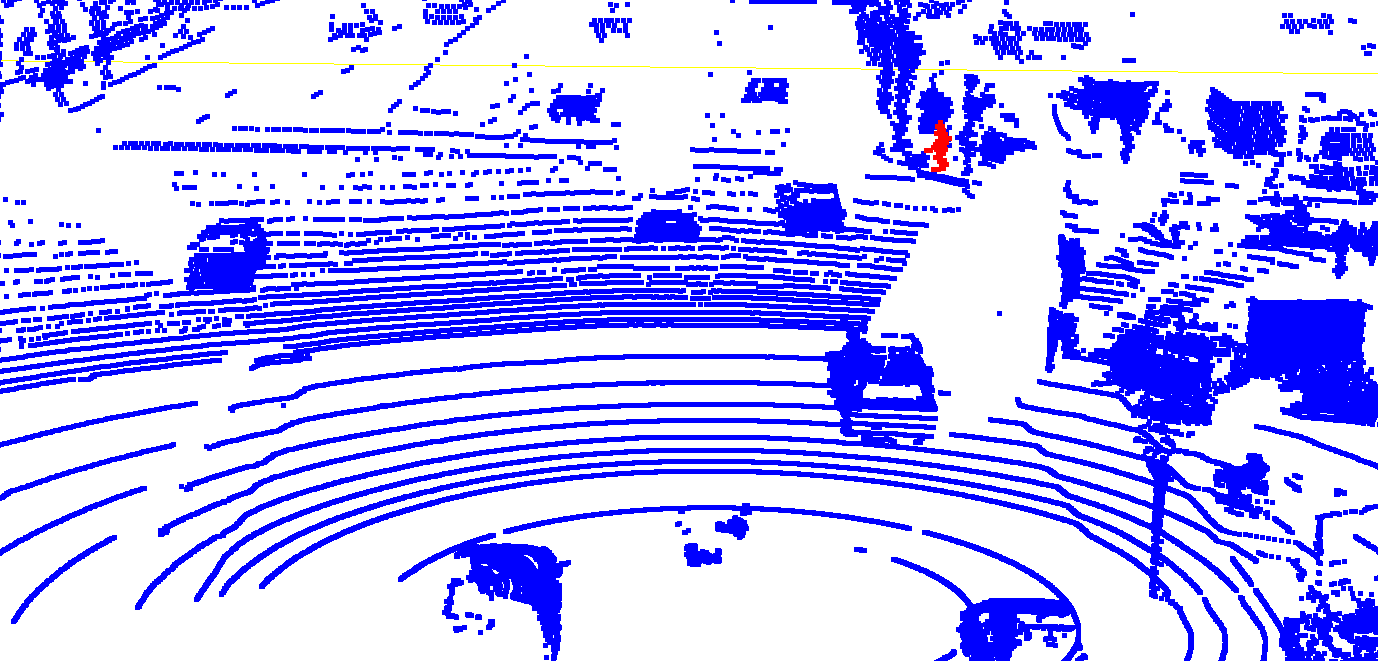}    
    \end{minipage}
    \begin{minipage}{0.24\linewidth}
    \includegraphics[width=\linewidth]{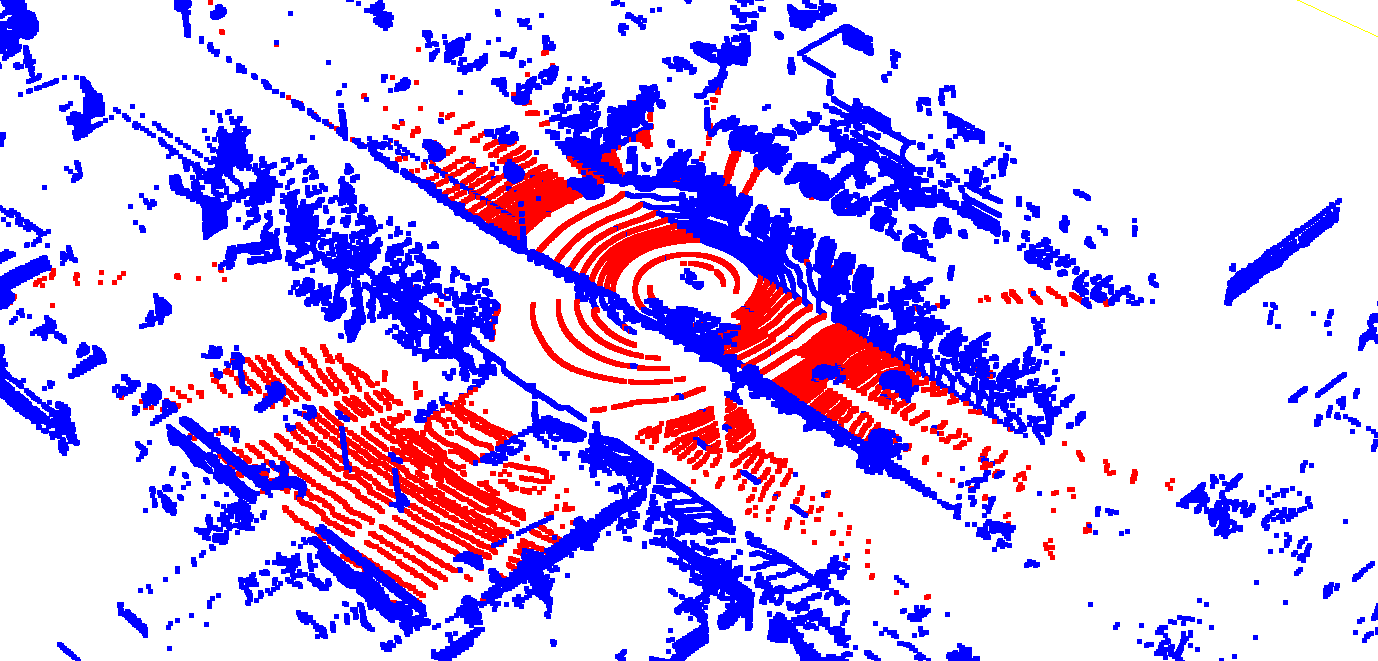}    
    \end{minipage}
    \begin{minipage}{0.24\linewidth}
    \includegraphics[width=\linewidth]{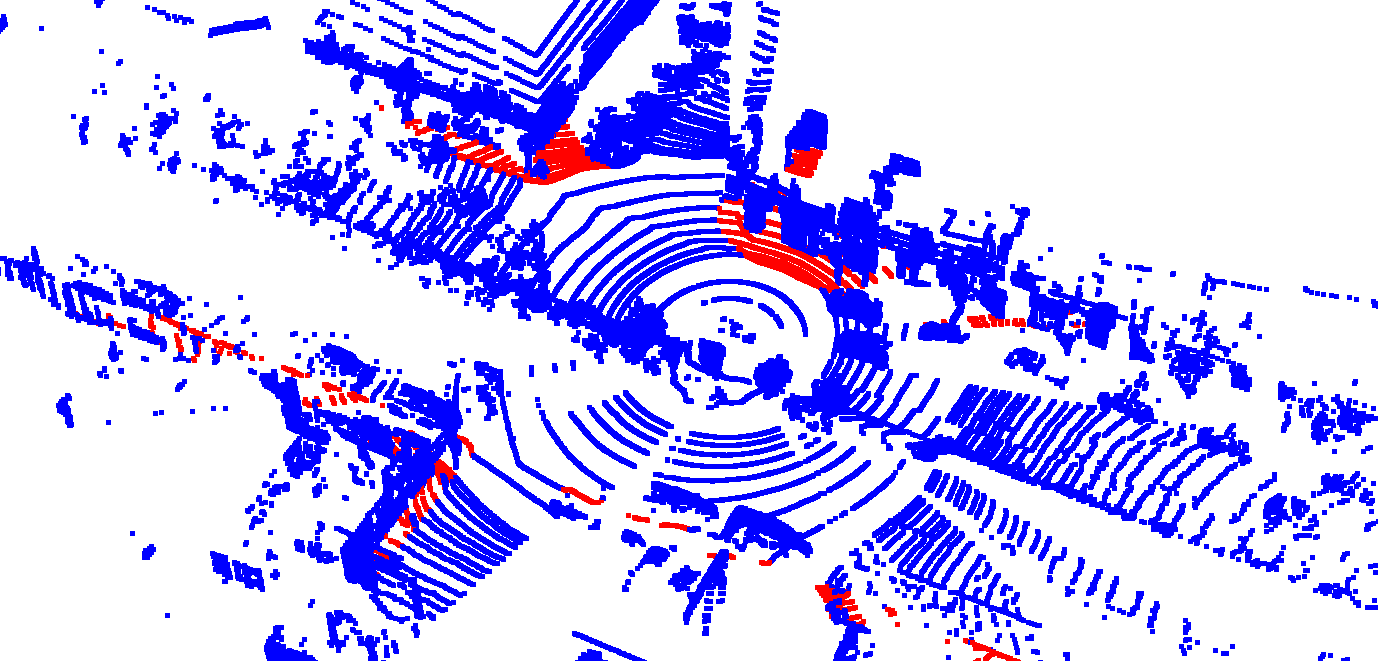}    
    \end{minipage}
    \vspace{1mm}
    \\    
    \hrule
    \vspace{1mm}
    \begin{minipage}{3mm}
    \rotatebox[origin=c]{90}{Pan. GT - Sim}
    \end{minipage}
    \begin{minipage}{0.24\linewidth}
    \includegraphics[width=\linewidth]{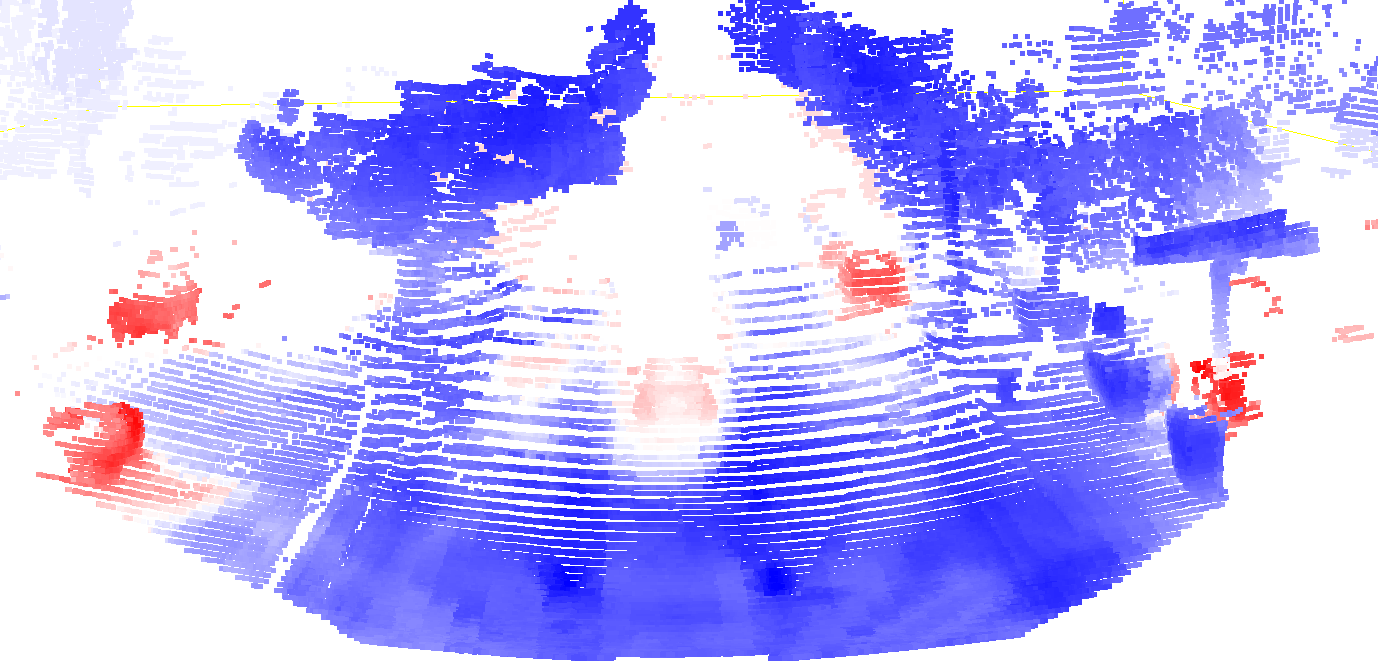}
    \end{minipage}
    \begin{minipage}{0.24\linewidth}
    \includegraphics[width=\linewidth]{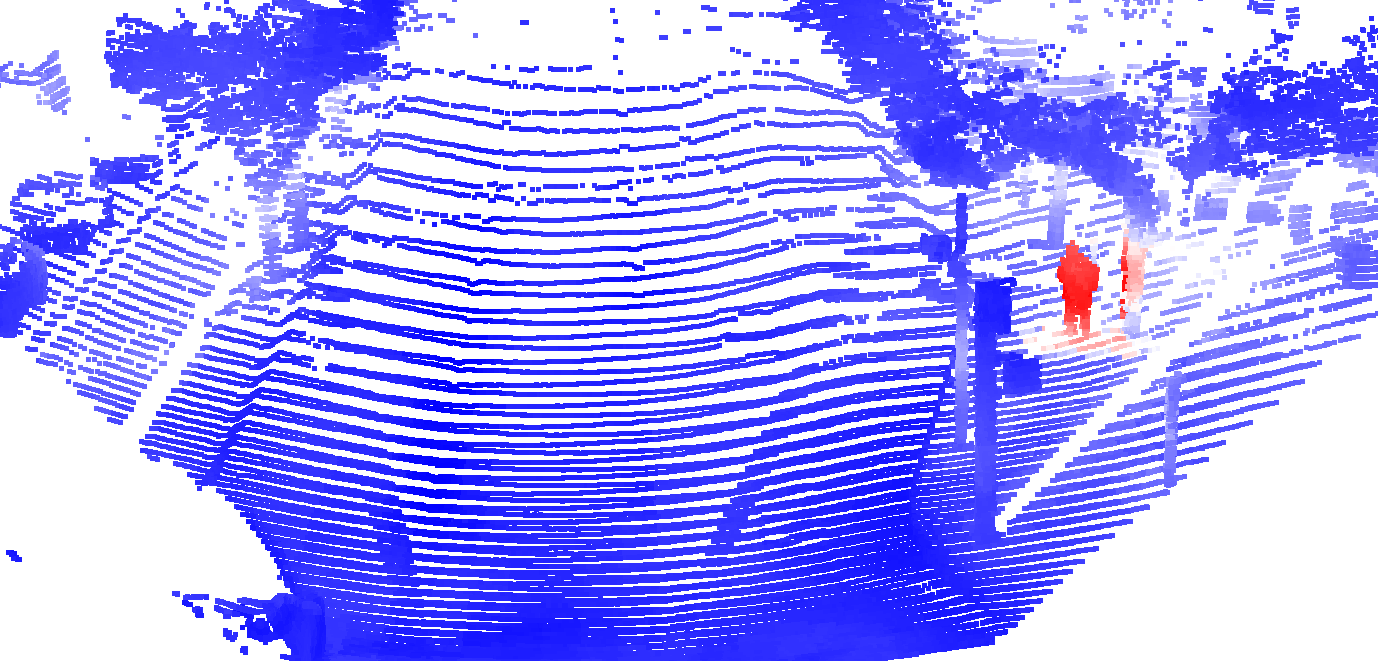}
    \end{minipage}
    \begin{minipage}{0.24\linewidth}
    \includegraphics[width=\linewidth]{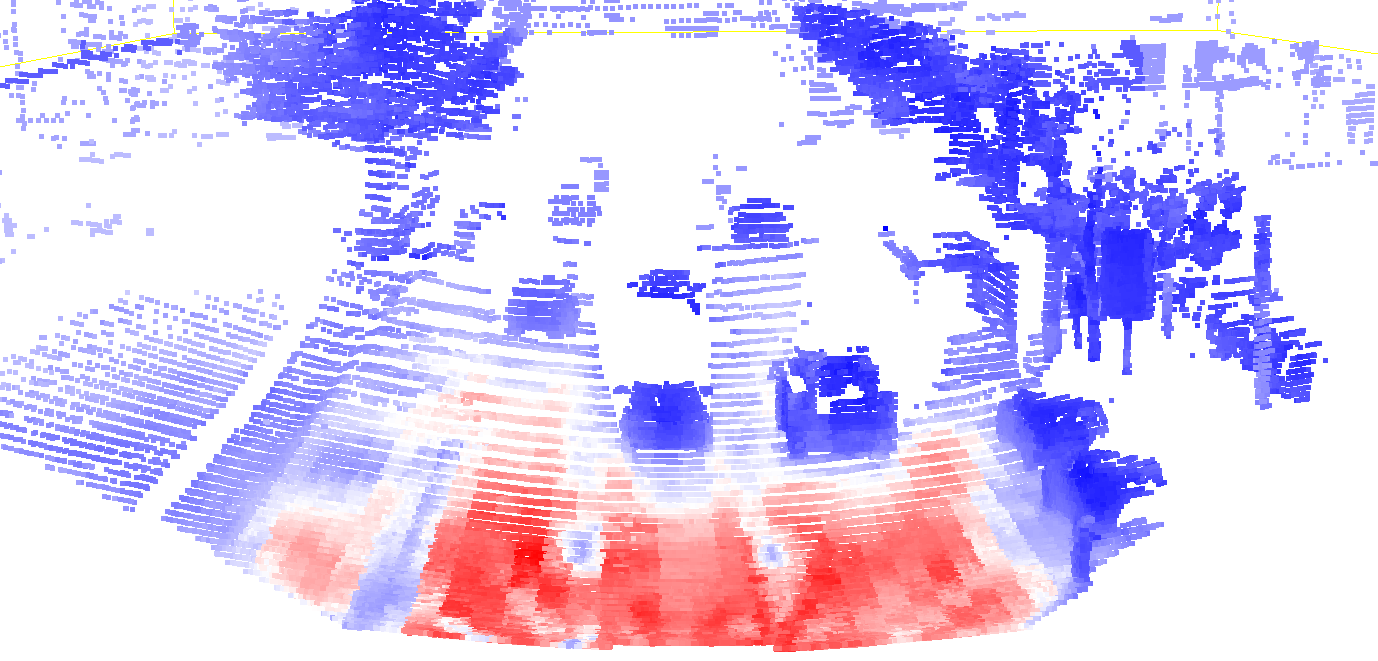}
    \end{minipage}
    \begin{minipage}{0.24\linewidth}
    \includegraphics[width=\linewidth]{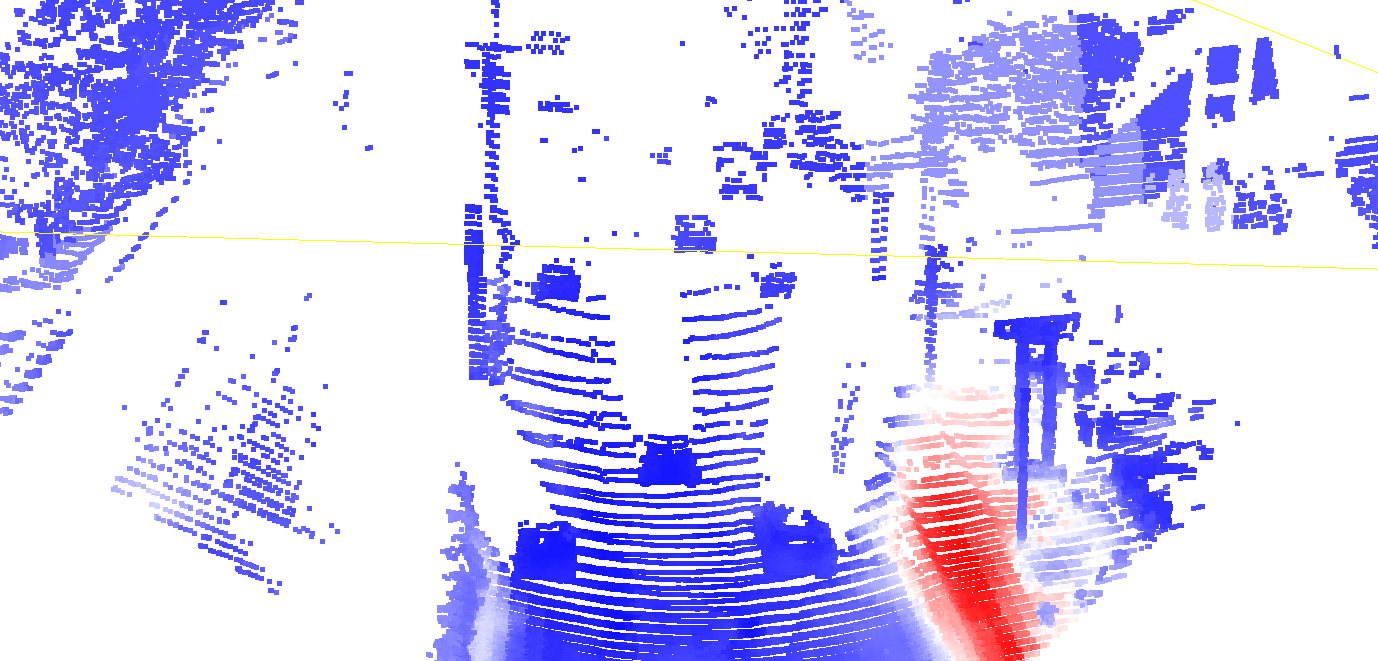}
    \end{minipage}    
    \\
    \begin{minipage}{3mm}
    \rotatebox[origin=c]{90}{Pan. GT - Label}
    \end{minipage}
    \begin{minipage}{0.24\linewidth}
    \includegraphics[width=\linewidth]{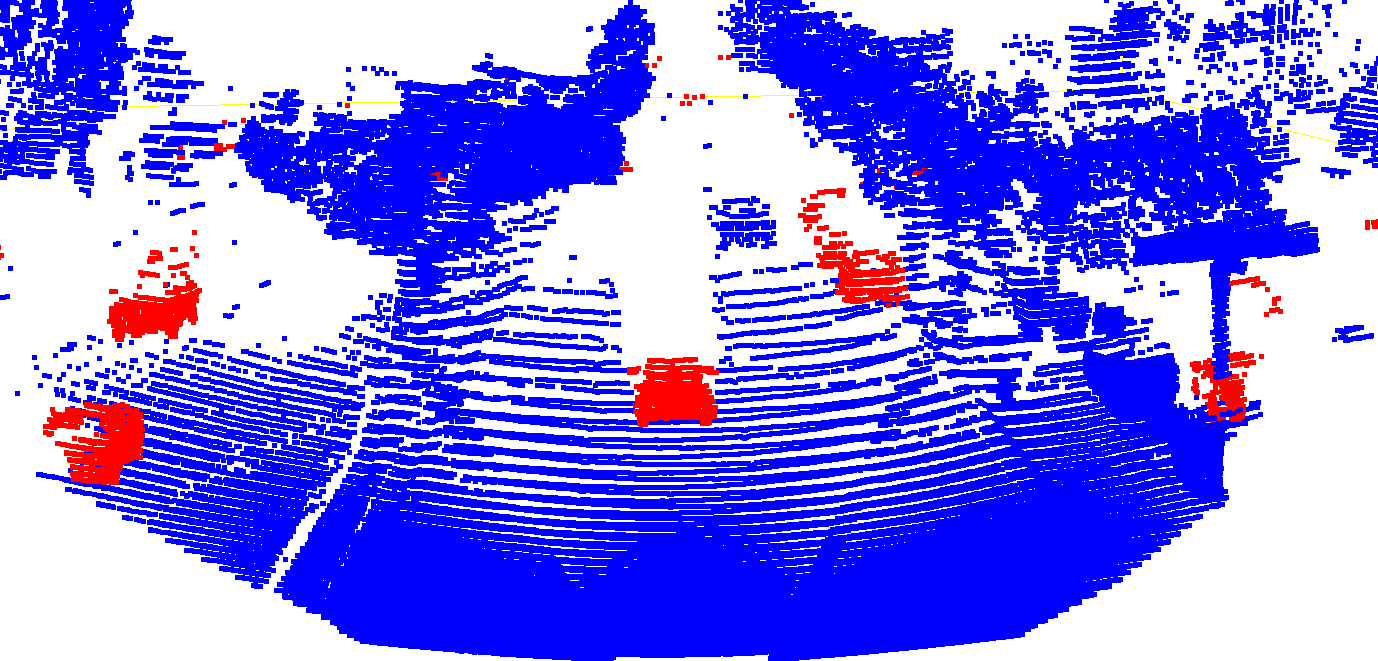}    
    \end{minipage}
    \begin{minipage}{0.24\linewidth}
    \includegraphics[width=\linewidth]{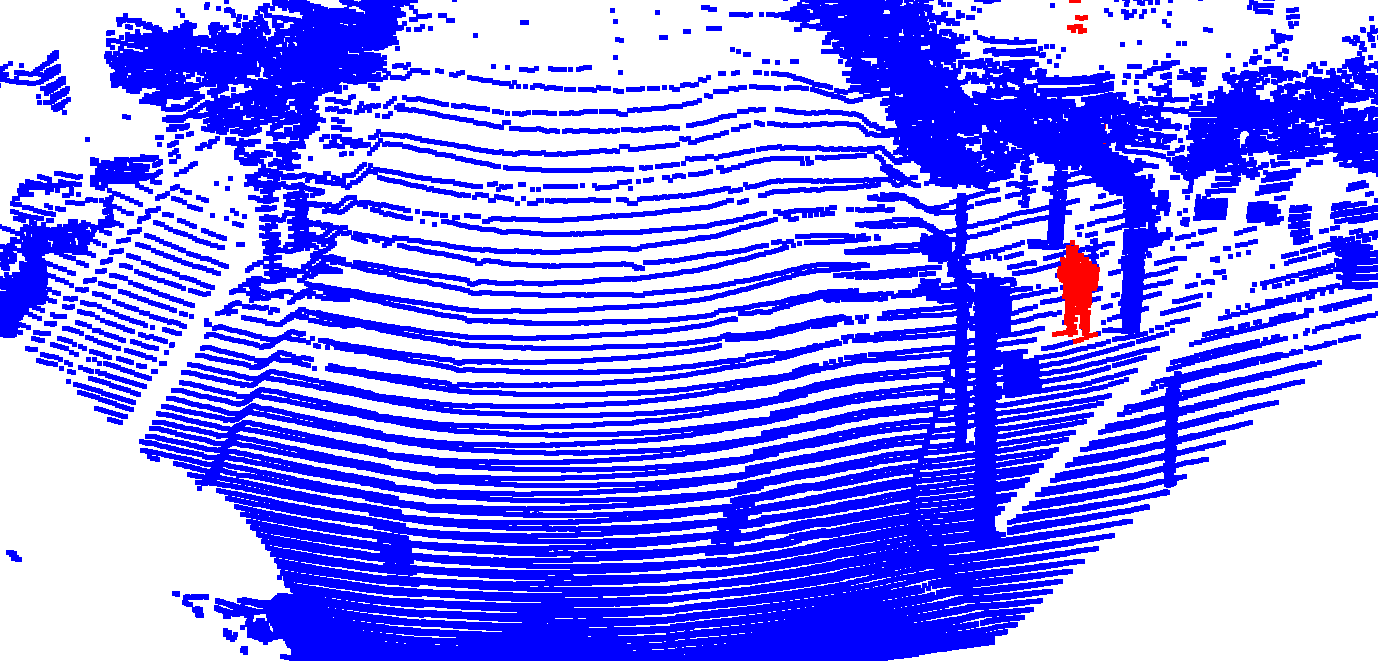}    
    \end{minipage}
    \begin{minipage}{0.24\linewidth}
    \includegraphics[width=\linewidth]{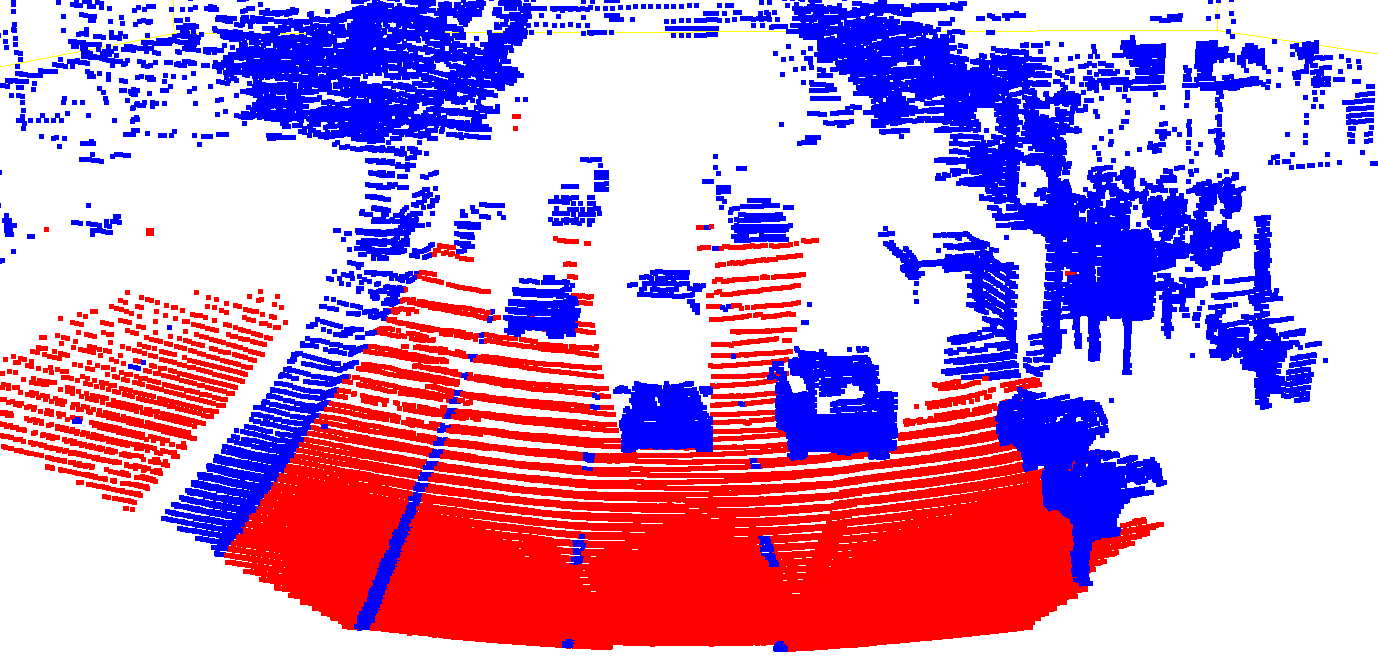}
    \end{minipage}
    \begin{minipage}{0.24\linewidth}
    \includegraphics[width=\linewidth]{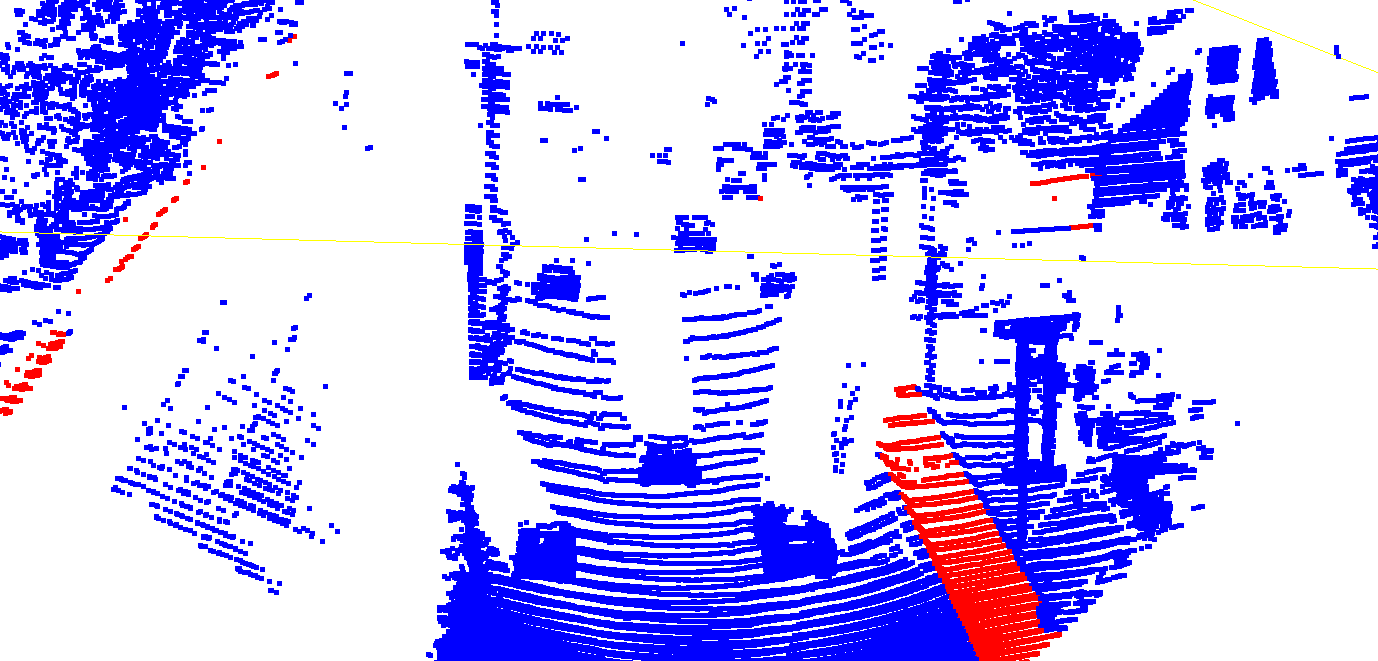}    
    \end{minipage}
    \caption{\textbf{Similarity map with class prototype}. For each scan, we use the ground-truth labels (presented on even rows) of four classes (car, pedestrian, road, sidewalk) to compute a class prototype (mean feature of the point belonging to the considered class). We then compute the feature similarity map (presented on odd rows) with respect to that class prototype. Color goes from blue to red for low and high values.}
    \label{fig:class_prototype}
\end{figure*}

We continue our visual inspection of the distilled features by presenting feature similarity map with respect to class prototypes in \cref{fig:class_prototype}. The features are extracted at the output of $\backthreed$ and are $\ell_2$-normalized. The similarity maps are then obtained as follows. For each scan, we use the ground-truth labels to extract the point features of a class of interest (car, pedestrian, road or sidewalk). We average all the corresponding features to obtain a single class prototype for that class. Finally, we compute the similarity of all point features with respect to this class prototype. This is a similar procedure as the one used in Fig.~\textcolor{red}{1} but using a mean feature instead of a single point feature.

In all cases, we notice that the most similar features to a class prototype belongs to corresponding class, as expected. This is another indication that the feature space is well structured where: the features of a same semantic class are close to each other; the features of two different semantic class are well separated. Nevertheless, when inspecting closely the similarity map, we notice sometimes some ``leakage'' around the objects of interest. This phenomenon is mostly visible for the class pedestrian. We believe that these artifacts are due to errors when projecting the points onto the camera plane, which affects the boundary of the objects. Finally, we remark as well that the similarity maps are less sharp on Pandar64 and PandarGT than on nuScenes and SemanticKITTI, likely because of the small number of scans available in Pandaset.

\section{Limitations}

Our work raises the possibility of replicating undesirable biases present in the large pretrained 2D models used for distillation. These models are known to harbor problematic biases related, e.g., to geographic location, gender, skin tone, and age. When distilling these 2D vision models into 3D lidar models, there is a potential for these biases to be amplified or mirrored. Our resulting lidar models may exhibit varying performance across different geographical regions, influenced by how these regions are represented in the training datasets of the original 2D models and in the 2D-to-3D distillation training data. For real-world applications of this distillation strategy, practitioners are expected to be mindful about the 2D foundation model used and the nature of the data it was trained upon (e.g., potential biases, privacy breaches, licenses, etc.)

Our study in Sec.~\textcolor{red}{3.3} shows that the linear probing mIoU has a standard deviation around $1.0$ percentage point between different pretrainings. Some possibilities to reduce these small fluctuations might be to explore longer pretraining schedules, or re-increase the number of loaded images per scan (from 1 to 6).

\end{document}